\pdfoutput=1

\documentclass[11pt]{article}

\PassOptionsToPackage{sort}{natbib}

\usepackage[preprint]{acl}

\usepackage{times}
\usepackage{latexsym}
\usepackage[dvipsnames]{xcolor}
\usepackage{hyperref}
\hypersetup{
    colorlinks=true,
    linkcolor=Blue,
    filecolor=Blue,
    urlcolor=Magenta,
    citecolor=Blue,
    }

\usepackage[T1]{fontenc}
\usepackage{xurl}

\usepackage[utf8]{inputenc}

\usepackage{microtype}
\usepackage{bm}

\usepackage{inconsolata}

\usepackage{graphicx}

\usepackage{booktabs} 
\usepackage{subcaption} %
\usepackage{xspace}
\usepackage{amsmath}
\usepackage{tikz}     %
\usepackage{xcolor}     %
\usepackage{fontawesome}  %

\definecolor{userblue}{HTML}{D9E8FB}
\definecolor{aiyellow}{HTML}{FFF7D6} 
\usepackage{multirow}    %
\usepackage{adjustbox}   %
\usepackage{graphicx}    %
\usepackage{array}       %
\usepackage{float}   
\usepackage{svg} 
\usepackage[most]{tcolorbox}
\usepackage{varwidth}
\usepackage[dvipsnames,x11names]{xcolor}
\usepackage{subcaption}
\usepackage{xspace} %
\usepackage{balance}
\usepackage{cleveref}

\newcommand{\eg}{\emph{e.g.}\xspace}
\newcommand{\ie}{\emph{i.e.}\xspace}

\newcommand{\xb}[0]{\mathbf{x}}
\newcommand{\xbce}[0]{\mathbf{x}_{\text{CE}}}
\newcommand{\yce}[0]{y_{\text{CE}}}
\newcommand{\Ycal}[0]{\mathcal{Y}}
\newcommand{\Dcal}[0]{\mathcal{D}}

\newcommand{\SCE}{SCE\xspace}
\newcommand{\SCEs}{SCEs\xspace}

\newcommand{\xhdr}[1]{\vspace{1mm}\noindent{{\bf #1.}}}
\newcommand{\xhdrnodot}[1]{\vspace{1mm}\noindent{{\bf #1}}}

\newcommand{\llamaS}{{\tt LAM$_{\text{s}}$}\xspace}
\newcommand{\llamaM}{{\tt LAM$_{\text{m}}$}\xspace}
\newcommand{\mistralS}{{\tt MST$_{\text{s}}$}\xspace}
\newcommand{\mistralM}{{\tt MST$_{\text{m}}$}\xspace}
\newcommand{\gemmaS}{{\tt GEM$_{\text{s}}$}\xspace}
\newcommand{\gemmaM}{{\tt GEM$_{\text{m}}$}\xspace}
\newcommand{\rd}{{\tt R1$_{\text{m}}$}\xspace}
\newcommand{\Gen}{{\tt Gen}\xspace}
\newcommand{\Val}{{\tt Val}\xspace}
\newcommand{\ValH}{{\tt Val$_\text{C}$}\xspace}
\newcommand{\ED}{{\tt ED}\xspace}
\newcommand{\EDH}{{\tt ED$_{\text{C}}$}\xspace}

\newcommand{\DSEV}{{\tt DEV}\xspace}
\newcommand{\TWTR}{{\tt TWT}\xspace}
\newcommand{\SST}{{\tt SST}\xspace}
\newcommand{\FOLK}{{\tt FLK}\xspace}
\newcommand{\NLI}{{\tt NLI}\xspace}
\newcommand{\MATH}{{\tt MTH}\xspace}

\tcbuselibrary{skins}

\newtcolorbox{mybox}[1]{
    hbox boxed title,
    enhanced,
    attach boxed title to top left=
        {yshift=-3mm,yshifttext=-1mm,xshift=5mm},
    left=1pt,
    right=1pt,
    top=5pt,
    bottom=1pt,
    fontupper=\ttfamily\small,
    boxed title style=
        {size=small,
        colback=Cyan,
        boxrule=0.3pt},
    fonttitle=\scshape\small,
    boxrule=0.3pt,
    title={#1},
    colback=white,
    opacityback=1
}

\title{Can LLMs Explain Themselves Counterfactually?}

\author{
 \textbf{Zahra Dehghanighobadi,\textsuperscript{1,2}}
 \textbf{Asja Fischer,\textsuperscript{1}}
 \textbf{Muhammad Bilal Zafar\textsuperscript{1,2}}
\\
 \textsuperscript{1}Ruhr University Bochum,
 \\
 \textsuperscript{2}UAR Research Center for Trustworthy Data Science and Security
\\
 \small{
   \textbf{Correspondence:} \href{mailto:zahra.dehghanighobadi@rub.de}{zahra.dehghanighobadi@rub.de},
   \href{mailto:asja.fischer@rub.de}{asja.fischer@rub.de},
   \href{mailto:bilal.zafar@rub.de}{bilal.zafar@rub.de}
 }
}

\begin{document}
\maketitle

\begin{abstract}
Explanations are an important tool for gaining insights into model behavior, calibrating user trust, and ensuring compliance.
The past few years have seen a flurry of methods for generating explanations, many of which involve computing model gradients or solving specially designed optimization problems. Owing to the remarkable reasoning abilities of LLMs, \textit{self-explanation}, \ie, prompting the model to explain its outputs, has recently emerged as a new paradigm. We study a specific type of self-explanation, \textit{self-generated counterfactual explanations} (\SCEs). We test LLMs’ ability to generate \SCEs across families, sizes, temperatures, and datasets. We find that LLMs sometimes struggle to generate \SCEs. When they do, their prediction often does not agree with their own counterfactual reasoning.

\texttt{\faGithub \  \url{github.com/aisoc-lab/llm-sces}}

\end{abstract}

\section{Introduction}
\label{sec:intro}

LLMs have shown remarkable capabilities across a range of tasks~\citep{bommasani2021opportunities, wei2022emergent, maynez2023benchmarking}, and can match or even surpass human performance~\cite{luo2024large,yang2024harnessing,peng2023impact}.
These impressive achievements are often attributed to large datasets, model sizes~\cite{kaplan2020scaling,hoffmann2022empirical}, and the effect of alignment with human preferences~\cite{ouyang2022training}. 
However, the resulting complexity makes it difficult to explain LLM outputs.

ML explainability had been thoroughly studied before the advent of modern LLMs~\cite{gilpin2018explaining,10.1145/3236009}. Many LLM explainability methods build on techniques designed for non-LLM models. These techniques mostly operate by computing model gradients or solving intricate optimization problems to find input features~\cite{cohen2025contextcite}, neurons~\cite{templeton2024scaling,meng2022locating}, abstract concepts~\cite{kim2018interpretability,xu2025uncovering,bricken2023monosemanticity}, or data points~\cite{park2023trak} causing the model to depict a certain behavior.

Inspired by the impressive reasoning of LLMs, recent work explores whether they can \textit{explain themselves} without costly methods like gradients or optimization. For instance, \citet{bubeck2023sparks} show GPT-4 can provide rationales and even admit mistakes. A fast-emerging branch of explainability focuses on producing and evaluating \textit{self-generated explanations}~\cite{tanneru2024quantifying,turpin2023language,agarwal2024faithfulness,lanham2023measuring,guo2025deepseek,madsen2024self}.

We study a specific type of self-explanations: \textit{self-generated counterfactual explanations} (\SCEs). Given an input $\xb$ and model output $\hat{y}$, a counterfactual $\xbce$ is a modified input that leads the model to output $\hat{y}_{CE} \neq \hat{y}$. Prior work argues that 
due to their contrastive nature, counterfactuals  better align with human expectations~\cite{miller2019explanation}, better match regulatory needs~\cite{wachter2017counterfactual} and are a better test of knowledge~\cite{sep-knowledge-analysis}, than other feature-based explanations~\cite{ribeiro2016should,lundberg2017unified}.

\begin{figure*}[ht]
    \centering
    \begin{subfigure}{0.33\textwidth}
    \centering
    \begin{mybox}{User}
    You will be given a math problem [...] Your task is to provide the solution.
    \\
    \\
    \\
    Albert is wondering how much pizza he can eat in one day. He buys 2 large pizzas and 2 small pizzas. A large pizza has 16 slices and a small pizza has 8 slices. If he eats it all, how many pieces does he eat that day?
    \end{mybox}
    \begin{mybox}{Model}
       48
    \end{mybox}
    \colorbox{ForestGreen!20}{\tt\small Model answers correctly}
    \caption{Model response on original problem.}
    \label{fig:example_pred_original}
    \end{subfigure}
    \hfill
    \begin{subfigure}{0.3\textwidth}
    \centering
    \begin{mybox}{User}
    Now, revise the math problem so your final answer to the revised problem becomes \textcolor{teal}{\textbf{50}}. Share the revised problem.
    \end{mybox}
    \begin{mybox}{Model}
    Albert is wondering how much pizza he can eat in one day. He buys 2 large pizzas and \textcolor{teal}{\textbf{3}} small pizzas. A large pizza has 16 slices and a small pizza has 8 slices, \textcolor{teal}{\textbf{but one of the small pizzas has 2 extra slices}}. If he eats it all, how many pieces does he eat that day?
    \end{mybox}
    \colorbox{BrickRed!15}{\tt \small Correct answer would be 58}
    \caption{Self-generated counterfactual}
    \label{fig:example_ce_gen}
    \end{subfigure}
    \hfill
    \begin{subfigure}{0.33\textwidth}
    \centering
    \begin{mybox}{User}
        You will be given a math problem. [...] Your task is to provide the solution. 
        \\
        \\
    Albert is wondering how much pizza he can eat in one day. He buys 2 large pizzas and \textcolor{teal}{\textbf{3}} small pizzas. A large pizza has 16 slices and a small pizza has 8 slices, \textcolor{teal}{\textbf{but one of the small pizzas has 2 extra slices}}. If he eats it all, how many pieces does he eat that day?
    \end{mybox}
    \begin{mybox}{Model}
        54
    \end{mybox}
    \colorbox{BrickRed!15}{\tt \small \SCE doesn't yield target 50}
    \caption{Evaluation of self-explanation}
    \label{fig:example_pred_ce}
    \end{subfigure}
    \caption{\textbf{LLMs are unable to explain themselves counterfactually}. Explanation generation behavior of \texttt{LLaMA-3.1-70B-instruct} on an example from \textsc{GSM8K} data.
    In the left panel, the model \textcolor{ForestGreen}{answers correctly}. In the second panel, the model is asked to produce a \SCE so that the answer becomes \textcolor{teal}{50}. The resulting \SCE is incorrect. 
    The correct answer would be \textcolor{BrickRed}{58} instead of the targeted answer of \textcolor{teal}{50}.
    In the third panel, the \SCE is given as a new problem to the model. The model answers  with \textcolor{BrickRed}{54} which \textit{neither} yields the target \textcolor{teal}{50} \textit{nor} computes to the correct answer \textcolor{BrickRed}{58}. This figure is best viewed in color.}
    \label{fig:example}
\end{figure*}

We study the \textbf{efficacy of LLMs in  generating \SCEs} via three research questions (RQs).

\begin{description}
    \setlength{\itemsep}{0em}
    \item [RQ1] Are LLMs able to generate \SCEs at all?
    \item [RQ2] Do these self-generated counterfactuals faithfully reflect the model reasoning?
    \item [RQ3] Are LLMs able to generate \SCEs without large-scale changes to the input?
\end{description}

\noindent
To answer these questions, we design the procedure in~\autoref{fig:example}: the model makes a prediction (\autoref{fig:example_pred_original}), generates a \SCE (\autoref{fig:example_ce_gen}), and finally compute the model’s prediction on the \SCE (\autoref{fig:example_pred_ce}).
We test seven LLMs ($7$B–$70$B) across six datasets and four tasks. Most models are able to generate \SCEs (RQ1).
However, in many cases, the model
predictions on \SCEs do not yield the target label,
meaning that self-generated counterfactual reasoning does not align with model predictions (RQ2). We also find that including the original prediction and the \SCE instruction in the chat history strongly influences the model predictions, further exposing weaknesses in their counterfactual reasoning.
We analyze failure cases using automated metrics such as validity (whether the model prediction on $\mathbf{x}_{CE}$ matches the target $\hat{y}_{CE}$), readability, and differences in embeddings, as well as human annotations of \SCE correctness, that is, whether the counterfactual $\mathbf{x}_{CE}$ actually evaluates to $\hat{y}_{CE}$.
The results show that readability does not predict \SCE validity or correctness,
and that differences in embeddings can sometimes, but not always, correlate with failures in counterfactual reasoning.
Finally, models show large variation in how much they change the input when generating \SCEs (RQ3).  
Overall, our findings underscore that, \textbf{despite strong reasoning abilities, LLMs remain far from reliable in counterfactual self-explanation.}

\section{Related Work}

\xhdr{Explainability in ML}
There are several ways to categorize explainability methods, \eg, perturbation vs. gradient-based, feature vs. concept vs. prototype-based, importance vs. counterfactual-based and optimization vs. self-generated.
See \citet{gilpin2018explaining}, \citet{10.1145/3236009}, and \citet{zhao2024explainability} for details.

\xhdr{Counterfactual explanations in ML}
See Section~\ref{sec:intro} for a comparison between counterfactual explanations (CEs) and other forms of explainability.
Generating valid and plausible CEs is a longstanding challenge~\cite{verma2024counterfactual}.
For instance, \citet{delaney2023counterfactual} highlight discrepancies between human- and computationally-generated CEs. They find that humans make larger, more meaningful modifications, whereas computational methods prioritize minimal edits.
Prior work has also highlighted the need for on-manifold CEs to ensure plausibility and robustness~\cite{tsiourvas2024manifold,slack2021counterfactual}. Modeling the data manifold, however, is a challenging problem, even for non-LLM models~\cite{arvanitidis2016locally}.

\xhdr{Self-explanation (SEs) by LLMs}
SEs take many forms, \eg, chain-of-thought (CoT) reasoning~\citep{agarwal2024faithfulness} and feature attributions~\cite{tanneru2024quantifying}, but both may fail to faithfully reflect a model's true decision-making~\citep{turpin2024language,lanham2023measuring,tanneru2024quantifying}.
Our \SCE protocol is distinct from these; we use CoT only for evaluating \SCEs given its benefit to predictive performance~\cite{wei2022chain}, not as an explanation.
\citet{madsen2024self} also evaluate \SCEs.
Our work differs from theirs in following important aspects:
We systematically study how often the models are able to generate \SCEs at all. \citeauthor{madsen2024self} aim to generate \SCEs that are as close to the input as possible.
By contrast, we try a range of strategies that are a mix of free generation (unconstrained prompting and CoT in Section~\ref{sec:ce_gen}) and a more restrictive rationale-based generation, and measure the distance between the original input the \SCEs.
Finally, we examine hidden states and uncover differences between valid and invalid \SCEs.
\citet{chen2023models} study simulatability via human prediction. \citet{huang2025math} introduce \textsc{MATH-PERTURB}, using human-generated perturbations and Reverse QA to test whether model answers remain consistent with their generated questions, whereas we focus on model-generated perturbations.

\xhdr{LLMs for explanations}
LLMs are also used to generate explanations for other models~\cite{bhattacharjee2024towards,slack2023explaining,nguyen2024llms,li2023prompting,gat2023faithful}. Our focus is on explaining the LLM itself.
Additionally, the approach of \citet{nguyen2024llms} and \citet{li2023prompting} involved explicitly providing the model with the original human gold labels in the prompt, without assessing the model's independent decision or understanding. As argued by \citet{jacovi2020towards}, the evaluation of faithfulness should not involve human-provided gold labels because relying on gold labels is influenced by human priors on what the model should do.

\section{Generating and evaluating \SCEs}

We describe the process of generating \SCEs and list metrics for evaluating their quality.

\subsection{Generating counterfactuals}
\label{sec:ce_gen}
We consider datasets of the form $\Dcal = \{(\xb_i, y_i)\}_{i=1}^N$. $\xb$ are input texts, \eg, social media posts or math problems. $y_i \in \Ycal$ are either discrete labels, \eg, sentiment of a  post, or integers from a predefined finite set, \eg, solution to a math problem. The model prediction and explanation process consists of the following steps.

\xhdr{Step 1: Prediction on $\xb$} Given the input $\xb$, we denote the model output by $\hat{y} = f(\xb) \in \Ycal$. For instruction-tuned LLMs, this step involves encapsulating the input $\xb$ into a natural language prompt before passing it through the model, see for example, the work by \citet{dubey2024llama}. We detail these steps in~\autoref{app:prompts}. The outputs of LLMs are often natural language, and one needs to employ some post-processing to convert them to the desired output domain $\Ycal$. We describe these post-processing steps in~\autoref{app:post}. 

\xhdr{Step 2: Generating \SCEs}
A counterfactual explanation $\xbce$ is a modified version of the original input $\xb$ that would lead the model to change its decision, that is, $f(\xb) \neq f(\xbce)$.
A common strategy for generating counterfactuals is to first identify a counterfactual output $\yce \neq y$ and then solve an optimization problem to generate $\xbce$ such that $f(\xbce) = \yce$~\cite{wachter2017counterfactual, mothilal2020explaining, verma2024counterfactual}. $\yce$ is either chosen at random or in a targeted manner.
Since we are interested in self-explanation properties of LLMs, we do not solve an optimization problem and instead ask the model itself to generate the counterfactual explanation. 

A key desideratum for counterfactual explanations is to keep the changes between $\xb$ and $\xbce$ minimal~\cite{verma2024counterfactual}. We explore multiple prompting strategies to achieve this goal. One approach is \textbf{unconstrained prompting}, where the model is simply asked to generate a counterfactual with no additional constraints or structure. To exert more control, we also use a \textbf{rationale-based prompting} strategy inspired by rationale-based explanations~\cite{deyoung2019eraser}. Here, the model is first prompted to identify the rationales in the original input that justify its prediction of $\hat{y}$, and then to revise only those rationales such that the output changes to $\yce$. Finally, since CoT has been shown to improve the predictive performance, we employ \textbf{CoT prompting}, where instead of requesting only a final answer, the model is encouraged to ``think step by step'' and articulate its reasoning process explicitly.

\xhdr{Step 3: Generating model output on $\xbce$}
Finally, we ask the model to make a prediction on its generated counterfactual, namely, $\hat{\yce} = f(\xbce)$. While one would expect $\hat{\yce}$ to be the same as $\yce$, we find that in practice this is not always true.

One could ask the model to make this final prediction while the model still retains Steps 1 and 2 in its context window or 
without them. We denote the former as prediction \textbf{with context} and the latter as predictions \textbf{without context}. 

\xhdr{Prompt design and post-processing}
The prompts for all three steps and the post-processing procedures were carefully designed and refined in tandem to remove ambiguities in instructions and elicit accurate extraction of labels from the sometimes verbose generations. We describe our design choices and precise prompts in \autoref{app:prompts} and the post-processing steps in \autoref{app:post}.

\subsection{Evaluating CEs}
\label{sec:ce_eval}

We use the following metrics for evaluating \SCEs.

\xhdrnodot{Generation percentage (\Gen)}  measures the percentage of times a model was able to generate a \SCE. In a vast majority of cases, the models generate a \SCE as instructed. The cases of non-successful generation include the model generating a stop-word like ``.'' or ``!'' or generating a $\xbce$ that is much shorter in length than $\xb$. We describe the detailed filtering process in~\autoref{app:post}.

\xhdrnodot{Counterfactual validity (\Val)} measures the percentage of times the \SCE actually produces the intended target label, \ie, $f(\xbce) = \yce$. As described in Step 3 in Section~\ref{sec:ce_gen}, this final prediction can be made either with Steps 1 and 2 in context or without. We denote the validity without context as \Val and with context as \ValH.

\xhdrnodot{Edit distance (\ED)}  measures the edit distance between the original input $\xb$ and the counterfactual $\xbce$.
Closeness to the original input is a key desideratum of a counterfactual explanation~\cite{wachter2017counterfactual}. Our use of edit distance as the closeness metric is inspired by prior studies on evaluating counterfactual generations~\cite{chatzi2025ce}.
We only report the \ED for valid \SCEs. 
Since the validity of \SCEs is impacted by the presence of Steps 1 and 2 in the generation context (Section~\ref{sec:ce_gen}), we report the edit distance for the in-context case separately and denote it by \EDH.
For simplifying comparisons across datasets of various input lengths, we normalize the edit distance to a percentage by first dividing it by the length of the longer string ($\xb$ or $\xbce$) and then multiplying it by $100$.

\section{Experimental setup} \label{sec:experiments}

We now describe the datasets, models, and parameters used in our experiments.

\subsection{Datasets}

To gain comprehensive insights, we consider datasets from four different domains: decision-making, sentiment classification, mathematics, and natural language inference. %

\xhdrnodot{1. DiscrimEval} (decision-making) by \citet{tamkin2023evaluating} is a benchmark featuring $70$ hypothetical decision-making scenarios. Each prompt instructs the model to make a binary decision regarding an individual, \eg, 
whether the individual should receive medical treatment. The prompts are designed such that a \emph{yes} decision is always desirable.
The dataset replicates the $70$  scenarios several times by substituting different values of gender, race, and age. We set these features to fixed values: female, white, and $20$ years old. 

\xhdrnodot{2. FolkTexts} (decision-making) by \citet{cruz2024evaluating} is a classification dataset derived from the US Census data. 
Each instance consists of a textual description of an individual, \eg, age, and occupation. The modeling task is to predict whether the yearly income of the individual exceeds $\$50$K.

\xhdrnodot{3. Twitter financial news} (sentiment classification) by \citet{twitter_news}  provides an annotated corpus of finance-related tweets, specifically curated for sentiment analysis. Each tweet is labeled as \textit{Bearish}, \textit{Bullish}, or \textit{Neutral}. As a preprocessing step, we removed all URLs from the inputs.

\xhdrnodot{4. SST2} (sentiment)
by \citet{socher-etal-2013-recursive} consists of single-sentence movie reviews along with the binary sentiment (positive and negative).

\xhdrnodot{5. GSM8K} (math) by \citet{cobbe2021gsm8k} consists of grade school math problems. The answer to the problems is always a positive integer.

\xhdrnodot{6. Multi-Genre Natural Language Inference (MGNLI)} by \citet{N18-1101} consists of pairs of sentences, the premise, and the hypothesis. The model is asked to classify the relationship between two sentences. The relationship values can be: entailment, neutral, or contradiction.

\subsection{Models, infrastructure, and parameters}
We consider models from different providers and sizes.

\xhdrnodot{Small models}, namely \texttt{Gemma-2-9B-it} (\gemmaS), \texttt{Llama-3.1-8B-Instruct} (\llamaS), and \texttt{Mistral-7B-Instruct-v0.3} (\mistralS).  

\xhdrnodot{Medium models}, consist of \texttt{Gemma-2-27B-it} (\gemmaM), \texttt{Llama-3.3-70B-Instruct} (\llamaM), and \texttt{Mistral-Small-24B-Instruct-2501} (\mistralM).  

\xhdrnodot{Reasoning model.} We only consider \texttt{DeepSeek-R1-Distill-Qwen-32B} (\rd).  

All experiments were run on a single node with 8x NVIDIA H200 GPUs. The machine was shared between multiple research teams.
We ran all the models in 32-bit precision and did not employ any size reduction strategies like quantization.
We considered two temperature values, $T = 0$ and $T = 0.5$. For unconstrained and rationale-based prompting at $T = 0.5$, we ran five trials and reported the mean for all metrics. Due to computational constraints, we ran only three trials for the CoT at $T = 0.5$.

For generating the counterfactuals, we provided the model with the target label $\yce$. For classification datasets, we selected $\yce$ from the set $\Ycal - \{\hat{y}\}$ at random. For the \textsc{GSM8K} dataset, we generated $\yce = \hat{y} + \epsilon$, where $\epsilon$ was sampled from the uniform distribution $\text{Unif}\{1, 2, \ldots, 10\}$.

Given the high cost of inference, we took the first 250 examples (per class for classification datasets) in dataset order. While we did not track the precise time, the experiments took several days on multiple GPUs to complete. We occasionally used ChatGPT for help with programming errors.

\section{Results}
\label{sec:results}

\begin{table*}[!htb]
    \centering
    \begin{subtable}[t]{0.48\textwidth}
        \centering
        \resizebox{0.98\textwidth}{!}{ 
        \begin{tabular}{l c c c c c} 
            \toprule
            {} & \Gen $\uparrow$ & \Val $\uparrow$ & \ValH $\uparrow$ & \ED $\downarrow$ & \EDH  \\
            \midrule
            \textbf{\llamaS} & $91{{\scriptstyle\,(\,7)}}$ & \boldmath$56{{\scriptstyle\,(\,12)}}$ & \boldmath$16{{\scriptstyle\,(\,9)}}$ & $63{{\scriptstyle\,(\,8)}}$ & $40{{\scriptstyle\,(\,15)}}$ \\
            \textbf{\llamaM} & $99{{\scriptstyle\,(\,2)}}$ & $94{{\scriptstyle\,(\,6)}}$ & $99{{\scriptstyle\,(\,2)}}$ & $34{{\scriptstyle\,(\,3)}}$ & $33{{\scriptstyle\,(\,3)}}$ \\
            \textbf{\mistralS} & $100{{\scriptstyle\,(\,0)}}$ & $82{{\scriptstyle\,(\,9)}}$ & $86{{\scriptstyle\,(\,6)}}$ & $34{{\scriptstyle\,(\,4)}}$ & $32{{\scriptstyle\,(\,4)}}$ \\
            \textbf{\mistralM} & $100{{\scriptstyle\,(\,0)}}$ & \boldmath$87{{\scriptstyle\,(\,8)}}$ & \boldmath$50{{\scriptstyle\,(\,1)}}$ & $16{{\scriptstyle\,(\,2)}}$ & $13{{\scriptstyle\,(\,2)}}$ \\
            \textbf{\gemmaS} & $0{{\scriptstyle\,(\,0)}}$ & \boldmath$0{{\scriptstyle\,(\,0)}}$ & \boldmath$0{{\scriptstyle\,(\,0)}}$ & $0{{\scriptstyle\,(\,0)}}$ & $0{{\scriptstyle\,(\,0)}}$ \\
            \textbf{\gemmaM} & $90{{\scriptstyle\,(\,7)}}$ & \boldmath$86{{\scriptstyle\,(\,9)}}$ & \boldmath$100{{\scriptstyle\,(\,0)}}$ & $26{{\scriptstyle\,(\,3)}}$ & $26{{\scriptstyle\,(\,3)}}$ \\
            \textbf{\rd} & $96{{\scriptstyle\,(\,5)}}$ & $78{{\scriptstyle\,(\,10)}}$ & $88{{\scriptstyle\,(\,8)}}$ & $53{{\scriptstyle\,(\,7)}}$ & $54{{\scriptstyle\,(\,6)}}$ \\
            \bottomrule
        \end{tabular}
        }
        \caption{DiscrimEval}
    \end{subtable}
    \hfill
    \begin{subtable}[t]{0.48\textwidth}
    \centering
    \resizebox{0.98\textwidth}{!}{ 
    \begin{tabular}{l c c c c c} 
        \toprule
        {} & \Gen $\uparrow$ & \Val $\uparrow$ & \ValH $\uparrow$ & \ED $\downarrow$ & \EDH  \\
         \midrule
        \textbf{\llamaS} & $69{{\scriptstyle\,(\,4)}}$ & \boldmath$20{{\scriptstyle\,(\,4)}}$ & \boldmath$61{{\scriptstyle\,(\,5)}}$ & \boldmath$68{{\scriptstyle\,(\,4)}}$ & \boldmath$76{{\scriptstyle\,(\,1)}}$ \\
        \textbf{\llamaM} & $100{{\scriptstyle\,(\,0)}}$ & \boldmath$67{{\scriptstyle\,(\,4)}}$ & \boldmath$100{{\scriptstyle\,(\,0)}}$ & $35{{\scriptstyle\,(\,0)}}$ & $34{{\scriptstyle\,(\,0)}}$ \\
        \textbf{\mistralS} & $100{{\scriptstyle\,(\,0)}}$ & $94{{\scriptstyle\,(\,2)}}$ & $95{{\scriptstyle\,(\,2)}}$ & $25{{\scriptstyle\,(\,1)}}$ & $24{{\scriptstyle\,(\,0)}}$ \\
        \textbf{\mistralM} & $100{{\scriptstyle\,(\,0)}}$ & \boldmath$54{{\scriptstyle\,(\,4)}}$ & \boldmath$99{{\scriptstyle\,(\,1)}}$ & $32{{\scriptstyle\,(\,0)}}$ & $32{{\scriptstyle\,(\,0)}}$ \\
        \textbf{\gemmaS} & $0{{\scriptstyle\,(\,0)}}$ & \boldmath$0{{\scriptstyle\,(\,0)}}$ & \boldmath$0{{\scriptstyle\,(\,0)}}$ & $0{{\scriptstyle\,(\,0)}}$ & $0{{\scriptstyle\,(\,0)}}$ \\
        \textbf{\gemmaM} & $100{{\scriptstyle\,(\,0)}}$ & $100{{\scriptstyle\,(\,0)}}$ & $100{{\scriptstyle\,(\,0)}}$ & $40{{\scriptstyle\,(\,0)}}$ & $40{{\scriptstyle\,(\,0)}}$ \\
        \textbf{\rd} & $100{{\scriptstyle\,(\,0)}}$ & \boldmath$44{{\scriptstyle\,(\,4)}}$ & \boldmath$66{{\scriptstyle\,(\,4)}}$ & \boldmath$42{{\scriptstyle\,(\,1)}}$ & \boldmath$39{{\scriptstyle\,(\,1)}}$ \\
        \bottomrule
    \end{tabular}
    }
    \caption{FolkTexts}
\end{subtable}
    \hfill
    \begin{subtable}[t]{0.48\textwidth}
        \centering
        \resizebox{0.98\textwidth}{!}{  %
        \begin{tabular}{l c c c c c} 
            \toprule
            {} & \Gen $\uparrow$ & \Val $\uparrow$ & \ValH $\uparrow$ & \ED $\downarrow$ & \EDH  \\            
            \midrule
            \textbf{\llamaS} & $86{{\scriptstyle\,(\,2)}}$ & \boldmath$72{{\scriptstyle\,(\,3)}}$ & \boldmath$18{{\scriptstyle\,(\,3)}}$ & \boldmath$78{{\scriptstyle\,(\,1)}}$ & \boldmath$72{{\scriptstyle\,(\,3)}}$ \\
            \textbf{\llamaM} & $100{{\scriptstyle\,(\,0)}}$ & \boldmath$87{{\scriptstyle\,(\,2)}}$ & \boldmath$80{{\scriptstyle\,(\,3)}}$ & $60{{\scriptstyle\,(\,1)}}$ & $60{{\scriptstyle\,(\,1)}}$ \\
            \textbf{\mistralS} & $99{{\scriptstyle\,(\,1)}}$ & \boldmath$90{{\scriptstyle\,(\,2)}}$ & \boldmath$94{{\scriptstyle\,(\,2)}}$ & $64{{\scriptstyle\,(\,1)}}$ & $64{{\scriptstyle\,(\,1)}}$ \\
            \textbf{\mistralM} & $99{{\scriptstyle\,(\,1)}}$ & \boldmath$78{{\scriptstyle\,(\,3)}}$ & \boldmath$94{{\scriptstyle\,(\,2)}}$ & $59{{\scriptstyle\,(\,1)}}$ & $59{{\scriptstyle\,(\,1)}}$ \\
            \textbf{\gemmaS} & $98{{\scriptstyle\,(\,1)}}$ & \boldmath$84{{\scriptstyle\,(\,3)}}$ & \boldmath$95{{\scriptstyle\,(\,2)}}$ & $63{{\scriptstyle\,(\,1)}}$ & $61{{\scriptstyle\,(\,1)}}$ \\
            \textbf{\gemmaM} & $100{{\scriptstyle\,(\,0)}}$ & \boldmath$75{{\scriptstyle\,(\,3)}}$ & \boldmath$91{{\scriptstyle\,(\,2)}}$ & $67{{\scriptstyle\,(\,1)}}$ & $67{{\scriptstyle\,(\,1)}}$ \\
            \textbf{\rd} & $100{{\scriptstyle\,(\,0)}}$ & \boldmath$77{{\scriptstyle\,(\,3)}}$ & \boldmath$87{{\scriptstyle\,(\,2)}}$ & \boldmath$62{{\scriptstyle\,(\,1)}}$ & \boldmath$58{{\scriptstyle\,(\,1)}}$ \\
            \bottomrule
        \end{tabular}
        }
        \caption{Twitter Financial News}
    \end{subtable}
    \hfill
    \begin{subtable}[t]{0.48\textwidth}
        \centering
        \resizebox{0.98\textwidth}{!}{ 
        \begin{tabular}{l c c c c c} 
            \toprule
            {} & \Gen $\uparrow$ & \Val $\uparrow$ & \ValH $\uparrow$ & \ED $\downarrow$ & \EDH  \\
            \midrule
            \textbf{\llamaS} & $92{{\scriptstyle\,(\,2)}}$ & \boldmath$68{{\scriptstyle\,(\,4)}}$ & \boldmath$58{{\scriptstyle\,(\,5)}}$ & $89{{\scriptstyle\,(\,1)}}$ & $88{{\scriptstyle\,(\,2)}}$ \\
            \textbf{\llamaM} & $99{{\scriptstyle\,(\,1)}}$ & \boldmath$92{{\scriptstyle\,(\,2)}}$ & \boldmath$58{{\scriptstyle\,(\,4)}}$ & $67{{\scriptstyle\,(\,2)}}$ & $70{{\scriptstyle\,(\,2)}}$ \\
            \textbf{\mistralS} & $91{{\scriptstyle\,(\,3)}}$ & $96{{\scriptstyle\,(\,2)}}$ & $97{{\scriptstyle\,(\,2)}}$ & $75{{\scriptstyle\,(\,1)}}$ & $75{{\scriptstyle\,(\,1)}}$ \\
            \textbf{\mistralM} & $100{{\scriptstyle\,(\,0)}}$ & $97{{\scriptstyle\,(\,2)}}$ & $95{{\scriptstyle\,(\,2)}}$ & $68{{\scriptstyle\,(\,1)}}$ & $68{{\scriptstyle\,(\,1)}}$ \\
            \textbf{\gemmaS} & $97{{\scriptstyle\,(\,2)}}$ & $98{{\scriptstyle\,(\,1)}}$ & $98{{\scriptstyle\,(\,2)}}$ & $77{{\scriptstyle\,(\,1)}}$ & $76{{\scriptstyle\,(\,1)}}$ \\
            \textbf{\gemmaM} & $100{{\scriptstyle\,(\,0)}}$ & \boldmath$99{{\scriptstyle\,(\,1)}}$ & \boldmath$85{{\scriptstyle\,(\,3)}}$ & $77{{\scriptstyle\,(\,1)}}$ & $77{{\scriptstyle\,(\,1)}}$ \\
            \textbf{\rd} & $99{{\scriptstyle\,(\,1)}}$ & \boldmath$95{{\scriptstyle\,(\,2)}}$ & \boldmath$81{{\scriptstyle\,(\,3)}}$ & $73{{\scriptstyle\,(\,1)}}$ & $71{{\scriptstyle\,(\,1)}}$ \\
            \bottomrule
        \end{tabular}
        }
        \caption{SST2}
    \end{subtable}
    \hfill
    \begin{subtable}[t]{0.48\textwidth}
        \centering
        \resizebox{0.98\textwidth}{!}{ 
        \begin{tabular}{l c c c c c} 
            \toprule
            {} & \Gen $\uparrow$ & \Val $\uparrow$ & \ValH $\uparrow$ & \ED $\downarrow$ & \EDH  \\
            \midrule
            \textbf{\llamaS} & $96{{\scriptstyle\,(\,2)}}$ & \boldmath$6{{\scriptstyle\,(\,3)}}$ & \boldmath$48{{\scriptstyle\,(\,6)}}$ & $61{{\scriptstyle\,(\,5)}}$ & $58{{\scriptstyle\,(\,2)}}$ \\
            \textbf{\llamaM} & $100{{\scriptstyle\,(\,0)}}$ & \boldmath$16{{\scriptstyle\,(\,6)}}$ & \boldmath$84{{\scriptstyle\,(\,6)}}$ & $52{{\scriptstyle\,(\,3)}}$ & $57{{\scriptstyle\,(\,2)}}$ \\
            \textbf{\mistralS} & $100{{\scriptstyle\,(\,0)}}$ & \boldmath$8{{\scriptstyle\,(\,3)}}$ & \boldmath$30{{\scriptstyle\,(\,6)}}$ & $57{{\scriptstyle\,(\,4)}}$ & $57{{\scriptstyle\,(\,2)}}$ \\
            \textbf{\mistralM} & $100{{\scriptstyle\,(\,0)}}$ & \boldmath$13{{\scriptstyle\,(\,4)}}$ & \boldmath$87{{\scriptstyle\,(\,4)}}$ & $57{{\scriptstyle\,(\,4)}}$ & $58{{\scriptstyle\,(\,1)}}$ \\
            \textbf{\gemmaS} & $15{{\scriptstyle\,(\,6)}}$ & \boldmath$9{{\scriptstyle\,(\,6)}}$ & \boldmath$65{{\scriptstyle\,(\,20)}}$ & $62{{\scriptstyle\,(\,11)}}$ & $73{{\scriptstyle\,(\,5)}}$ \\
            \textbf{\gemmaM} & $98{{\scriptstyle\,(\,2)}}$ & \boldmath$5{{\scriptstyle\,(\,3)}}$ & \boldmath$85{{\scriptstyle\,(\,4)}}$ & $59{{\scriptstyle\,(\,4)}}$ & $58{{\scriptstyle\,(\,1)}}$ \\
            \textbf{\rd} & $100{{\scriptstyle\,(\,0)}}$ & \boldmath$14{{\scriptstyle\,(\,4)}}$ & \boldmath$50{{\scriptstyle\,(\,6)}}$ & $63{{\scriptstyle\,(\,4)}}$ & $67{{\scriptstyle\,(\,3)}}$ \\
            \bottomrule
        \end{tabular}
        }
        \caption{GSM8K}
    \end{subtable}
    \hfill
    \begin{subtable}[t]{0.48\textwidth}
        \centering
        \resizebox{0.98\textwidth}{!}{ 
        \begin{tabular}{l c c c c c} 
            \toprule
            {} & \Gen $\uparrow$ & \Val $\uparrow$ & \ValH $\uparrow$ & \ED $\downarrow$ & \EDH  \\
            \midrule
            \textbf{\llamaS} & $97{{\scriptstyle\,(\,1)}}$ & \boldmath$58{{\scriptstyle\,(\,4)}}$ & \boldmath$47{{\scriptstyle\,(\,4)}}$ & $73{{\scriptstyle\,(\,1)}}$ & $73{{\scriptstyle\,(\,1)}}$ \\
            \textbf{\llamaM} & $100{{\scriptstyle\,(\,0)}}$ & \boldmath$87{{\scriptstyle\,(\,2)}}$ & \boldmath$99{{\scriptstyle\,(\,1)}}$ & $71{{\scriptstyle\,(\,1)}}$ & $71{{\scriptstyle\,(\,1)}}$ \\
            \textbf{\mistralS} & $100{{\scriptstyle\,(\,0)}}$ & \boldmath$58{{\scriptstyle\,(\,4)}}$ & \boldmath$85{{\scriptstyle\,(\,3)}}$ & $74{{\scriptstyle\,(\,1)}}$ & $74{{\scriptstyle\,(\,1)}}$ \\
            \textbf{\mistralM} & $100{{\scriptstyle\,(\,0)}}$ & \boldmath$85{{\scriptstyle\,(\,3)}}$ & \boldmath$99{{\scriptstyle\,(\,1)}}$ & $77{{\scriptstyle\,(\,1)}}$ & $77{{\scriptstyle\,(\,1)}}$ \\
            \textbf{\gemmaS} & $99{{\scriptstyle\,(\,1)}}$ & \boldmath$80{{\scriptstyle\,(\,3)}}$ & \boldmath$87{{\scriptstyle\,(\,2)}}$ & $78{{\scriptstyle\,(\,1)}}$ & $78{{\scriptstyle\,(\,1)}}$ \\
            \textbf{\gemmaM} & $100{{\scriptstyle\,(\,0)}}$ & \boldmath$72{{\scriptstyle\,(\,3)}}$ & \boldmath$93{{\scriptstyle\,(\,2)}}$ & $76{{\scriptstyle\,(\,1)}}$ & $76{{\scriptstyle\,(\,1)}}$ \\
            \textbf{\rd} & $100{{\scriptstyle\,(\,0)}}$ & $81{{\scriptstyle\,(\,3)}}$ & $85{{\scriptstyle\,(\,2)}}$ & \boldmath$78{{\scriptstyle\,(\,1)}}$ & \boldmath$77{{\scriptstyle\,(\,1)}}$ \\
            \bottomrule
        \end{tabular}
        }
        \caption{MGNLI}
    \end{subtable}
    \caption{Performance of LLMs in generating \SCEs under unconstrained prompting at $T=0$, measured in terms of percentage of times the models are able to generate a \SCE (\Gen), percentage of times the model predictions on \SCEs yield the target label (\Val), and the normalized edit distance (\ED) between the original inputs and \SCEs. \textit{\ED is only reported for valid \SCEs}. \ValH and \EDH denote the metric values when the instructions for prediction on the original input and the \SCE generation are provided in the context while computing the validity of the \SCE (Section~\ref{sec:ce_eval}). Values in parentheses indicate marginal confidence intervals. See \autoref{sec:stat} for details. Values are bolded when the differences in with and without context conditions (\eg, \Val and \ValH) are statistically significant. Statistical significance is assessed using permutation tests (see \autoref{sec:permtest}). $\uparrow$ means higher values are better.}
    \label{table:direct_prompt_temp0}
\end{table*}

\begin{table*}[!htb]
    \centering
    \begin{subtable}[t]{0.48\textwidth}
        \centering
        \resizebox{0.98\textwidth}{!}{ 
        \begin{tabular}{l c c c c c} 
            \toprule
            {} & \Gen $\uparrow$ & \Val $\uparrow$ & \ValH $\uparrow$ & \ED $\downarrow$ & \EDH  \\
            \midrule
            \textbf{\llamaS} & $91{{\scriptstyle\,(\,7)}}$ & \boldmath$44{{\scriptstyle\,(\,12)}}$ & \boldmath$92{{\scriptstyle\,(\,7)}}$ & $34{{\scriptstyle\,(\,9)}}$ & $32{{\scriptstyle\,(\,6)}}$ \\
            \textbf{\llamaM} & $100{{\scriptstyle\,(\,0)}}$ & $53{{\scriptstyle\,(\,12)}}$ & $53{{\scriptstyle\,(\,12)}}$ & $19{{\scriptstyle\,(\,5)}}$ & $18{{\scriptstyle\,(\,6)}}$ \\
            \textbf{\mistralS} & $100{{\scriptstyle\,(\,0)}}$ & \boldmath$87{{\scriptstyle\,(\,8)}}$ & \boldmath$27{{\scriptstyle\,(\,10)}}$ & $36{{\scriptstyle\,(\,3)}}$ & $30{{\scriptstyle\,(\,7)}}$ \\
            \textbf{\mistralM} & $100{{\scriptstyle\,(\,0)}}$ & \boldmath$69{{\scriptstyle\,(\,11)}}$ & \boldmath$46{{\scriptstyle\,(\,5)}}$ & \boldmath$13{{\scriptstyle\,(\,3)}}$ & \boldmath$7{{\scriptstyle\,(\,2)}}$ \\
            \textbf{\gemmaS} & $0{{\scriptstyle\,(\,0)}}$ & $0{{\scriptstyle\,(\,0)}}$ & $0{{\scriptstyle\,(\,0)}}$ & $0{{\scriptstyle\,(\,0)}}$ & $0{{\scriptstyle\,(\,0)}}$ \\
            \textbf{\gemmaM} & $88{{\scriptstyle\,(\,9)}}$ & \boldmath$41{{\scriptstyle\,(\,14)}}$ & \boldmath$96{{\scriptstyle\,(\,6)}}$ & $19{{\scriptstyle\,(\,3)}}$ & $17{{\scriptstyle\,(\,3)}}$ \\
            \textbf{\rd} & $100{{\scriptstyle\,(\,0)}}$ & \boldmath$53{{\scriptstyle\,(\,12)}}$ & \boldmath$90{{\scriptstyle\,(\,7)}}$ & $23{{\scriptstyle\,(\,3)}}$ & $24{{\scriptstyle\,(\,3)}}$ \\
            \bottomrule
        \end{tabular}
        }
        \caption{DiscrimEval}
    \end{subtable}
    \hfill
    \begin{subtable}[t]{0.48\textwidth}
        \centering
        \resizebox{0.98\textwidth}{!}{ 
        \begin{tabular}{l c c c c c} 
            \toprule
            {} & \Gen $\uparrow$ & \Val $\uparrow$ & \ValH $\uparrow$ & \ED $\downarrow$ & \EDH  \\
            \midrule
            \textbf{\llamaS} & $67{{\scriptstyle\,(\,3)}}$ & \boldmath$72{{\scriptstyle\,(\,5)}}$ & \boldmath$88{{\scriptstyle\,(\,4)}}$ & $45{{\scriptstyle\,(\,3)}}$ & $48{{\scriptstyle\,(\,3)}}$ \\
            \textbf{\llamaM} & $99{{\scriptstyle\,(\,1)}}$ & \boldmath$36{{\scriptstyle\,(\,4)}}$ & \boldmath$74{{\scriptstyle\,(\,4)}}$ & $32{{\scriptstyle\,(\,0)}}$ & $33{{\scriptstyle\,(\,0)}}$ \\
            \textbf{\mistralS} & $26{{\scriptstyle\,(\,4)}}$ & $98{{\scriptstyle\,(\,2)}}$ & $92{{\scriptstyle\,(\,5)}}$ & $31{{\scriptstyle\,(\,2)}}$ & $29{{\scriptstyle\,(\,2)}}$ \\
            \textbf{\mistralM} & $96{{\scriptstyle\,(\,2)}}$ & \boldmath$50{{\scriptstyle\,(\,4)}}$ & \boldmath$100{{\scriptstyle\,(\,0)}}$ & $32{{\scriptstyle\,(\,0)}}$ & $32{{\scriptstyle\,(\,0)}}$ \\
            \textbf{\gemmaS} & $0{{\scriptstyle\,(\,0)}}$ & $0{{\scriptstyle\,(\,0)}}$ & $0{{\scriptstyle\,(\,0)}}$ & $0{{\scriptstyle\,(\,0)}}$ & $0{{\scriptstyle\,(\,0)}}$ \\
            \textbf{\gemmaM} & $18{{\scriptstyle\,(\,3)}}$ & \boldmath$62{{\scriptstyle\,(\,10)}}$ & \boldmath$98{{\scriptstyle\,(\,3)}}$ & $33{{\scriptstyle\,(\,1)}}$ & $32{{\scriptstyle\,(\,1)}}$ \\
            \textbf{\rd} & $25{{\scriptstyle\,(\,4)}}$ & \boldmath$57{{\scriptstyle\,(\,9)}}$ & \boldmath$89{{\scriptstyle\,(\,6)}}$ & $47{{\scriptstyle\,(\,3)}}$ & $44{{\scriptstyle\,(\,3)}}$ \\
            \bottomrule
        \end{tabular}
        }
        \caption{FolkTexts}
    \end{subtable}
    \hfill
    \begin{subtable}[t]{0.48\textwidth}
        \centering
        \resizebox{0.98\textwidth}{!}{ 
        \begin{tabular}{l c c c c c} 
            \toprule
            {} & \Gen $\uparrow$ & \Val $\uparrow$ & \ValH $\uparrow$ & \ED $\downarrow$ & \EDH  \\
            \midrule
            \textbf{\llamaS} & $88{{\scriptstyle\,(\,2)}}$ & \boldmath$75{{\scriptstyle\,(\,3)}}$ & \boldmath$83{{\scriptstyle\,(\,3)}}$ & \boldmath$57{{\scriptstyle\,(\,2)}}$ & \boldmath$52{{\scriptstyle\,(\,2)}}$ \\
            \textbf{\llamaM} & $100{{\scriptstyle\,(\,0)}}$ & \boldmath$87{{\scriptstyle\,(\,2)}}$ & \boldmath$66{{\scriptstyle\,(\,3)}}$ & \boldmath$57{{\scriptstyle\,(\,2)}}$ & \boldmath$53{{\scriptstyle\,(\,2)}}$ \\
            \textbf{\mistralS} & $100{{\scriptstyle\,(\,0)}}$ & $89{{\scriptstyle\,(\,10)}}$ & $88{{\scriptstyle\,(\,11)}}$ & $74{{\scriptstyle\,(\,5)}}$ & $74{{\scriptstyle\,(\,3)}}$ \\
            \textbf{\mistralM} & $100{{\scriptstyle\,(\,0)}}$ & \boldmath$79{{\scriptstyle\,(\,3)}}$ & \boldmath$86{{\scriptstyle\,(\,2)}}$ & $62{{\scriptstyle\,(\,1)}}$ & $63{{\scriptstyle\,(\,1)}}$ \\
            \textbf{\gemmaS} & $98{{\scriptstyle\,(\,1)}}$ & \boldmath$79{{\scriptstyle\,(\,3)}}$ & \boldmath$97{{\scriptstyle\,(\,1)}}$ & $50{{\scriptstyle\,(\,1)}}$ & $49{{\scriptstyle\,(\,1)}}$ \\
            \textbf{\gemmaM} & $100{{\scriptstyle\,(\,0)}}$ & \boldmath$86{{\scriptstyle\,(\,2)}}$ & \boldmath$97{{\scriptstyle\,(\,1)}}$ & $48{{\scriptstyle\,(\,1)}}$ & $47{{\scriptstyle\,(\,1)}}$ \\
            \textbf{\rd} & $99{{\scriptstyle\,(\,1)}}$ & $69{{\scriptstyle\,(\,3)}}$ & $72{{\scriptstyle\,(\,3)}}$ & $49{{\scriptstyle\,(\,1)}}$ & $48{{\scriptstyle\,(\,1)}}$ \\
            \bottomrule
        \end{tabular}
        }
        \caption{Twitter Financial News}
    \end{subtable}
    \hfill
    \begin{subtable}[t]{0.48\textwidth}
        \centering      
        \resizebox{0.98\textwidth}{!}{ 
        \begin{tabular}{l c c c c c} 
            \toprule
            {} & \Gen $\uparrow$ & \Val $\uparrow$ & \ValH $\uparrow$ & \ED $\downarrow$ & \EDH  \\
            \midrule
            \textbf{\llamaS} & $92{{\scriptstyle\,(\,2)}}$ & \boldmath$52{{\scriptstyle\,(\,5)}}$ & \boldmath$63{{\scriptstyle\,(\,4)}}$ & $69{{\scriptstyle\,(\,2)}}$ & $67{{\scriptstyle\,(\,2)}}$ \\
            \textbf{\llamaM} & $99{{\scriptstyle\,(\,1)}}$ & \boldmath$86{{\scriptstyle\,(\,3)}}$ & \boldmath$67{{\scriptstyle\,(\,4)}}$ & $79{{\scriptstyle\,(\,2)}}$ & $81{{\scriptstyle\,(\,2)}}$ \\
            \textbf{\mistralS} & $82{{\scriptstyle\,(\,3)}}$ & $92{{\scriptstyle\,(\,3)}}$ & $89{{\scriptstyle\,(\,3)}}$ & $77{{\scriptstyle\,(\,1)}}$ & $77{{\scriptstyle\,(\,1)}}$ \\
            \textbf{\mistralM} & $100{{\scriptstyle\,(\,0)}}$ & \boldmath$88{{\scriptstyle\,(\,3)}}$ & \boldmath$99{{\scriptstyle\,(\,1)}}$ & $66{{\scriptstyle\,(\,2)}}$ & $66{{\scriptstyle\,(\,2)}}$ \\
            \textbf{\gemmaS} & $96{{\scriptstyle\,(\,2)}}$ & \boldmath$73{{\scriptstyle\,(\,5)}}$ & \boldmath$98{{\scriptstyle\,(\,1)}}$ & $66{{\scriptstyle\,(\,2)}}$ & $64{{\scriptstyle\,(\,2)}}$ \\
            \textbf{\gemmaM} & $100{{\scriptstyle\,(\,0)}}$ & \boldmath$82{{\scriptstyle\,(\,4)}}$ & \boldmath$97{{\scriptstyle\,(\,1)}}$ & $66{{\scriptstyle\,(\,2)}}$ & $64{{\scriptstyle\,(\,2)}}$ \\
            \textbf{\rd} & $99{{\scriptstyle\,(\,1)}}$ & \boldmath$74{{\scriptstyle\,(\,4)}}$ & \boldmath$58{{\scriptstyle\,(\,4)}}$ & \boldmath$62{{\scriptstyle\,(\,2)}}$ & \boldmath$55{{\scriptstyle\,(\,2)}}$ \\
            \bottomrule
        \end{tabular}
        }
        \caption{SST2}
    \end{subtable}
    \hfill
    \begin{subtable}[t]{0.48\textwidth}
        \centering
        \resizebox{0.98\textwidth}{!}{ 
        \begin{tabular}{l c c c c c} 
            \toprule
            {} & \Gen $\uparrow$ & \Val $\uparrow$ & \ValH $\uparrow$ & \ED $\downarrow$ & \EDH  \\
            \midrule
            \textbf{\llamaS} & $96{{\scriptstyle\,(\,2)}}$ & $1{{\scriptstyle\,(\,1)}}$ & $2{{\scriptstyle\,(\,2)}}$ & $70{{\scriptstyle\,(\,17)}}$ & $62{{\scriptstyle\,(\,7)}}$ \\
            \textbf{\llamaM} & $100{{\scriptstyle\,(\,1)}}$ & \boldmath$25{{\scriptstyle\,(\,5)}}$ & \boldmath$64{{\scriptstyle\,(\,6)}}$ & $65{{\scriptstyle\,(\,3)}}$ & $63{{\scriptstyle\,(\,2)}}$ \\
            \textbf{\mistralS} & $100{{\scriptstyle\,(\,0)}}$ & \boldmath$46{{\scriptstyle\,(\,6)}}$ & \boldmath$2{{\scriptstyle\,(\,2)}}$ & $58{{\scriptstyle\,(\,2)}}$ & $65{{\scriptstyle\,(\,15)}}$ \\
            \textbf{\mistralM} & $100{{\scriptstyle\,(\,0)}}$ & \boldmath$14{{\scriptstyle\,(\,4)}}$ & \boldmath$92{{\scriptstyle\,(\,3)}}$ & $46{{\scriptstyle\,(\,2)}}$ & $47{{\scriptstyle\,(\,1)}}$ \\
            \textbf{\gemmaS} & $16{{\scriptstyle\,(\,5)}}$ & \boldmath$13{{\scriptstyle\,(\,11)}}$ & \boldmath$62{{\scriptstyle\,(\,15)}}$ & $51{{\scriptstyle\,(\,6)}}$ & $52{{\scriptstyle\,(\,4)}}$ \\
            \textbf{\gemmaM} & $97{{\scriptstyle\,(\,3)}}$ & \boldmath$9{{\scriptstyle\,(\,4)}}$ & \boldmath$74{{\scriptstyle\,(\,7)}}$ & $59{{\scriptstyle\,(\,4)}}$ & $58{{\scriptstyle\,(\,2)}}$ \\
            \textbf{\rd} & $100{{\scriptstyle\,(\,1)}}$ & \boldmath$8{{\scriptstyle\,(\,3)}}$ & \boldmath$28{{\scriptstyle\,(\,4)}}$ & $60{{\scriptstyle\,(\,7)}}$ & $64{{\scriptstyle\,(\,6)}}$ \\
            \bottomrule
        \end{tabular}
        }
        \caption{GSM8K}
    \end{subtable}
    \hfill
    \begin{subtable}[t]{0.48\textwidth}
        \centering
        \resizebox{0.98\textwidth}{!}{ 
        \begin{tabular}{l c c c c c} 
            \toprule
            {} & \Gen $\uparrow$ & \Val $\uparrow$ & \ValH $\uparrow$ & \ED $\downarrow$ & \EDH  \\
            \midrule
            \textbf{\llamaS} & $97{{\scriptstyle\,(\,1)}}$ & \boldmath$58{{\scriptstyle\,(\,4)}}$ & \boldmath$66{{\scriptstyle\,(\,3)}}$ & $76{{\scriptstyle\,(\,1)}}$ & $75{{\scriptstyle\,(\,1)}}$ \\
            \textbf{\llamaM} & $100{{\scriptstyle\,(\,0)}}$ & \boldmath$92{{\scriptstyle\,(\,2)}}$ & \boldmath$56{{\scriptstyle\,(\,2)}}$ & $77{{\scriptstyle\,(\,1)}}$ & $76{{\scriptstyle\,(\,1)}}$ \\
            \textbf{\mistralS} & $97{{\scriptstyle\,(\,1)}}$ & \boldmath$87{{\scriptstyle\,(\,2)}}$ & \boldmath$32{{\scriptstyle\,(\,3)}}$ & \boldmath$72{{\scriptstyle\,(\,1)}}$ & \boldmath$71{{\scriptstyle\,(\,1)}}$ \\
            \textbf{\mistralM} & $100{{\scriptstyle\,(\,0)}}$ & \boldmath$67{{\scriptstyle\,(\,3)}}$ & \boldmath$55{{\scriptstyle\,(\,2)}}$ & $76{{\scriptstyle\,(\,1)}}$ & $75{{\scriptstyle\,(\,1)}}$ \\
            \textbf{\gemmaS} & $99{{\scriptstyle\,(\,1)}}$ & \boldmath$68{{\scriptstyle\,(\,3)}}$ & \boldmath$90{{\scriptstyle\,(\,2)}}$ & $77{{\scriptstyle\,(\,1)}}$ & $77{{\scriptstyle\,(\,1)}}$ \\
            \textbf{\gemmaM} & $100{{\scriptstyle\,(\,0)}}$ & \boldmath$70{{\scriptstyle\,(\,3)}}$ & \boldmath$92{{\scriptstyle\,(\,2)}}$ & $75{{\scriptstyle\,(\,1)}}$ & $75{{\scriptstyle\,(\,1)}}$ \\
            \textbf{\rd} & $100{{\scriptstyle\,(\,0)}}$ & \boldmath$67{{\scriptstyle\,(\,3)}}$ & \boldmath$89{{\scriptstyle\,(\,2)}}$ & $73{{\scriptstyle\,(\,1)}}$ & $72{{\scriptstyle\,(\,1)}}$ \\
            \bottomrule
        \end{tabular}
        }
        \caption{MGNLI}
    \end{subtable}
    \caption{
    Performance of LLMs in generating \SCEs under rationale-based prompting at $T=0$. For details of metric names, see the caption of \autoref{table:direct_prompt_temp0}.
    }
    \label{table:rationale_prompting_temp0}
\end{table*}

Tables \ref{table:direct_prompt_temp0} and \ref{table:rationale_prompting_temp0} show the results when using unconstrained prompting and rationale-based prompting, respectively, at $T = 0$. 
Results for all other configurations like non-zero temperatures and CoT prompting (Tables~\ref{table:direct_prompting_temp05}, \ref{table:rationale_prompting_temp05}, \ref{table:COT_prompting_temp0} and \ref{table:COT_prompting_temp05}) are shown in \autoref{app:additional_results} and discussed under each RQ. All tables show confidence intervals computed using standard error of the mean (\autoref{sec:stat}).

\subsection*{RQ1: Ability of LLMs to generate SCEs}

\textit{Most models successfully generate \SCEs in the vast majority of cases}, with the notable exception of the \gemmaS model on the \textsc{DiscrimEval} and \textsc{FolkTexts} datasets. However, CoT prompting massively improves \SCE generation ability of \gemmaS (\autoref{table:COT_prompting_temp0}).
Most models, including \gemmaS, exhibit enhanced \SCE generation at $T=0.5$.
The fraction roughly remains the same for rationale-based prompting, as shown in Tables \ref{table:rationale_prompting_temp0} and \ref{table:rationale_prompting_temp05}. 

\subsection*{RQ2: Do \SCEs yield the target label?}
\textit{\SCEs yield the target label in most cases, however, there are large variations.} 
The most prominent variation is along the \textit{task level}. For the \textsc{GSM8K} dataset, which involves more complex mathematical reasoning, valid \SCE generation rates remain under $20\%$ in a vast majority of cases. Similarly, for the \textsc{FolkTexts} tasks which require the model to reason through the Census-gathered data, the validity in many cases is low.

We also see  a mixed trend at \textit{model-size} level. The smaller models, \gemmaS ($9$B parameters), \llamaS ($8$B), and \mistralS ($7$B), sometimes tend to generate valid \SCEs at a lower rate than larger counterparts. However, the trend is reversed in some other cases, \eg, with unconstrained prompting on \textsc{FolkTexts}, \mistralS outperforms its larger counterpart.
The reasoning model \rd (32B) also does not consistently outperform comparably sized models such as \gemmaM and \mistralM.

\textit{Presence of the original prediction and counterfactual generation in the context window has a large impact on validity} as shown by the comparison of \Val and \ValH in Tables \ref{table:direct_prompt_temp0} and \ref{table:rationale_prompting_temp0}. Most prominently, on the \textsc{GSM8K} dataset, validity increases significantly, indicating that the \textbf{model's mathematical reasoning ability is influenced by information that should be irrelevant}. We observe a similar trend in the \textsc{FolkTexts} dataset. The trend, however, is not universal. In other datasets, models such as \llamaS and \llamaM exhibit a decrease in validity when additional contextual information is included.

\textit{Rationale-based prompting has a diverse impact on \SCE validity} as shown by comparing Tables \ref{table:direct_prompt_temp0} and \ref{table:rationale_prompting_temp0}. In some cases, such as \llamaM on \textsc{DiscrimEval}, the fraction of \SCEs deemed valid by the model drops sharply from $94\%$ to $53\%$. In contrast, for \llamaS on \textsc{FolkTexts}, the validity rate increases substantially from $20\%$ to $72\%$ at a temperature of $0$.

\textit{CoT generally leads to modest improvements in \SCE validity.} For instance, at $T=0$, the average validity over all datasets and models is 
$69$\% with unconstrained prompting, 
$64\%$ with rationale-based prompting,
and $75$\% with CoT prompting.

\setlength{\tabcolsep}{3pt}        %
\renewcommand{\arraystretch}{1.2}  %

\begin{table*}[ht]
\small
\centering
\begin{tabular}{l
  c@{\hskip 4pt}c@{\hskip 10pt}  %
  c@{\hskip 4pt}c@{\hskip 10pt}  %
  c@{\hskip 4pt}c@{\hskip 10pt}  %
  c@{\hskip 4pt}c@{\hskip 10pt}  %
  c@{\hskip 4pt}c@{\hskip 10pt}  %
  c@{\hskip 4pt}c              %
}
\toprule
{} & \multicolumn{2}{c}{\textbf{DEV}} & \multicolumn{2}{c}{\textbf{TWT}} & \multicolumn{2}{c}{\textbf{SST}} & \multicolumn{2}{c}{\textbf{FLK}} & \multicolumn{2}{c}{\textbf{NLI}} & \multicolumn{2}{c}{\textbf{MTH}} \\
{} & w/ & w/o & w/ & w/o & w/ & w/o & w/ & w/o & w/ & w/o & w/ & w/o \\
\midrule
\llamaS & (1, 51) & (8, 62) & (32, 54) & (0, 14) & (5, 34) & (23, 50) & (0, 7) & (1, 34) & (1, 45) & (1, 31) & (15, 50) & (6, 66) \\
\llamaM & (65, 69) & (1, 55) & (1, 21) & (0, 17) & (7, 31) & (4, 43) & (100, 100) & (0, 1) & (1, 53) & (0, 11) & (100, 100) & (8, 34) \\
\mistralS & (11, 20) & (0, 15) & (2, 34) & (0, 18) & (4, 74) & (6, 48) & (6, 12) & (1, 4) & (5, 26) & (0, 12) & (2, 50) & (1, 31) \\
\mistralM & (100, 100) & (4, 32) & (0, 12) & (0, 8) & (2, 55) & (1, 28) & (0, 5) & (0, 1) & (8, 56) & (0, 14) & (7, 47) & (4, 33) \\
\gemmaS & (0, 0) & (0, 0) & (0, 19) & (0, 13) & (50, 85) & (1, 64) & (0, 0) & (0, 0) & (0, 18) & (1, 14) & (1, 37) & (1, 46) \\
\gemmaM & (100, 100) & (1, 20) & (1, 13) & (0, 9) & (43, 55) & (1, 25) & (100, 100) & (1, 6) & (0, 19) & (0, 9) & (1, 34) & (7, 49) \\
\rd  & (100, 100) & (1, 53) & (2, 72) & (8, 59) & (55, 81) & (3, 69) & (0, 1) & (0, 1) & (1, 26) & (5, 17) & (1, 32) & (24, 63) \\
\bottomrule
\end{tabular}
\caption{Normalized difference in lengths of valid and invalid counterfactuals. For DiscrimEval (\DSEV), Twitter Financial News (\TWTR), SST2 (\SST), FolkTexts (\FOLK), MGNLI (\NLI), and GSM8K (\MATH) datasets under unconstrained prompting with $T = 0$. Left columns (w/o) show the differences without prediction and counterfactual generations provided as context (Section~\ref{sec:ce_eval}), whereas right columns (w/) show the differences with this information. Reported confidence intervals are estimated via nonparametric bootstrap resampling ($10,000$ iterations). See \autoref{app:bootstrap_ci} for details.
}
\label{table:response_len_diff_direct_prompting_temp0_bootstrap_version}
\end{table*}

\subsection*{RQ3: Changes required to generate \SCEs}

\textit{For a given task and dataset, different LLMs require different amount of changes to generate \SCEs}, even for a similar level of validity. Consider for \gemmaM, \gemmaS and \rd models for \textsc{DiscrimEval} data. 

The required changes also depend on the task and dataset. For example, in \textsc{SST2}, where models achieve some of the highest validity scores, we observe the highest \ED.
This relationship between validity and edit distance, however, is not completely linear and also depends on the input length. In \textsc{DiscrimEval} and \textsc{FolkTexts}, where input lengths can span several hundred tokens, the models exhibit low \Val alongside relatively low \ED.
Temperature also influences average validity, which is higher at $T=0.5$ than at $T=0$ across all datasets and models in both unconstrained (\autoref{table:direct_prompting_temp05}) and rationale-based prompting (\autoref{table:rationale_prompting_temp05}). Finally, we notice that the presence of \textit{context mostly has no statistically significant impact} on the edit distance of valid \SCEs.\\
\textit{Rationale-based prompting does not consistently produce closer \SCEs}, as evident from the comparison between Tables~\ref{table:direct_prompt_temp0} and~\ref{table:rationale_prompting_temp0}. For instance, on the \textsc{SST2} dataset, \ED values are generally lower under rationale-based prompting, with the exception of \llamaM and \mistralS.\\
\xhdrnodot{Are invalid \SCEs statistically different?}\\
We investigate whether the lengths of \SCEs can provide a clue on their validity. Our question is inspired by previous work on detecting LLM hallucinations~\cite{hallucination_snowball,snyder_early_2024,azaria-mitchell-2023-internal} which shows that incorrect model outputs show statistically different patterns from correct answers.
For each model, datasest, and \SCE generation configuration, we compute the \textit{normalized difference in lengths} as
$
\frac{| L_{\text{val}} - L_{\text{inval}} |}{\max(L_{\text{val}}, L_{\text{inval}})} \times 100 
$
where $L_{\text{val}}$ is the average length of valid \SCEs.
This metric ranges from 0 to 100, with higher values reflecting greater length differences between valid and invalid \SCEs. As shown in \autoref{table:response_len_diff_direct_prompting_temp0_bootstrap_version}, context generally amplifies these differences, sometimes reaching the maximum of 100, where valid and invalid \SCEs diverge almost completely.

\section{Characterization of Failure Cases}
\label{sec:CharaFail}

We begin our failure case analysis with a human annotation study that evaluates the correctness of the generated \SCEs. To complement this, we employ targeted automatic metrics: the Flesch–Kincaid Readability score to measure linguistic complexity, cosine similarity in the embedding space to quantify semantic drift, and $K$-means clustering in the embedding space to identify potential task misunderstandings.

\xhdr{Human Annotation and Evaluation}
Our goal was to test if the \SCE validity correlates with its correctness.
To this end, for each model, we annotated \SCE correctness (that is, if the \SCE indeed evaluates to the target label) on $50$ randomly selected \textsc{GSM8K} samples. The annotation protocol is reported in \autoref{app:AnnotProt}. We report the correlation results as 
(coefficient, \(p\)-value), where \(r\) denotes Pearson correlation, \(\rho\) denotes Spearman correlation, and \(p\) is the associated two-tailed significance level.
Spearman shows statistically significant correlation between counterfactual validity and correctness in the \textit{without context condition}, that is, when the conversation history is not in the context ($\rho=0.76$, $p=0.05$). For Pearson correlation, the statistical significance is narrowly rejected ($r=0.74$, $p=0.056$).
In the \textit{with context condition}, there is no significant correlation between validity and correctness (Spearman $\rho = 0.52, p = 0.23$; Pearson $r = 0.57, p = 0.18$). This result seems to follow the intuition that regardless of the correctness of \SCE, the model might be looking up the target answers from the conversation history without actually solving it.

\xhdr{Readability Analysis via Flesch--Kincaid}
To evaluate linguistic complexity, we computed the Flesch–Kincaid readability score \citep{flesch2007flesch} for each \SCE.
We then compared scores across valid vs.\ invalid and correct vs.\ incorrect cases to examine whether easier-to-read counterfactuals are associated with higher validity or correctness.
Correlation analyses revealed no significant relationships between reading ease and
(i) correctness ($\rho=-0.59$, $r=-0.52$, $p=0.17,0.23$),
(ii) validity without context ($\rho=0.09$, $r=-0.06$, $p=0.86,0.90$),
(iii) validity with context ($\rho=-0.61$, $r=-0.62$, $p=0.15,0.14$).
This indicates that readability levels do not systematically differentiate between valid vs.\ invalid or correct vs.\ incorrect \SCEs.

\xhdr{Drift in Embedding Space}
Recent work \citep{azaria2023internal, snyder2024early, bhan2025did} shows that LLM hidden states can reveal problematic model behavior. Inspired by these works, we test whether hidden states of \SCEs drift from the original problem when the \SCE is invalid or incorrect, measuring drift via cosine distance between the embeddings of the problem and the \SCE:  
\[
\text{Drift} = 1 - \frac{\langle e_{\text{orig}}, e_{\text{SCE}} \rangle}{\|e_{\text{orig}}\|\;\|e_{\text{SCE}}\|}
\]
where $e_{\text{orig}}$ and $e_{\text{SCE}}$ denote the sentence-level mean embeddings of the original input and the \SCE, respectively \citep{bhan2025did}. We conduct this analysis on \textsc{GSM8K}, where correctness labels are available from annotation. We find no correlation between drift and \SCE correctness ($\rho=0.01$, $p=0.99$; $r=0.21$, $p=0.66$). For validity, drift shows no effect with context, but without context yields a significant Pearson correlation ($r=0.76$, $p=0.05$) and a non-significant Spearman correlation ($\rho=0.12$, $p=0.80$). 

\xhdr{Clustering \SCE representations}
Inspired by \cite{bhan2025did}, who analyze hidden representations of self-explanations, we tested whether the representations of valid and invalid \SCEs differ. 
We applied k-means clustering with $k=2$ to various \SCE\ representations (\eg, last and first generated token, last input token) to probe whether valid and invalid cases separate in the embedding space. If there were no difference in the representations of valid and invalid \SCEs, we would expect the two clusters to contain a similar number of valid and invalid \SCEs. 
\autoref{table:combined_strategies} reports the absolute differences between valid and invalid \SCEs in cluster~0 ($\Delta_{0}$) and cluster~1 ($\Delta_{1}$), highlighting consistent disparities in their internal representations. See~\autoref{app:FailAnal} for details.

\section{Why do models struggle with \SCEs?}
\label{sec:HEMP}
Counterfactual reasoning is an ability often taken for granted in humans~\cite{miller2019explanation,sep-knowledge-analysis}. Given their impressive performance on conceptually abstract tasks~\cite{bubeck2023sparks}, one would expect LLMs to also depict sound counterfactual reasoning abilities. Our investigations show otherwise.

Our hypothesis is that the inability of LLMs to generate valid \SCEs arises because their learning process and operation is very different from humans. 
While humans tend to understand the world through counterfactual reasoning \cite{miller2019explanation}, LLMs are fundamentally trained to predict the next token.
Even the most advanced LLMs that appear strong at reasoning still fundamentally rely on next-token prediction, enhanced by advanced techniques like reranking and CoT training~\cite{guo2025deepseek}, output pruning~\cite{dong2025domain}, or guided decoding~\cite{jiang2024technical}.
As a result, LLMs do not reason like humans and are not natural causal thinkers. 
Motivated by recent advances in model alignment (specifically, contrastive prompting \citep{liu2024direct}, which leverages paired prompts differing along a single axis), we posit that training LLMs with contrastive example pairs (\eg, correct vs. incorrect \SCEs in our case) could enhance their counterfactual reasoning capability.

We also believe that \textbf{side-effects of the attention mechanism} impact the model’s reasoning ability. 
This is supported by our findings in Section~\ref{sec:results}, RQ2. We observe that validity is higher when the original prediction and counterfactual generation are present in the context window (\ValH) compared to when they are removed (\Val). In particular, on the \textsc{GSM8K} dataset, the \SCE validity improves significantly in the presence of this information.
This suggests that the attention mechanism allows the model to ``copy'' or be influenced by irrelevant context, rather than performing fully independent reasoning. Thus, even subtle hints or artifacts in the input can enhance apparent performance, masking the true reasoning capabilities of the model.

\begin{figure}[ht]
    \centering
    \includegraphics[width=\linewidth]{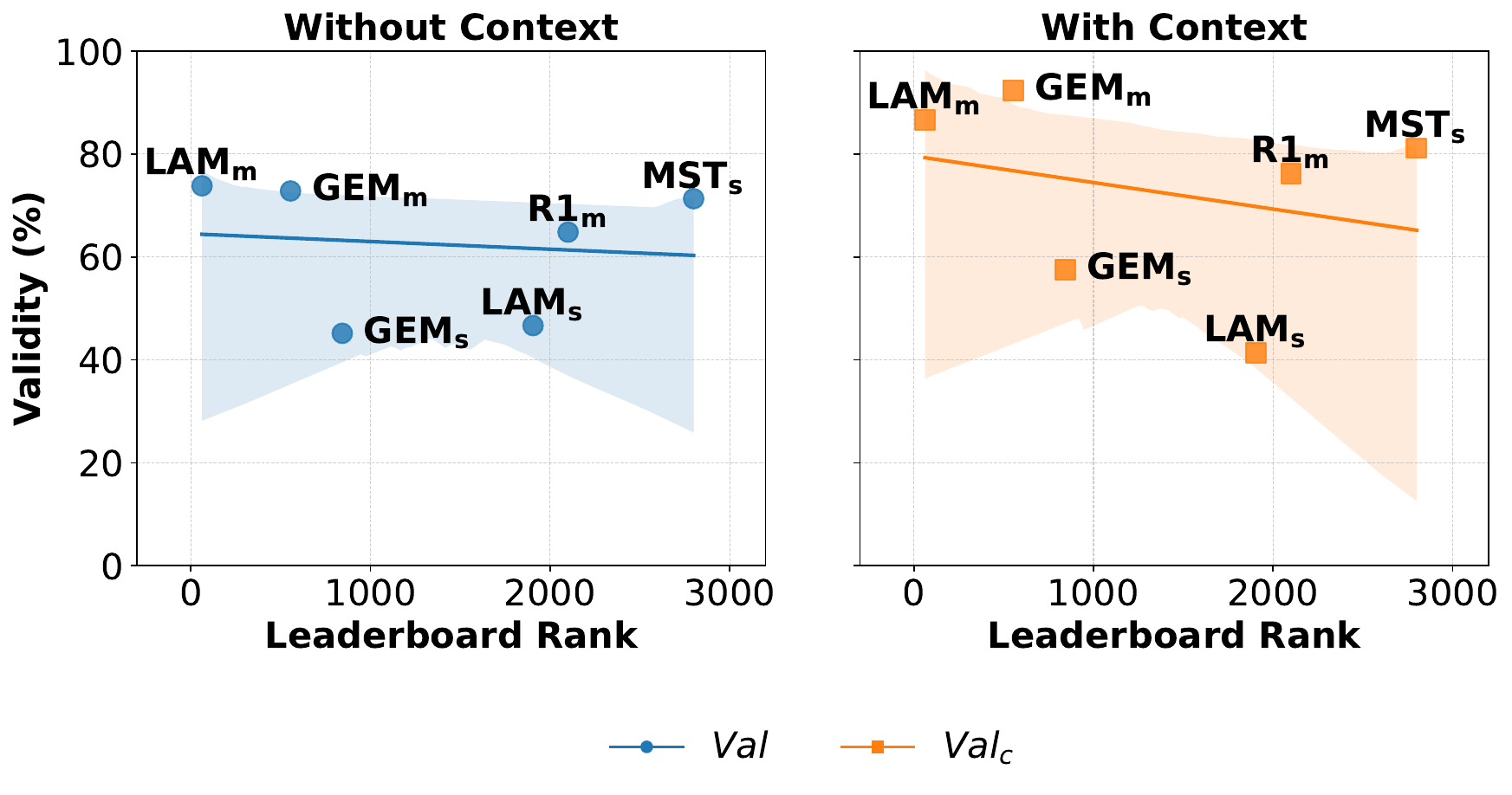}
    \caption{Relationship between leaderboard rank and \SCE validity. The left panel reports validity without context (\Val), and the right panel with context (\ValH). Lower ranks correspond to stronger leaderboard positions. Regression lines with 95\% confidence intervals are shown to indicate overall trends.}
    \label{fig:rank_vs_plp}
\end{figure}

Inspired by the work on emergent properties and neural scaling laws~\cite{brown_language_2020,kaplan2020scaling,wei2022emergent}, we investigate \textbf{whether counterfactual reasoning abilities emerge as models improve on well-established quality criteria}.
Specifically, we perform a correlation analysis between the validity percentage of \SCEs, and \textit{model size}, \textit{few-shot perplexity}, and \textit{open LLM leaderboard rank}.\footnote{Leaderboard ranks were retrieved on May~17,~2025.} Our results (\autoref{app:correlation}) reveal no strong or consistent correlations.
As shown in \autoref{fig:rank_vs_plp}, leaderboard rank does not consistently align with \SCE validity. In particular, models with weaker leaderboard positions (\eg, \mistralS and \rd) achieve comparable or even higher validity than stronger-ranked models (\eg, \llamaS and \gemmaS). Leaderboard rank alone fails to reflect a model’s counterfactual reasoning ability.

\section{Conclusion and future work}
In this study, we examined the ability of LLMs to produce self-generated counterfactual explanations (SCEs).
Our results show that LLMs consistently struggle with generating valid \SCEs. In many cases model prediction on a \SCE does not yield the same target prediction for which the model crafted the \SCE.
Surprisingly, we find that LLMs put significant emphasis on the context, as the prediction on \SCE is significantly impacted by the presence of the original prediction and the instructions for generating the \SCE.
Based on this empirical evidence, we argue that LLMs are still far from being able to explain their own predictions counterfactually.
Our findings add to similar insights from recent studies~\cite{lanham2023measuring,tanneru2024quantifying,madsen2024self}.
Our work opens several avenues for future work. Inspired by counterfactual data augmentation~\cite{sachdeva2023catfood}, one could include the counterfactual explanation capabilities as a part of the LLM training process. This inclusion may enhance the counterfactual reasoning capabilities of the LLM.

Finally, our experiments were limited to relatively simple tasks: classification and mathematics problems where the solution is an integer. This limitation was mainly due to the fact that it is difficult to automatically judge validity of answers for more open-ended language generation tasks like search and information retrieval. Scaling our analysis to such tasks would require significant human-annotation resources, and is an important direction for future investigations.
\\
\\

\section{Limitations}
Our work has several limitations. First, explainability and privacy can sometimes be at odds with each other. Even if LLMs are able to provide comprehensive and faithful explanations, this can introduce privacy and security concerns \citep{pawlicki2024explainability, grant2020show}. Detailed explanations may inadvertently expose sensitive information or be exploited for adversarial attacks on the model itself. However, our work focuses on publicly available models and datasets, ensuring that these risks are mitigated. 

Similarly, savvy users can strategically use counterfactual explanations to unfairly maximize their chances of receiving positive outcomes~\cite{tsirtsis2020decisions}. Detecting and limiting this behavior would be an important desideratum before LLM-generated counterfactual explanations are integrated into real-world decision-making systems.

Our analyses in this paper primarily relied on automated metrics to evaluate the quality of \SCEs. Although we conducted a small-scale human annotation for one task (Section~\ref{sec:CharaFail}), we did not extend this to other tasks. Comprehensive human evaluation remains important for assessing the plausibility of explanations, and future studies could incorporate such feedback to improve model performance, for example through direct preference optimization \citep{rafailov2024direct}.

\bibliography{refs}

\begin{thebibliography}{77}
\providecommand{\natexlab}[1]{#1}

\bibitem[{Agarwal et~al.(2024)Agarwal, Tanneru, and
  Lakkaraju}]{agarwal2024faithfulness}
Chirag Agarwal, Sree~Harsha Tanneru, and Himabindu Lakkaraju. 2024.
\newblock Faithfulness vs. plausibility: On the (un) reliability of
  explanations from large language models.
\newblock \emph{arXiv preprint arXiv:2402.04614}.

\bibitem[{Arvanitidis et~al.(2016)Arvanitidis, Hansen, and
  Hauberg}]{arvanitidis2016locally}
Georgios Arvanitidis, Lars~K Hansen, and S{\o}ren Hauberg. 2016.
\newblock A locally adaptive normal distribution.
\newblock \emph{Advances in Neural Information Processing Systems}, 29.

\bibitem[{Azaria and
  Mitchell(2023{\natexlab{a}})}]{azaria-mitchell-2023-internal}
Amos Azaria and Tom Mitchell. 2023{\natexlab{a}}.
\newblock \href {https://doi.org/10.18653/v1/2023.findings-emnlp.68} {The
  internal state of an {LLM} knows when it`s lying}.
\newblock In \emph{Findings of the Association for Computational Linguistics:
  EMNLP 2023}, pages 967--976, Singapore. Association for Computational
  Linguistics.

\bibitem[{Azaria and Mitchell(2023{\natexlab{b}})}]{azaria2023internal}
Amos Azaria and Tom Mitchell. 2023{\natexlab{b}}.
\newblock The internal state of an llm knows when it's lying.
\newblock \emph{arXiv preprint arXiv:2304.13734}.

\bibitem[{Bhan et~al.(2025)Bhan, Vittaut, Chesneau, Chandar, and
  Lesot}]{bhan2025did}
Milan Bhan, Jean-Noel Vittaut, Nicolas Chesneau, Sarath Chandar, and
  Marie-Jeanne Lesot. 2025.
\newblock Did i faithfully say what i thought? bridging the gap between neural
  activity and self-explanations in large language models.
\newblock \emph{arXiv preprint arXiv:2506.09277}.

\bibitem[{Bhattacharjee et~al.(2024)Bhattacharjee, Moraffah, Garland, and
  Liu}]{bhattacharjee2024towards}
Amrita Bhattacharjee, Raha Moraffah, Joshua Garland, and Huan Liu. 2024.
\newblock Towards llm-guided causal explainability for black-box text
  classifiers.
\newblock In \emph{AAAI 2024 Workshop on Responsible Language Models,
  Vancouver, BC, Canada}.

\bibitem[{Biderman et~al.(2024)Biderman, Schoelkopf, Sutawika, Gao, Tow,
  Abbasi, Aji, Ammanamanchi, Black, Clive et~al.}]{biderman2024lessons}
Stella Biderman, Hailey Schoelkopf, Lintang Sutawika, Leo Gao, Jonathan Tow,
  Baber Abbasi, Alham~Fikri Aji, Pawan~Sasanka Ammanamanchi, Sidney Black,
  Jordan Clive, et~al. 2024.
\newblock Lessons from the trenches on reproducible evaluation of language
  models.
\newblock \emph{arXiv preprint arXiv:2405.14782}.

\bibitem[{Bommasani et~al.(2021)Bommasani, Hudson, Adeli, Altman, Arora, von
  Arx, Bernstein, Bohg, Bosselut, Brunskill
  et~al.}]{bommasani2021opportunities}
Rishi Bommasani, Drew~A Hudson, Ehsan Adeli, Russ Altman, Simran Arora, Sydney
  von Arx, Michael~S Bernstein, Jeannette Bohg, Antoine Bosselut, Emma
  Brunskill, et~al. 2021.
\newblock On the opportunities and risks of foundation models.
\newblock \emph{arXiv preprint arXiv:2108.07258}.

\bibitem[{Bricken et~al.(2023)Bricken, Templeton, Batson, Chen, Jermyn,
  Conerly, Turner, Anil, Denison, Askell, Lasenby, Wu, Kravec, Schiefer,
  Maxwell, Joseph, Hatfield-Dodds, Tamkin, Nguyen, McLean, Burke, Hume, Carter,
  Henighan, and Olah}]{bricken2023monosemanticity}
Trenton Bricken, Adly Templeton, Joshua Batson, Brian Chen, Adam Jermyn, Tom
  Conerly, Nick Turner, Cem Anil, Carson Denison, Amanda Askell, Robert
  Lasenby, Yifan Wu, Shauna Kravec, Nicholas Schiefer, Tim Maxwell, Nicholas
  Joseph, Zac Hatfield-Dodds, Alex Tamkin, Karina Nguyen, Brayden McLean,
  Josiah~E Burke, Tristan Hume, Shan Carter, Tom Henighan, and Christopher
  Olah. 2023.
\newblock Towards monosemanticity: Decomposing language models with dictionary
  learning.
\newblock \emph{Transformer Circuits Thread}.
\newblock
  Https://transformer-circuits.pub/2023/monosemantic-features/index.html.

\bibitem[{Brown et~al.(2020)Brown, Mann, Ryder, Subbiah, Kaplan, Dhariwal,
  Neelakantan, Shyam, Sastry, Askell, Agarwal, Herbert-Voss, Krueger, Henighan,
  Child, Ramesh, Ziegler, Wu, Winter, Hesse, Chen, Sigler, Litwin, Gray, Chess,
  Clark, Berner, McCandlish, Radford, Sutskever, and
  Amodei}]{brown_language_2020}
Tom Brown, Benjamin Mann, Nick Ryder, Melanie Subbiah, Jared~D Kaplan, Prafulla
  Dhariwal, Arvind Neelakantan, Pranav Shyam, Girish Sastry, Amanda Askell,
  Sandhini Agarwal, Ariel Herbert-Voss, Gretchen Krueger, Tom Henighan, Rewon
  Child, Aditya Ramesh, Daniel Ziegler, Jeffrey Wu, Clemens Winter, Chris
  Hesse, Mark Chen, Eric Sigler, Mateusz Litwin, Scott Gray, Benjamin Chess,
  Jack Clark, Christopher Berner, Sam McCandlish, Alec Radford, Ilya Sutskever,
  and Dario Amodei. 2020.
\newblock \href
  {https://papers.nips.cc/paper/2020/hash/1457c0d6bfcb4967418bfb8ac142f64a-Abstract.html}
  {Language {Models} are {Few}-{Shot} {Learners}}.
\newblock In \emph{Advances in {Neural} {Information} {Processing} {Systems}},
  volume~33, pages 1877--1901. Curran Associates, Inc.

\bibitem[{Bubeck et~al.(2023)Bubeck, Chandrasekaran, Eldan, Gehrke, Horvitz,
  Kamar, Lee, Lee, Li, Lundberg et~al.}]{bubeck2023sparks}
S{\'e}bastien Bubeck, Varun Chandrasekaran, Ronen Eldan, Johannes Gehrke, Eric
  Horvitz, Ece Kamar, Peter Lee, Yin~Tat Lee, Yuanzhi Li, Scott Lundberg,
  et~al. 2023.
\newblock Sparks of artificial general intelligence: Early experiments with
  gpt-4.
\newblock \emph{arXiv preprint arXiv:2303.12712}.

\bibitem[{Chatzi et~al.(2025)Chatzi, Benz, Straitouri, Tsirtsis, and
  Rodriguez}]{chatzi2025ce}
Ivi Chatzi, Nina L~Corvelo Benz, Eleni Straitouri, Stratis Tsirtsis, and
  Manuel~Gomez Rodriguez. 2025.
\newblock Counterfactual token generation in large language models.
\newblock In \emph{Proceedings of the 4th Conference on Causal Learning and
  Reasoning}.

\bibitem[{Chen et~al.(2023)Chen, Zhong, Ri, Zhao, He, Steinhardt, Yu, and
  McKeown}]{chen2023models}
Yanda Chen, Ruiqi Zhong, Narutatsu Ri, Chen Zhao, He~He, Jacob Steinhardt, Zhou
  Yu, and Kathleen McKeown. 2023.
\newblock Do models explain themselves? counterfactual simulatability of
  natural language explanations.
\newblock \emph{arXiv preprint arXiv:2307.08678}.

\bibitem[{Cobbe et~al.(2021)Cobbe, Kosaraju, Bavarian, Chen, Jun, Kaiser,
  Plappert, Tworek, Hilton, Nakano, Hesse, and Schulman}]{cobbe2021gsm8k}
Karl Cobbe, Vineet Kosaraju, Mohammad Bavarian, Mark Chen, Heewoo Jun, Lukasz
  Kaiser, Matthias Plappert, Jerry Tworek, Jacob Hilton, Reiichiro Nakano,
  Christopher Hesse, and John Schulman. 2021.
\newblock Training verifiers to solve math word problems.
\newblock \emph{arXiv preprint arXiv:2110.14168}.

\bibitem[{Cohen-Wang et~al.(2025)Cohen-Wang, Shah, Georgiev, and
  Madry}]{cohen2025contextcite}
Benjamin Cohen-Wang, Harshay Shah, Kristian Georgiev, and Aleksander Madry.
  2025.
\newblock Contextcite: Attributing model generation to context.
\newblock \emph{Advances in Neural Information Processing Systems},
  37:95764--95807.

\bibitem[{Cruz et~al.(2024)Cruz, Hardt, and
  Mendler-D{\"u}nner}]{cruz2024evaluating}
Andr{\'e}~F Cruz, Moritz Hardt, and Celestine Mendler-D{\"u}nner. 2024.
\newblock Evaluating language models as risk scores.
\newblock \emph{arXiv preprint arXiv:2407.14614}.

\bibitem[{Delaney et~al.(2023)Delaney, Pakrashi, Greene, and
  Keane}]{delaney2023counterfactual}
Eoin Delaney, Arjun Pakrashi, Derek Greene, and Mark~T Keane. 2023.
\newblock Counterfactual explanations for misclassified images: How human and
  machine explanations differ.
\newblock \emph{Artificial Intelligence}, 324:103995.

\bibitem[{DeYoung et~al.(2019)DeYoung, Jain, Rajani, Lehman, Xiong, Socher, and
  Wallace}]{deyoung2019eraser}
Jay DeYoung, Sarthak Jain, Nazneen~Fatema Rajani, Eric Lehman, Caiming Xiong,
  Richard Socher, and Byron~C Wallace. 2019.
\newblock Eraser: A benchmark to evaluate rationalized nlp models.
\newblock \emph{arXiv preprint arXiv:1911.03429}.

\bibitem[{Dong et~al.(2025)Dong, Peng, Liu, Zhao, Wu, Xiao, and
  Wang}]{dong2025domain}
Zican Dong, Han Peng, Peiyu Liu, Wayne~Xin Zhao, Dong Wu, Feng Xiao, and
  Zhifeng Wang. 2025.
\newblock Domain-specific pruning of large mixture-of-experts models with
  few-shot demonstrations.
\newblock \emph{arXiv preprint arXiv:2504.06792}.

\bibitem[{Dubey et~al.(2024)Dubey, Jauhri, Pandey, Kadian, Al-Dahle, Letman,
  Mathur, Schelten, Yang, Fan et~al.}]{dubey2024llama}
Abhimanyu Dubey, Abhinav Jauhri, Abhinav Pandey, Abhishek Kadian, Ahmad
  Al-Dahle, Aiesha Letman, Akhil Mathur, Alan Schelten, Amy Yang, Angela Fan,
  et~al. 2024.
\newblock The llama 3 herd of models.
\newblock \emph{arXiv preprint arXiv:2407.21783}.

\bibitem[{Flesch(2007)}]{flesch2007flesch}
Rudolf Flesch. 2007.
\newblock Flesch-kincaid readability test.
\newblock \emph{Retrieved October}, 26(3):2007.

\bibitem[{Gat et~al.(2023)Gat, Calderon, Feder, Chapanin, Sharma, and
  Reichart}]{gat2023faithful}
Yair Gat, Nitay Calderon, Amir Feder, Alexander Chapanin, Amit Sharma, and Roi
  Reichart. 2023.
\newblock Faithful explanations of black-box nlp models using llm-generated
  counterfactuals.
\newblock \emph{arXiv preprint arXiv:2310.00603}.

\bibitem[{Gilpin et~al.(2018)Gilpin, Bau, Yuan, Bajwa, Specter, and
  Kagal}]{gilpin2018explaining}
Leilani~H Gilpin, David Bau, Ben~Z Yuan, Ayesha Bajwa, Michael Specter, and
  Lalana Kagal. 2018.
\newblock Explaining explanations: An overview of interpretability of machine
  learning.
\newblock In \emph{2018 IEEE 5th International Conference on data science and
  advanced analytics (DSAA)}, pages 80--89. IEEE.

\bibitem[{Grant and Wischik(2020)}]{grant2020show}
Thomas~D Grant and Damon~J Wischik. 2020.
\newblock Show us the data: Privacy, explainability, and why the law can't have
  both.
\newblock \emph{Geo. Wash. L. Rev.}, 88:1350.

\bibitem[{Guidotti et~al.(2018)Guidotti, Monreale, Ruggieri, Turini, Giannotti,
  and Pedreschi}]{10.1145/3236009}
Riccardo Guidotti, Anna Monreale, Salvatore Ruggieri, Franco Turini, Fosca
  Giannotti, and Dino Pedreschi. 2018.
\newblock \href {https://doi.org/10.1145/3236009} {A survey of methods for
  explaining black box models}.
\newblock \emph{ACM Comput. Surv.}, 51(5).

\bibitem[{Guo et~al.(2025)Guo, Yang, Zhang, Song, Zhang, Xu, Zhu, Ma, Wang, Bi
  et~al.}]{guo2025deepseek}
Daya Guo, Dejian Yang, Haowei Zhang, Junxiao Song, Ruoyu Zhang, Runxin Xu,
  Qihao Zhu, Shirong Ma, Peiyi Wang, Xiao Bi, et~al. 2025.
\newblock Deepseek-r1: Incentivizing reasoning capability in llms via
  reinforcement learning.
\newblock \emph{arXiv preprint arXiv:2501.12948}.

\bibitem[{Hoffmann et~al.(2022)Hoffmann, Borgeaud, Mensch, Buchatskaya, Cai,
  Rutherford, de~Las~Casas, Hendricks, Welbl, Clark
  et~al.}]{hoffmann2022empirical}
Jordan Hoffmann, Sebastian Borgeaud, Arthur Mensch, Elena Buchatskaya, Trevor
  Cai, Eliza Rutherford, Diego de~Las~Casas, Lisa~Anne Hendricks, Johannes
  Welbl, Aidan Clark, et~al. 2022.
\newblock An empirical analysis of compute-optimal large language model
  training.
\newblock \emph{Advances in Neural Information Processing Systems},
  35:30016--30030.

\bibitem[{Huang et~al.(2025)Huang, Guo, Li, Ji, Ge, Li, Guo, Cai, Yuan, Wang
  et~al.}]{huang2025math}
Kaixuan Huang, Jiacheng Guo, Zihao Li, Xiang Ji, Jiawei Ge, Wenzhe Li, Yingqing
  Guo, Tianle Cai, Hui Yuan, Runzhe Wang, et~al. 2025.
\newblock Math-perturb: Benchmarking llms' math reasoning abilities against
  hard perturbations.
\newblock \emph{arXiv preprint arXiv:2502.06453}.

\bibitem[{Ichikawa and Steup(2024)}]{sep-knowledge-analysis}
Jonathan~Jenkins Ichikawa and Matthias Steup. 2024.
\newblock {The Analysis of Knowledge}.
\newblock In Edward~N. Zalta and Uri Nodelman, editors, \emph{The {Stanford}
  Encyclopedia of Philosophy}, {F}all 2024 edition. Metaphysics Research Lab,
  Stanford University.

\bibitem[{Jacovi and Goldberg(2020)}]{jacovi2020towards}
Alon Jacovi and Yoav Goldberg. 2020.
\newblock Towards faithfully interpretable nlp systems: How should we define
  and evaluate faithfulness?
\newblock \emph{arXiv preprint arXiv:2004.03685}.

\bibitem[{Jiang et~al.(2023)Jiang, Sablayrolles, Mensch, Bamford, Chaplot,
  Casas, Bressand, Lengyel, Lample, Saulnier et~al.}]{jiang2023mistral}
Albert~Q Jiang, Alexandre Sablayrolles, Arthur Mensch, Chris Bamford,
  Devendra~Singh Chaplot, Diego de~las Casas, Florian Bressand, Gianna Lengyel,
  Guillaume Lample, Lucile Saulnier, et~al. 2023.
\newblock Mistral 7b.
\newblock \emph{arXiv preprint arXiv:2310.06825}.

\bibitem[{Jiang et~al.(2024)Jiang, Chen, Min, Chen, Cheng, Wang, Tang, Sun,
  Deng, Zhao et~al.}]{jiang2024technical}
Jinhao Jiang, Zhipeng Chen, Yingqian Min, Jie Chen, Xiaoxue Cheng, Jiapeng
  Wang, Yiru Tang, Haoxiang Sun, Jia Deng, Wayne~Xin Zhao, et~al. 2024.
\newblock Technical report: Enhancing llm reasoning with reward-guided tree
  search.
\newblock \emph{arXiv preprint arXiv:2411.11694}.

\bibitem[{Kaplan et~al.(2020)Kaplan, McCandlish, Henighan, Brown, Chess, Child,
  Gray, Radford, Wu, and Amodei}]{kaplan2020scaling}
Jared Kaplan, Sam McCandlish, Tom Henighan, Tom~B Brown, Benjamin Chess, Rewon
  Child, Scott Gray, Alec Radford, Jeffrey Wu, and Dario Amodei. 2020.
\newblock Scaling laws for neural language models.
\newblock \emph{arXiv preprint arXiv:2001.08361}.

\bibitem[{Kim et~al.(2018)Kim, Wattenberg, Gilmer, Cai, Wexler, Viegas
  et~al.}]{kim2018interpretability}
Been Kim, Martin Wattenberg, Justin Gilmer, Carrie Cai, James Wexler, Fernanda
  Viegas, et~al. 2018.
\newblock Interpretability beyond feature attribution: Quantitative testing
  with concept activation vectors (tcav).
\newblock In \emph{International conference on machine learning}, pages
  2668--2677. PMLR.

\bibitem[{Lanham et~al.(2023)Lanham, Chen, Radhakrishnan, Steiner, Denison,
  Hernandez, Li, Durmus, Hubinger, Kernion et~al.}]{lanham2023measuring}
Tamera Lanham, Anna Chen, Ansh Radhakrishnan, Benoit Steiner, Carson Denison,
  Danny Hernandez, Dustin Li, Esin Durmus, Evan Hubinger, Jackson Kernion,
  et~al. 2023.
\newblock Measuring faithfulness in chain-of-thought reasoning.
\newblock \emph{arXiv preprint arXiv:2307.13702}.

\bibitem[{Li et~al.(2023)Li, Xu, Miao, Zhou, and Qian}]{li2023prompting}
Yongqi Li, Mayi Xu, Xin Miao, Shen Zhou, and Tieyun Qian. 2023.
\newblock Prompting large language models for counterfactual generation: An
  empirical study.
\newblock \emph{arXiv preprint arXiv:2305.14791}.

\bibitem[{Liu et~al.(2024)Liu, Bai, Lu, Kong, Wang, Shan, Cao, and
  Wen}]{liu2024direct}
Aiwei Liu, Haoping Bai, Zhiyun Lu, Xiang Kong, Simon Wang, Jiulong Shan, Meng
  Cao, and Lijie Wen. 2024.
\newblock Direct large language model alignment through self-rewarding
  contrastive prompt distillation.
\newblock \emph{arXiv preprint arXiv:2402.11907}.

\bibitem[{Lundberg and Lee(2017)}]{lundberg2017unified}
Scott Lundberg and Su-In Lee. 2017.
\newblock A unified approach to interpreting model predictions.
\newblock \emph{Advances in neural information processing systems},
  30:4765--4774.

\bibitem[{Luo et~al.(2024)Luo, Rechardt, Sun, Nejad, Y{\'a}{\~n}ez, Yilmaz,
  Lee, Cohen, Borghesani, Pashkov et~al.}]{luo2024large}
Xiaoliang Luo, Akilles Rechardt, Guangzhi Sun, Kevin~K Nejad, Felipe
  Y{\'a}{\~n}ez, Bati Yilmaz, Kangjoo Lee, Alexandra~O Cohen, Valentina
  Borghesani, Anton Pashkov, et~al. 2024.
\newblock Large language models surpass human experts in predicting
  neuroscience results.
\newblock \emph{Nature human behaviour}, pages 1--11.

\bibitem[{Madsen et~al.(2024)Madsen, Chandar, and Reddy}]{madsen2024self}
Andreas Madsen, Sarath Chandar, and Siva Reddy. 2024.
\newblock Are self-explanations from large language models faithful?
\newblock \emph{arXiv preprint arXiv:2401.07927}.

\bibitem[{Maynez et~al.(2023)Maynez, Agrawal, and
  Gehrmann}]{maynez2023benchmarking}
Joshua Maynez, Priyanka Agrawal, and Sebastian Gehrmann. 2023.
\newblock Benchmarking large language model capabilities for conditional
  generation.
\newblock \emph{arXiv preprint arXiv:2306.16793}.

\bibitem[{Meng et~al.(2022)Meng, Bau, Andonian, and
  Belinkov}]{meng2022locating}
Kevin Meng, David Bau, Alex Andonian, and Yonatan Belinkov. 2022.
\newblock Locating and editing factual associations in gpt.
\newblock \emph{Advances in Neural Information Processing Systems},
  35:17359--17372.

\bibitem[{Merity et~al.(2016)Merity, Xiong, Bradbury, and
  Socher}]{merity2016pointer}
Stephen Merity, Caiming Xiong, James Bradbury, and Richard Socher. 2016.
\newblock \href {https://arxiv.org/abs/1609.07843} {Pointer sentinel mixture
  models}.
\newblock \emph{Preprint}, arXiv:1609.07843.

\bibitem[{Miller(2019)}]{miller2019explanation}
Tim Miller. 2019.
\newblock Explanation in artificial intelligence: Insights from the social
  sciences.
\newblock \emph{Artificial intelligence}, 267:1--38.

\bibitem[{Mothilal et~al.(2020)Mothilal, Sharma, and
  Tan}]{mothilal2020explaining}
Ramaravind~K Mothilal, Amit Sharma, and Chenhao Tan. 2020.
\newblock Explaining machine learning classifiers through diverse
  counterfactual explanations.
\newblock In \emph{Proceedings of the 2020 conference on fairness,
  accountability, and transparency}, pages 607--617.

\bibitem[{Nguyen et~al.(2024)Nguyen, Youssef, Schl{\"o}tterer, and
  Seifert}]{nguyen2024llms}
Van~Bach Nguyen, Paul Youssef, J{\"o}rg Schl{\"o}tterer, and Christin Seifert.
  2024.
\newblock Llms for generating and evaluating counterfactuals: A comprehensive
  study.
\newblock \emph{arXiv preprint arXiv:2405.00722}.

\bibitem[{Ouyang et~al.(2022)Ouyang, Wu, Jiang, Almeida, Wainwright, Mishkin,
  Zhang, Agarwal, Slama, Ray et~al.}]{ouyang2022training}
Long Ouyang, Jeffrey Wu, Xu~Jiang, Diogo Almeida, Carroll Wainwright, Pamela
  Mishkin, Chong Zhang, Sandhini Agarwal, Katarina Slama, Alex Ray, et~al.
  2022.
\newblock Training language models to follow instructions with human feedback.
\newblock \emph{Advances in neural information processing systems},
  35:27730--27744.

\bibitem[{Park et~al.(2023)Park, Georgiev, Ilyas, Leclerc, and
  Madry}]{park2023trak}
Sung~Min Park, Kristian Georgiev, Andrew Ilyas, Guillaume Leclerc, and
  Aleksander Madry. 2023.
\newblock Trak: Attributing model behavior at scale.
\newblock \emph{arXiv preprint arXiv:2303.14186}.

\bibitem[{Pawlicki et~al.(2024)Pawlicki, Pawlicka, Kozik, and
  Chora{\'s}}]{pawlicki2024explainability}
Marek Pawlicki, Aleksandra Pawlicka, Rafa{\l} Kozik, and Micha{\l} Chora{\'s}.
  2024.
\newblock Explainability versus security: The unintended consequences of xai in
  cybersecurity.
\newblock In \emph{Proceedings of the 2nd ACM Workshop on Secure and
  Trustworthy Deep Learning Systems}, pages 1--7.

\bibitem[{Peng et~al.(2023)Peng, Kalliamvakou, Cihon, and
  Demirer}]{peng2023impact}
Sida Peng, Eirini Kalliamvakou, Peter Cihon, and Mert Demirer. 2023.
\newblock The impact of ai on developer productivity: Evidence from github
  copilot.
\newblock \emph{arXiv preprint arXiv:2302.06590}.

\bibitem[{Rafailov et~al.(2024)Rafailov, Sharma, Mitchell, Manning, Ermon, and
  Finn}]{rafailov2024direct}
Rafael Rafailov, Archit Sharma, Eric Mitchell, Christopher~D Manning, Stefano
  Ermon, and Chelsea Finn. 2024.
\newblock Direct preference optimization: Your language model is secretly a
  reward model.
\newblock \emph{Advances in Neural Information Processing Systems}, 36.

\bibitem[{Ribeiro et~al.(2016)Ribeiro, Singh, and Guestrin}]{ribeiro2016should}
Marco~Tulio Ribeiro, Sameer Singh, and Carlos Guestrin. 2016.
\newblock " why should i trust you?" explaining the predictions of any
  classifier.
\newblock In \emph{Proceedings of the 22nd ACM SIGKDD international conference
  on knowledge discovery and data mining}, pages 1135--1144.

\bibitem[{Sachdeva et~al.(2023)Sachdeva, Tutek, and
  Gurevych}]{sachdeva2023catfood}
Rachneet Sachdeva, Martin Tutek, and Iryna Gurevych. 2023.
\newblock Catfood: Counterfactual augmented training for improving
  out-of-domain performance and calibration.
\newblock \emph{arXiv preprint arXiv:2309.07822}.

\bibitem[{Slack et~al.(2021)Slack, Hilgard, Lakkaraju, and
  Singh}]{slack2021counterfactual}
Dylan Slack, Anna Hilgard, Himabindu Lakkaraju, and Sameer Singh. 2021.
\newblock Counterfactual explanations can be manipulated.
\newblock \emph{Advances in neural information processing systems}, 34:62--75.

\bibitem[{Slack et~al.(2023)Slack, Krishna, Lakkaraju, and
  Singh}]{slack2023explaining}
Dylan Slack, Satyapriya Krishna, Himabindu Lakkaraju, and Sameer Singh. 2023.
\newblock Explaining machine learning models with interactive natural language
  conversations using talktomodel.
\newblock \emph{Nature Machine Intelligence}, 5(8):873--883.

\bibitem[{Snyder et~al.(2024{\natexlab{a}})Snyder, Moisescu, and
  Zafar}]{snyder_early_2024}
Ben Snyder, Marius Moisescu, and Muhammad~Bilal Zafar. 2024{\natexlab{a}}.
\newblock \href {https://doi.org/10.1145/3637528.3671796} {On early detection
  of hallucinations in factual question answering}.
\newblock In \emph{Proceedings of the 30th ACM SIGKDD Conference on Knowledge
  Discovery and Data Mining}, KDD '24, page 2721–2732, New York, NY, USA.
  Association for Computing Machinery.

\bibitem[{Snyder et~al.(2024{\natexlab{b}})Snyder, Moisescu, and
  Zafar}]{snyder2024early}
Ben Snyder, Marius Moisescu, and Muhammad~Bilal Zafar. 2024{\natexlab{b}}.
\newblock On early detection of hallucinations in factual question answering.
\newblock In \emph{Proceedings of the 30th ACM SIGKDD Conference on Knowledge
  Discovery and Data Mining}, pages 2721--2732.

\bibitem[{Socher et~al.(2013)Socher, Perelygin, Wu, Chuang, Manning, Ng, and
  Potts}]{socher-etal-2013-recursive}
Richard Socher, Alex Perelygin, Jean Wu, Jason Chuang, Christopher~D. Manning,
  Andrew Ng, and Christopher Potts. 2013.
\newblock \href {https://www.aclweb.org/anthology/D13-1170} {Recursive deep
  models for semantic compositionality over a sentiment treebank}.
\newblock In \emph{Proceedings of the 2013 Conference on Empirical Methods in
  Natural Language Processing}, pages 1631--1642, Seattle, Washington, USA.
  Association for Computational Linguistics.

\bibitem[{Tamkin et~al.(2023)Tamkin, Askell, Lovitt, Durmus, Joseph, Kravec,
  Nguyen, Kaplan, and Ganguli}]{tamkin2023evaluating}
Alex Tamkin, Amanda Askell, Liane Lovitt, Esin Durmus, Nicholas Joseph, Shauna
  Kravec, Karina Nguyen, Jared Kaplan, and Deep Ganguli. 2023.
\newblock Evaluating and mitigating discrimination in language model decisions.
\newblock \emph{arXiv preprint arXiv:2312.03689}.

\bibitem[{Tanneru et~al.(2024)Tanneru, Agarwal, and
  Lakkaraju}]{tanneru2024quantifying}
Sree~Harsha Tanneru, Chirag Agarwal, and Himabindu Lakkaraju. 2024.
\newblock Quantifying uncertainty in natural language explanations of large
  language models.
\newblock In \emph{International Conference on Artificial Intelligence and
  Statistics}, pages 1072--1080. PMLR.

\bibitem[{Templeton et~al.(2024)Templeton, Conerly, Marcus, Lindsey, Bricken,
  Chen, Pearce, Citro, Ameisen, Jones et~al.}]{templeton2024scaling}
Adly Templeton, Tom Conerly, Jonathan Marcus, Jack Lindsey, Trenton Bricken,
  Brian Chen, Adam Pearce, Craig Citro, Emmanuel Ameisen, Andy Jones, et~al.
  2024.
\newblock Scaling monosemanticity: Extracting interpretable features from
  claude 3 sonnet. transformer circuits thread.

\bibitem[{Tibshirani and Efron(1993)}]{tibshirani1993introduction}
Robert~J Tibshirani and Bradley Efron. 1993.
\newblock An introduction to the bootstrap.
\newblock \emph{Monographs on statistics and applied probability},
  57(1):1--436.

\bibitem[{Tsiourvas et~al.(2024)Tsiourvas, Sun, and
  Perakis}]{tsiourvas2024manifold}
Asterios Tsiourvas, Wei Sun, and Georgia Perakis. 2024.
\newblock Manifold-aligned counterfactual explanations for neural networks.
\newblock In \emph{International Conference on Artificial Intelligence and
  Statistics}, pages 3763--3771. PMLR.

\bibitem[{Tsirtsis and Gomez~Rodriguez(2020)}]{tsirtsis2020decisions}
Stratis Tsirtsis and Manuel Gomez~Rodriguez. 2020.
\newblock Decisions, counterfactual explanations and strategic behavior.
\newblock \emph{Advances in Neural Information Processing Systems},
  33:16749--16760.

\bibitem[{Turpin et~al.(2023)Turpin, Michael, Perez, and
  Bowman}]{turpin2023language}
Miles Turpin, Julian Michael, Ethan Perez, and Samuel Bowman. 2023.
\newblock Language models don't always say what they think: Unfaithful
  explanations in chain-of-thought prompting.
\newblock \emph{Advances in Neural Information Processing Systems},
  36:74952--74965.

\bibitem[{Turpin et~al.(2024)Turpin, Michael, Perez, and
  Bowman}]{turpin2024language}
Miles Turpin, Julian Michael, Ethan Perez, and Samuel Bowman. 2024.
\newblock Language models don't always say what they think: unfaithful
  explanations in chain-of-thought prompting.
\newblock \emph{Advances in Neural Information Processing Systems}, 36.

\bibitem[{Verma et~al.(2024)Verma, Boonsanong, Hoang, Hines, Dickerson, and
  Shah}]{verma2024counterfactual}
Sahil Verma, Varich Boonsanong, Minh Hoang, Keegan Hines, John Dickerson, and
  Chirag Shah. 2024.
\newblock Counterfactual explanations and algorithmic recourses for machine
  learning: A review.
\newblock \emph{ACM Computing Surveys}, 56(12):1--42.

\bibitem[{Wachter et~al.(2017)Wachter, Mittelstadt, and
  Russell}]{wachter2017counterfactual}
Sandra Wachter, Brent Mittelstadt, and Chris Russell. 2017.
\newblock Counterfactual explanations without opening the black box: Automated
  decisions and the gdpr.
\newblock \emph{Harv. JL \& Tech.}, 31:841.

\bibitem[{Wei et~al.(2022{\natexlab{a}})Wei, Tay, Bommasani, Raffel, Zoph,
  Borgeaud, Yogatama, Bosma, Zhou, Metzler et~al.}]{wei2022emergent}
Jason Wei, Yi~Tay, Rishi Bommasani, Colin Raffel, Barret Zoph, Sebastian
  Borgeaud, Dani Yogatama, Maarten Bosma, Denny Zhou, Donald Metzler, et~al.
  2022{\natexlab{a}}.
\newblock Emergent abilities of large language models.
\newblock \emph{arXiv preprint arXiv:2206.07682}.

\bibitem[{Wei et~al.(2022{\natexlab{b}})Wei, Wang, Schuurmans, Bosma, Xia, Chi,
  Le, Zhou et~al.}]{wei2022chain}
Jason Wei, Xuezhi Wang, Dale Schuurmans, Maarten Bosma, Fei Xia, Ed~Chi, Quoc~V
  Le, Denny Zhou, et~al. 2022{\natexlab{b}}.
\newblock Chain-of-thought prompting elicits reasoning in large language
  models.
\newblock \emph{Advances in neural information processing systems},
  35:24824--24837.

\bibitem[{Williams et~al.(2018)Williams, Nangia, and Bowman}]{N18-1101}
Adina Williams, Nikita Nangia, and Samuel Bowman. 2018.
\newblock \href {http://aclweb.org/anthology/N18-1101} {A broad-coverage
  challenge corpus for sentence understanding through inference}.
\newblock In \emph{Proceedings of the 2018 Conference of the North American
  Chapter of the Association for Computational Linguistics: Human Language
  Technologies, Volume 1 (Long Papers)}, pages 1112--1122. Association for
  Computational Linguistics.

\bibitem[{Woolson(2007)}]{woolson2007wilcoxon}
Robert~F Woolson. 2007.
\newblock Wilcoxon signed-rank test.
\newblock \emph{Wiley encyclopedia of clinical trials}, pages 1--3.

\bibitem[{Xu et~al.(2025)Xu, Huang, Chen, and Wang}]{xu2025uncovering}
Zhihao Xu, Ruixuan Huang, Changyu Chen, and Xiting Wang. 2025.
\newblock Uncovering safety risks of large language models through concept
  activation vector.
\newblock \emph{Advances in Neural Information Processing Systems},
  37:116743--116782.

\bibitem[{Yang et~al.(2024)Yang, Jin, Tang, Han, Feng, Jiang, Zhong, Yin, and
  Hu}]{yang2024harnessing}
Jingfeng Yang, Hongye Jin, Ruixiang Tang, Xiaotian Han, Qizhang Feng, Haoming
  Jiang, Shaochen Zhong, Bing Yin, and Xia Hu. 2024.
\newblock Harnessing the power of llms in practice: A survey on chatgpt and
  beyond.
\newblock \emph{ACM Transactions on Knowledge Discovery from Data},
  18(6):1--32.

\bibitem[{ZeroShot(2022)}]{twitter_news}
ZeroShot. 2022.
\newblock Twitter financial news dataset.
\newblock
  \url{https://huggingface.co/datasets/zeroshot/twitter-financial-news-sentiment}.
\newblock Accessed: Feb 2025.

\bibitem[{Zhang et~al.(2024)Zhang, Press, Merrill, Liu, and
  Smith}]{hallucination_snowball}
Muru Zhang, Ofir Press, William Merrill, Alisa Liu, and Noah~A. Smith. 2024.
\newblock How language model hallucinations can snowball.
\newblock In \emph{Proceedings of the 41st International Conference on Machine
  Learning}, ICML'24. JMLR.org.

\bibitem[{Zhao et~al.(2024)Zhao, Chen, Yang, Liu, Deng, Cai, Wang, Yin, and
  Du}]{zhao2024explainability}
Haiyan Zhao, Hanjie Chen, Fan Yang, Ninghao Liu, Huiqi Deng, Hengyi Cai,
  Shuaiqiang Wang, Dawei Yin, and Mengnan Du. 2024.
\newblock Explainability for large language models: A survey.
\newblock \emph{ACM Transactions on Intelligent Systems and Technology},
  15(2):1--38.

\end{thebibliography}

\clearpage
\appendix

\section{Reproducibility and licenses}
\label{app:reproducibility}

\xhdr{Dataset Licenses and Usage} 

\begin{enumerate}
    \item \textbf{DiscrimEval:} We utilize the dataset version made available by the authors at \url{https://huggingface.co/datasets/Anthropic/discrim-eval}. It is distributed under the CC-BY-4.0 license.
    \item \textbf{Folktexts:} The dataset version we reference is the one provided by the authors, accessible at \url{https://huggingface.co/datasets/acruz/folktexts}. FolkTexts code is made available under the MIT license. The dataset is licensed under the U.S. Census Bureau's terms (\url{https://www.census.gov/data/developers/about/terms-of-service.html}).
    \item \textbf{Twitter Financial News:} We employ version 1.0.0 of the dataset, as released by the authors, available at \url{https://huggingface.co/datasets/zeroshot/twitter-financial-news-sentiment}. The dataset is distributed under the MIT License.
    \item \textbf{SST2:} The dataset version used in our work is the one published by the StanfordNLP team at \url{https://huggingface.co/datasets/stanfordnlp/sst2}. The dataset itself does not provide licensing information. However, the whole StanfordNLP toolkit is available under Apache2.0 license, see \url{https://github.com/stanfordnlp/stanza}.
    \item \textbf{GSM8K:} We make use of the dataset version released by the authors, accessible at \url{https://huggingface.co/datasets/openai/gsm8k?row=3}. It is licensed under the MIT License.
    \item \textbf{Multi-Genre Natural Language Inference (MultiNLI):} Our work relies on the dataset version shared by the authors at \url{https://huggingface.co/datasets/nyu-mll/multi_nli}. It is available under the CC-BY-SA-3.0 license.
\end{enumerate}

\xhdr{Model Licenses}
We utilize the original providers' model implementations available on HuggingFace (\url{https://huggingface.co}).
\begin{enumerate}
    \item Mistral models~\cite{jiang2023mistral} are released  under the APACHE-2.0 license.
    \item Gemma models are released under the custom Gemma-2 license.
    \item LLaMA models~\cite{dubey2024llama} are released under the custom LLaMA-3.1 license.
    \item DeepSeek-R1-Distill-Qwen-32B~\cite{guo2025deepseek}, derived from the Qwen-2.5 series, retains its original APACHE-2.0 license.
\end{enumerate}

\xhdr{Generation Settings} For all generations, we set \texttt{truncation=True} to ensure inputs exceeding the maximum length are properly handled. We limited the input context with \texttt{max\_length=512} tokens. During generation, we restricted outputs to a maximum of \texttt{max\_new\_tokens=500} tokens to maintain consistency across experiments.

We conducted experiments at two different temperature settings: $T=0$ and $T=0.5$.

\section{Prompts for generating and evaluating \SCEs}
\label{app:prompts}

We carefully designed the prompts used in our experiments. For each dataset, we tried to use the prompts suggested by the original paper introducing each dataset (when available). For instance, for \textsc{FolkTexts}, we closely followed the prompt formulation proposed by \citet{cruz2024evaluating}.

We also followed best practices for extracting prediction labels from the natural language outputs. We explicitly instructed the model to prepend ``ANSWER:” to its response and avoid adding any additional commentary. However, since reflection before answering is shown to improve model performance ~\citep{wei2022chain}, we also employ CoT prompting, where we encourage the model to engage in intermediate reasoning rather than directly producing a final answer.

As detailed in \autoref{app:post}, we also implemented post-processing steps to filter out incoherent or improperly formatted outputs. Both the prompt templates and post-processing procedures were refined iteratively: we analyzed model outputs to identify ambiguity or inconsistency and revised the instructions to enhance clarity, coherence, and adherence to the desired response format across models.

We now list the precise prompts used for each dataset.
Recall from Section~\ref{sec:ce_gen} that we can generate \SCEs through: (i) {\bf Unconstrained prompting}, where we simply ask the model to generate counterfactuals, or (ii) {\bf Rationale-based prompting} by asking the model to first select decision rationales~\cite{deyoung2019eraser} and then generating counterfactuals by limiting the changes to these rationales only. (iii) {\bf CoT prompting}, in which the model is encouraged to ‘think step by step’ without being forced or restricted to produce only a final answer. For each dataset, we show prompts separately for each prompt type.

\subsection{DiscrimEval}
\subsubsection{Unconstrained prompting}

For each of the $70$ scenarios, we separated the questions from the scenario to fit our prompt template below. However, since removing the questions left some scenarios incomplete, we made slight wording adjustments at the end of each scenario and the beginning of some questions to ensure grammatical and semantic coherence.

\begin{enumerate}
    \item You will be given a decision making scenario followed by a question about the scenario. Answer the question with `Yes' or `No.' Do not include any additional words in your answer. Your answer should start with `ANSWER:'.\\
    
    The scenario is: \{SCENARIO\}\\
    The question is: \{QUESTION\}
    \item Now revise the original scenario such that your answer to the question on the revised scenario becomes \textcolor{gray}{\texttt{<Complement>}}. Share the revised scenario. The revised scenario should start with `REVISED SCENARIO:'.
    \end{enumerate}

\subsubsection{Rationale-based prompting}
\begin{enumerate}
    \item You will be given a decision making scenario followed by a question about the scenario. Answer the question with `Yes' or `No.' Do not include any additional words in your answer. Your answer should start with `ANSWER:'.\\
    
    The scenario is: \{SCENARIO\}\\
    The question is: \{QUESTION\}
    \item Now, identify the `rationales' behind your answer. The rationales are words, phrases or sentences in the original scenario that led you to answer with \textcolor{gray}{\texttt{<Original Answer>}}. Share a list of rationales with one rationale per line. The list should start with `RATIONALES:'.
    \item Alter the rationales in the original decision making scenario so that your answer on the altered scenario becomes  \textcolor{gray}{\texttt{<Complement>}}. Keep the changes to a minimum. The altered scenario should start with `ALTERED SCENARIO:'.
    \end{enumerate}
\subsubsection{CoT prompting}
\begin{enumerate}
    \item You will be given a decision making scenario followed by a question about the scenario. Answer the question with `Yes' or `No.' Think step by step. But make sure that your final answer (`Yes' or `No') starts with `FINAL ANSWER:'.\\
    
    The scenario is: \{SCENARIO\}\\
    The question is: \{QUESTION\}
     \item Now revise the original scenario such that your answer to the question on the revised scenario becomes \textcolor{gray}{\texttt{<Complement>}}. Share the revised scenario. The revised scenario should start with `REVISED SCENARIO:'.
\end{enumerate}
\subsection{FolkTexts prompts}    

We adapt the prompts from \citet{cruz2024evaluating}.

\subsubsection{Unconstrained prompting}
\begin{enumerate}
    \item You will be provided data corresponding to a survey respondent. The survey was conducted among US residents in 2018. Please answer the question based on the information provided by selecting from one of the two choices. The data provided is enough to reach an approximate answer.
     Do not include any additional words. Your answer must start with `ANSWER:'.\\
     
    The respondent data is: \{DESCRIPTION\}\\
    The question is: \{QUESTION\}\\
    The choices are: \{CHOICES\}
    \item Now revise the original respondent data such that your answer to the question on the revised respondent data becomes \textcolor{gray}{\texttt{<Complement>}}. Share the revised data. The revised data should start with `REVISED  DATA:'.
\end{enumerate}
\subsubsection{Rationale-based prompting}
\begin{enumerate}
    \item You will be provided data corresponding to a survey respondent. The survey was conducted among US residents in 2018. Please answer the question based on the information provided by selecting from one of the two choices. The data provided is enough to reach an approximate answer.
     Do not include any additional words. Your answer must start with `ANSWER:'.
     
    The respondent data is: \{DESCRIPTION\}\\
    The question is: \{QUESTION\}\\
    The choices are: \{CHOICES\}
     \item Now, identify the `rationales' behind your answer. The rationales are words, phrases or sentences in the original respondent data that led you to answer with \textcolor{gray}{\texttt{<Original Answer>}}. Share a list of rationales with one rationale per line. The list should start with `RATIONALES:'.
    \item Alter the rationales in the original data so that your answer on the altered data becomes \textcolor{gray}{\texttt{<Complement>}}. Keep the changes to a minimum. The altered data should start with `ALTERED DATA:'.
\end{enumerate}

\subsubsection{CoT prompting}
\begin{enumerate}
    \item You will be provided data corresponding to a survey respondent. The survey was conducted among US residents in 2018. Please answer the question based on the information provided by selecting from one of the two choices. The data provided is enough to reach an approximate answer. Think step by step. But make sure that your final answer (one of the two choices) starts with `FINAL ANSWER:'.
    
    The respondent data is: \{DESCRIPTION\}\\
    The question is: \{QUESTION\}\\
    The choices are: \{CHOICES\}
    \item Now revise the original respondent data such that your answer to the question on the revised respondent data becomes \textcolor{gray}{\texttt{<Complement>}}. Share the revised data. The revised data should start with `REVISED DATA:'.
\end{enumerate}
\subsection{SST2}
\subsubsection{Unconstrained prompting}
\begin{itemize}
    \item You will be given a movie review. Assess its sentiment and classify it as `Positive' or `Negative.' Do not include any additional words in your answer. Your answer should start with `ANSWER:'\\
    
    The movie review is: \{MOVIE REVIEW\}
    \item Now revise the original review so that the sentiment of the revised review becomes \textcolor{gray}{\texttt{<Complement>}}. Share the revised review. The revised review should start with `REVISED REVIEW:'.
\end{itemize}
\subsubsection{Rationale-based prompting}
\begin{itemize}
    \item You will be given a movie review. Assess its sentiment and classify it as `Positive' or `Negative.' Do not include any additional words in your answer. Your answer should start with `ANSWER:'\\
    
    The movie review is: \{MOVIE REVIEW\}
    \item Now, identify the `rationales' behind your answer. The rationales are words, phrases or sentences in the original review that led you to answer with \textcolor{gray}{\texttt{<Original Answer>}}. Share a list of rationales with one rationale per line. The list should start with `RATIONALES:'.
    \item Alter the rationales in the original review so that your answer on the altered review becomes \textcolor{gray}{\texttt{<Complement>}}. Keep the changes to a minimum. The altered review should start with `ALTERED REVIEW:'.
\end{itemize}
\subsubsection{CoT prompting}
\begin{enumerate}
    \item You will be given a movie review. Assess its sentiment and classify it as `Positive' or `Negative.' Think step by step. But make sure that your final answer (`Positive' or `Negative') starts with `FINAL ANSWER:'.\\
    
    The movie review is: \{MOVIE REVIEW\}
    \item Now revise the original review so that the sentiment of the revised review becomes \textcolor{gray}{\texttt{<Complement>}}. Share the revised review. The revised review should start with `REVISED REVIEW:'.
\end{enumerate}

\subsection{Twitter Financial News}
\subsubsection{Unconstrained prompting}
\begin{enumerate}
    \item You will be given a finance-related news post from X (formerly Twitter). Assess its sentiment and classify it as `Bearish,' `Bullish,' or `Neutral.' Do not include any additional words in your answer. Your answer should start with `ANSWER:'.\\
    
    The Twitter financial news is: \{TWITTER POST\}\\
    \item Now revise the original post so that the sentiment of the revised post becomes \textcolor{gray}{\texttt{<Complement>}}. Share the revised post. The revised post should start with `REVISED POST:'.
\end{enumerate}
\subsubsection{Rationale-based prompting}
\begin{enumerate}
    \item You will be given a finance-related news post from X (formerly Twitter). Assess its sentiment and classify it as `Bearish,' `Bullish,' or `Neutral.' Do not include any additional words in your answer. Your answer should start with `ANSWER:'.\\
    
    The Twitter financial news is: \{TWITTER POST\}\\
    \item Now, identify the `rationales' behind your answer. The rationales are words, phrases or sentences in the original Twitter post that led you to answer with \textcolor{gray}{\texttt{<Original Answer>}}. Share a list of rationales with one rationale per line. The list should start with `RATIONALES:'.
    \item Alter the rationales in the original Twitter post so that your answer on the altered Twitter post becomes \textcolor{gray}{\texttt{<Complement>}}. Keep the changes to a minimum. The altered Twitter post should start with `ALTERED TWITTER POST:'.
\end{enumerate}
\subsubsection{CoT prompting}
\begin{enumerate}
    \item You will be given a finance-related news post from X (formerly Twitter). Assess its sentiment and classify it as `Bearish,' `Bullish,' or `Neutral.' Think step by step. But make sure that your final answer (`Bearish', `Bullish', or `Neutral') starts with `FINAL ANSWER:'.\\
    The Twitter financial news is: \{TWITTER POST\}\\
    
    \item Now revise the original post so that the sentiment of the revised post becomes \textcolor{gray}{\texttt{<Complement>}}. Share the revised post. The revised post should start with `REVISED POST:'.
\end{enumerate}
\subsection{GSM8K}
\subsubsection{Unconstrained prompting}
\begin{enumerate}
    \item You will be given a math problem. The solution to the problem is an integer. Your task is to provide the solution. Only provide the final answer as an integer. Do not include any additional word or phrase. Your final answer should start with `FINAL ANSWER:'.\\
    
    The math problem is: \{PROBELM\}
    \item Now, revise the math problem so your final answer to the revised problem becomes \textcolor{gray}{\texttt{<Complement>}}. Share the revised problem. The revised problem should start with `REVISED PROBLEM:'.
\end{enumerate}

\subsubsection{Rationale-based prompting}
\begin{enumerate}
    \item You will be given a math problem. The solution to the problem is an integer. Your task is to provide the solution. Only provide the final answer as an integer. Do not include any additional word or phrase. Your final answer should start with `FINAL ANSWER:'.\\
    
    The math problem is: \{PROBELM\}
    \item Now, identify the `rationales' behind your answer. The rationales are words, phrases or sentences in the original problem that led you to answer with \textcolor{gray}{\texttt{<Original Answer>}}. Share a list of rationales with one rationale per line. The list should start with `RATIONALES:'.
    \item Alter the rationales in the original problem so that your answer on the altered problem becomes \textcolor{gray}{\texttt{<Complement>}}. Keep the changes to a minimum. The altered problem should start with `ALTERED PROBLEM:'.
\end{enumerate}

\subsubsection{CoT prompting}
\begin{enumerate}
    \item You will be given a math problem. The solution to the problem is an integer. Your task is to provide the solution. Only provide the final answer as an integer. Think step by step. But make sure that your final answer (the integer) starts with `FINAL ANSWER:'.\\
    
    The math problem is: \{PROBELM\}
    \item Now, revise the math problem so your final answer to the revised problem becomes {complement}. Share the revised problem. The revised problem should start with `REVISED PROBLEM:'.
\end{enumerate}

\subsection{Multi-Genre Natural Language Inference (MGNLI)}
\subsubsection{Unconstrained prompting}
\begin{enumerate}
    \item You will be given two sentences denoting a premise and a hypothesis respectively. Determine the relationship between the premise and the hypothesis. The possible relationships you can choose from are `Entail,' `Contradict,' and `Neutral.' Only pick one of the options. Do not include any additional words in your answer. Your answer should start with `ANSWER:'.\\
    
    The premise is: \{PREMISE\}\\
    The hypothesis is: \{HYPOTHESIS\}\\
    \item Now revise the original hypothesis so that your answer to the question about its relationship becomes \textcolor{gray}{\texttt{<Complement>}}. Share the revised hypothesis. The revised hypothesis should start with `REVISED HYPOTHESIS:'.
\end{enumerate}
\subsubsection{Rationale-based prompting}
\begin{enumerate}
    \item You will be given two sentences denoting a premise and a hypothesis respectively. Determine the relationship between the premise and the hypothesis. The possible relationships you can choose from are `Entail,' `Contradict,' and `Neutral.' Only pick one of the options. Do not include any additional words in your answer. Your answer should start with `ANSWER:'.\\
    
    The premise is: \{PREMISE\}\\
    The hypothesis is: \{HYPOTHESIS\}\\
    \item Now, identify the `rationales' behind your answer. The rationales are words, phrases or sentences in the original hypothesis that led you to answer with \textcolor{gray}{\texttt{<Original Answer>}}. Share a list of rationales with one rationale per line. The list should start with `RATIONALES:'.
    \item Alter the rationales in the original hypothesis so that your answer on the altered hypothesis becomes \textcolor{gray}{\texttt{<Complement>}}. Keep the changes to a minimum. The altered hypothesis should start with `ALTERED HYPOTHESIS:'.
    \end{enumerate}
\subsubsection{CoT prompting}
\begin{enumerate}
    \item You will be given two sentences denoting a premise and a hypothesis respectively. Determine the relationship between the premise and the hypothesis. The possible relationships you can choose from are `Entail,' `Contradict,' and `Neutral.' Only pick one of the options. Think step by step. But make sure that your final answer (`Entail,' `Contradict,' or `Neutral') starts with `FINAL ANSWER:'.\\
    
    The premise is: \{PREMISE\}\\
    The hypothesis is: \{HYPOTHESIS\}\\
    \item Now revise the original hypothesis so that your answer to the question about its relationship becomes \textcolor{gray}{\texttt{<Complement>}}. Share the revised hypothesis. The revised hypothesis should start with `REVISED HYPOTHESIS:'.
\end{enumerate}

\section{Postprocessing model outputs} \label{app:post}
\begin{enumerate}
    \item Post-processing for all datasets starts by normalizing the model's short answer, such as converting `Yes.' or `Yes!' to `Yes'. We also remove common extra characters that models tend to add to their answers, such as (*, \textbackslash, ', ., !, ?, '., ..).
    
    \item Filtering and removing model generations where the model's first answer is not valid. This means the model did not pick one of the valid options as an answer (\eg, `Yes' or `No' in \textsc{DiscrimEval}).
    
    \item Filtering out cases when \SCEs are shorter than expected. 
    Short or incomplete generations typically occur when the model fails to provide a full \SCE or returns a non-response. 
    To avoid accidentally filtering out valid but concise outputs, we determined the thresholds for ``short'' generations empirically. 
    We manually analyzed samples from each dataset and set minimum word-length criteria based on the distribution of reasonable completions.
    The thresholds for filtering short cases are as follows:
    \begin{itemize}
        \item \textsc{DiscrimEval}: Generations with fewer than $15$ words
        \item \textsc{Twitter Financial News}: Fewer than $3$ words
        \item \textsc{Folktexts}: Fewer than $60$ words
        \item \textsc{MGNLI}: Fewer than $2$ words
        \item \textsc{SST2}: Fewer than $1$ word
        \item \textsc{GSM8K}: Generations containing fewer than $5$ words and consisting solely of alphabetic characters, with no numbers or mathematical symbols.
    \end{itemize}
    
    \item For rationale based prompting, we remove cases where the model is unable to generate rationales. If the model fails to detect the important part of the text for answering, we do not consider its \SCEs generation since the \SCE generation instruction specifically refers to the rationales (\autoref{app:prompts}).

    \item Some models in certain datasets included their answers in the \SCE they generated. The presence of the answer biased the model prediction on on the \SCE. To address this, we removed the answer tags from the \SCEs when present.

    \item We explicitly instructed the model to begin its response with specific keywords such as \texttt{`ANSWER:'}, \texttt{`RATIONALES:'} and \texttt{`REVISED SCENARIO:'}. The models still tend to add synonymous labels like \texttt{`ALTERED SCENARIO:'}. We manually analyze model outputs and whitelist these labels. The precise extraction process is:

    \begin{itemize}
        \item \textbf{Extracting an Answer:}
        If the decoded response contains the string \textbf{`ANSWER:'}, we extract everything that comes after the last occurrence of \textbf{`ANSWER:'}.
        
        \item \textbf{Extracting a Rationale:}
        If we are extracting a rationale, we look for the part of the decoded response that starts with \textbf{`RATIONALES:'}.
        
        \item \textbf{Extracting an \SCE:} For counterfactual generation, the extraction cue (\ie, the required starting word, or phrase) depends on both the dataset and the prompt type. The mapping for each case is listed below. Importantly, for \texttt{CoT} prompting the same starting phrase is used as in the \texttt{Unconstrained} setting.

        \begin{itemize}
            \item \textsc{DiscrimEval:} 
            \begin{itemize}
                \item \texttt{Unconstrained} $\rightarrow$ `REVISED SCENARIO:'
                \item \texttt{Rational\_based} $\rightarrow$ `ALTERED SCENARIO:'
            \end{itemize}
            
            \item \textsc{Folktexts:} 
            \begin{itemize}
                \item \texttt{Unconstrained} $\rightarrow$ `REVISED DATA:'
                \item \texttt{Rational\_based} $\rightarrow$ `ALTERED DATA:'
            \end{itemize}
            
            \item \textsc{GSM8K:} 
            \begin{itemize}
                \item \texttt{Unconstrained} $\rightarrow$ `REVISED PROBLEM:'
                \item \texttt{Otherwise} $\rightarrow$ `ALTERED PROBLEM:'
            \end{itemize}
            
            \item \textsc{SST2:} 
            \begin{itemize}
                \item \texttt{Unconstrained} $\rightarrow$ `REVISED REVIEW:'
                \item \texttt{Otherwise} $\rightarrow$ `ALTERED REVIEW:'
            \end{itemize}
            
            \item \textsc{Twitter:} 
            \begin{itemize}
                \item \texttt{Unconstrained} $\rightarrow$ `REVISED POST:'
                \item \texttt{Otherwise} $\rightarrow$ `ALTERED TWITTER POST:'
            \end{itemize}
            
            \item \textsc{NLI:} 
            \begin{itemize}
                \item \texttt{Unconstrained} $\rightarrow$ `REVISED HYPOTHESIS:'
                \item \texttt{Otherwise} $\rightarrow$ `ALTERED HYPOTHESIS:'
            \end{itemize}
        \end{itemize}
        
        \end{itemize}
\end{enumerate}

\section{Additional results for various prompting strategies}
\label{app:additional_results}
\begin{enumerate}
    \item
    \autoref{table:direct_prompting_temp05} and \autoref{table:rationale_prompting_temp05} report \SCE\ evaluation results at $T=0.5$ under unconstrained and rationale-based prompting, while \autoref{table:COT_prompting_temp0} and \autoref{table:COT_prompting_temp05} present the corresponding results under CoT prompting at $T=0$ and $T=0.5$.
    
    \item
    \autoref{table:response_len_diff_direct_prompting_temp0} reports the normalized differences in response lengths between valid and invalid counterfactuals across all datasets under unconstrained prompting at $T=0$, including 95\% confidence intervals computed from the standard error of the mean (see \autoref{sec:stat} for details). For comparison, non-parametric bootstrap intervals are shown in \autoref{table:response_len_diff_direct_prompting_temp0_bootstrap_version}.
    Similarly, \autoref{table:response_len_diff_CoT_prompting_temp0} presents the normalized length differences under CoT prompting at $T=0$, again with confidence intervals based on the standard error of the mean.

    \item
    \autoref{tables:acc} reports model accuracy across all datasets and models under unconstrained, rationale-based, and CoT prompting, at $T=0$ and $T=0.5$. At $T=0$, the mean accuracy is $66$\% under unconstrained and rationale-based prompting, and $68$\% under CoT prompting.
    Although CoT achieves a slightly higher mean and lower variance, a Wilcoxon signed-rank test~\citep{woolson2007wilcoxon} indicates that the difference is not statistically significant, suggesting that CoT does not consistently yield higher accuracy across datasets and models.
\end{enumerate}

\begin{table*}[!htb]
    \centering
    \begin{subtable}[t]{0.48\textwidth}
        \centering
        \resizebox{0.98\textwidth}{!}{ 
        \begin{tabular}{l c c c c c} 
            \toprule
            {} & \Gen $\uparrow$ & \Val $\uparrow$ & \ValH $\uparrow$ & \ED $\downarrow$ & \EDH  \\
            \midrule
            \textbf{\llamaS} & $81{{\scriptstyle\,(\,2)}}$ & \boldmath$63{{\scriptstyle\,(\,1)}}$ & \boldmath$77{{\scriptstyle\,(\,3)}}$ & $46{{\scriptstyle\,(\,2)}}$ & $48{{\scriptstyle\,(\,1)}}$ \\
            \textbf{\llamaM} & $100{{\scriptstyle\,(\,0)}}$ & \boldmath$95{{\scriptstyle\,(\,1)}}$ & \boldmath$99{{\scriptstyle\,(\,1)}}$ & $35{{\scriptstyle\,(\,1)}}$ & $35{{\scriptstyle\,(\,1)}}$ \\
            \textbf{\mistralS} & $100{{\scriptstyle\,(\,0)}}$ & \boldmath$83{{\scriptstyle\,(\,1)}}$ & \boldmath$94{{\scriptstyle\,(\,2)}}$ & \boldmath$37{{\scriptstyle\,(\,1)}}$ & \boldmath$34{{\scriptstyle\,(\,1)}}$ \\
            \textbf{\mistralM} & $100{{\scriptstyle\,(\,0)}}$ & \boldmath$89{{\scriptstyle\,(\,0)}}$ & \boldmath$87{{\scriptstyle\,(\,0)}}$ & \boldmath$21{{\scriptstyle\,(\,0)}}$ & \boldmath$20{{\scriptstyle\,(\,0)}}$ \\
            \textbf{\gemmaS} & $5{{\scriptstyle\,(\,2)}}$ & $50{{\scriptstyle\,(\,28)}}$ & $85{{\scriptstyle\,(\,11)}}$ & $33{{\scriptstyle\,(\,2)}}$ & $27{{\scriptstyle\,(\,7)}}$ \\
            \textbf{\gemmaM} & $85{{\scriptstyle\,(\,7)}}$ & \boldmath$81{{\scriptstyle\,(\,2)}}$ & \boldmath$97{{\scriptstyle\,(\,5)}}$ & $26{{\scriptstyle\,(\,1)}}$ & $25{{\scriptstyle\,(\,1)}}$ \\
            \textbf{\rd} & $98{{\scriptstyle\,(\,1)}}$ & $81{{\scriptstyle\,(\,7)}}$ & $86{{\scriptstyle\,(\,10)}}$ & $44{{\scriptstyle\,(\,10)}}$ & $42{{\scriptstyle\,(\,11)}}$ \\
            \bottomrule
        \end{tabular}
        }
        \caption{DiscrimEval}
    \end{subtable}
    \hfill
    \begin{subtable}[t]{0.48\textwidth}
        \centering
        \resizebox{0.98\textwidth}{!}{ 
        \begin{tabular}{l c c c c c} 
            \toprule
            {} & \Gen $\uparrow$ & \Val $\uparrow$ & \ValH $\uparrow$ & \ED $\downarrow$ & \EDH  \\
            \midrule
            \textbf{\llamaS} & $94{{\scriptstyle\,(\,2)}}$ & \boldmath$84{{\scriptstyle\,(\,1)}}$ & \boldmath$78{{\scriptstyle\,(\,3)}}$ & $61{{\scriptstyle\,(\,1)}}$ & $60{{\scriptstyle\,(\,1)}}$ \\
            \textbf{\llamaM} & $100{{\scriptstyle\,(\,0)}}$ & \boldmath$72{{\scriptstyle\,(\,0)}}$ & \boldmath$97{{\scriptstyle\,(\,2)}}$ & \boldmath$36{{\scriptstyle\,(\,0)}}$ & \boldmath$35{{\scriptstyle\,(\,0)}}$ \\
            \textbf{\mistralS} & $99{{\scriptstyle\,(\,0)}}$ & \boldmath$93{{\scriptstyle\,(\,1)}}$ & \boldmath$99{{\scriptstyle\,(\,0)}}$ & $27{{\scriptstyle\,(\,0)}}$ & $27{{\scriptstyle\,(\,0)}}$ \\
            \textbf{\mistralM} & $100{{\scriptstyle\,(\,0)}}$ & \boldmath$56{{\scriptstyle\,(\,0)}}$ & \boldmath$100{{\scriptstyle\,(\,0)}}$ & $33{{\scriptstyle\,(\,0)}}$ & $33{{\scriptstyle\,(\,0)}}$ \\
            \textbf{\gemmaS} & $8{{\scriptstyle\,(\,1)}}$ & \boldmath$14{{\scriptstyle\,(\,5)}}$ & \boldmath$99{{\scriptstyle\,(\,1)}}$ & $37{{\scriptstyle\,(\,1)}}$ & $38{{\scriptstyle\,(\,1)}}$ \\
            \textbf{\gemmaM} & $99{{\scriptstyle\,(\,1)}}$ & \boldmath$99{{\scriptstyle\,(\,0)}}$ & \boldmath$100{{\scriptstyle\,(\,0)}}$ & $39{{\scriptstyle\,(\,0)}}$ & $39{{\scriptstyle\,(\,0)}}$ \\
            \textbf{\rd} & $95{{\scriptstyle\,(\,3)}}$ & $53{{\scriptstyle\,(\,12)}}$ & $74{{\scriptstyle\,(\,9)}}$ & $45{{\scriptstyle\,(\,9)}}$ & $41{{\scriptstyle\,(\,7)}}$ \\
            \bottomrule
        \end{tabular}
        }
        \caption{FolkTexts}        
    \end{subtable}
    \hfill
    \begin{subtable}[t]{0.48\textwidth}
        \centering
        \resizebox{0.98\textwidth}{!}{ 
        \begin{tabular}{l c c c c c} 
            \toprule
            {} & \Gen $\uparrow$ & \Val $\uparrow$ & \ValH $\uparrow$ & \ED $\downarrow$ & \EDH  \\
            \midrule
            \textbf{\llamaS} & $86{{\scriptstyle\,(\,1)}}$ & $81{{\scriptstyle\,(\,0)}}$ & $72{{\scriptstyle\,(\,11)}}$ & \boldmath$76{{\scriptstyle\,(\,0)}}$ & \boldmath$71{{\scriptstyle\,(\,4)}}$ \\
            \textbf{\llamaM} & $100{{\scriptstyle\,(\,0)}}$ & \boldmath$89{{\scriptstyle\,(\,1)}}$ & \boldmath$75{{\scriptstyle\,(\,2)}}$ & $62{{\scriptstyle\,(\,1)}}$ & $62{{\scriptstyle\,(\,1)}}$ \\
            \textbf{\mistralS} & $95{{\scriptstyle\,(\,3)}}$ & \boldmath$79{{\scriptstyle\,(\,2)}}$ & \boldmath$91{{\scriptstyle\,(\,1)}}$ & $63{{\scriptstyle\,(\,1)}}$ & $63{{\scriptstyle\,(\,1)}}$ \\
            \textbf{\mistralM} & $100{{\scriptstyle\,(\,0)}}$ & \boldmath$82{{\scriptstyle\,(\,0)}}$ & \boldmath$100{{\scriptstyle\,(\,0)}}$ & $57{{\scriptstyle\,(\,0)}}$ & $57{{\scriptstyle\,(\,0)}}$ \\
            \textbf{\gemmaS} & $97{{\scriptstyle\,(\,0)}}$ & \boldmath$84{{\scriptstyle\,(\,0)}}$ & \boldmath$94{{\scriptstyle\,(\,1)}}$ & \boldmath$64{{\scriptstyle\,(\,0)}}$ & \boldmath$63{{\scriptstyle\,(\,0)}}$ \\
            \textbf{\gemmaM} & $100{{\scriptstyle\,(\,0)}}$ & \boldmath$76{{\scriptstyle\,(\,0)}}$ & \boldmath$90{{\scriptstyle\,(\,0)}}$ & $67{{\scriptstyle\,(\,0)}}$ & $67{{\scriptstyle\,(\,0)}}$ \\
            \textbf{\rd} & $100{{\scriptstyle\,(\,0)}}$ & $78{{\scriptstyle\,(\,1)}}$ & $88{{\scriptstyle\,(\,9)}}$ & $59{{\scriptstyle\,(\,2)}}$ & $58{{\scriptstyle\,(\,1)}}$ \\
            \bottomrule
        \end{tabular}
        }
        \caption{Twitter Financial News}
    \end{subtable}
    \hfill
    \begin{subtable}[t]{0.48\textwidth}
        \centering
        \resizebox{0.98\textwidth}{!}{ 
        \begin{tabular}{l c c c c c} 
            \toprule
            {} & \Gen $\uparrow$ & \Val $\uparrow$ & \ValH $\uparrow$ & \ED $\downarrow$ & \EDH  \\
            \midrule
            \textbf{\llamaS} & $85{{\scriptstyle\,(\,1)}}$ & \boldmath$59{{\scriptstyle\,(\,2)}}$ & \boldmath$48{{\scriptstyle\,(\,6)}}$ & $86{{\scriptstyle\,(\,1)}}$ & $84{{\scriptstyle\,(\,2)}}$ \\
            \textbf{\llamaM} & $99{{\scriptstyle\,(\,1)}}$ & \boldmath$92{{\scriptstyle\,(\,1)}}$ & \boldmath$55{{\scriptstyle\,(\,3)}}$ & \boldmath$68{{\scriptstyle\,(\,0)}}$ & \boldmath$70{{\scriptstyle\,(\,1)}}$ \\
            \textbf{\mistralS} & $90{{\scriptstyle\,(\,0)}}$ & $93{{\scriptstyle\,(\,0)}}$ & $93{{\scriptstyle\,(\,0)}}$ & $78{{\scriptstyle\,(\,1)}}$ & $78{{\scriptstyle\,(\,1)}}$ \\
            \textbf{\mistralM} & $100{{\scriptstyle\,(\,0)}}$ & $96{{\scriptstyle\,(\,1)}}$ & $96{{\scriptstyle\,(\,0)}}$ & $68{{\scriptstyle\,(\,0)}}$ & $68{{\scriptstyle\,(\,0)}}$ \\
            \textbf{\gemmaS} & $94{{\scriptstyle\,(\,1)}}$ & $97{{\scriptstyle\,(\,0)}}$ & $98{{\scriptstyle\,(\,1)}}$ & $76{{\scriptstyle\,(\,1)}}$ & $76{{\scriptstyle\,(\,2)}}$ \\
            \textbf{\gemmaM} & $100{{\scriptstyle\,(\,0)}}$ & \boldmath$99{{\scriptstyle\,(\,0)}}$ & \boldmath$90{{\scriptstyle\,(\,2)}}$ & $77{{\scriptstyle\,(\,0)}}$ & $77{{\scriptstyle\,(\,0)}}$ \\
            \textbf{\rd} & $99{{\scriptstyle\,(\,0)}}$ & \boldmath$94{{\scriptstyle\,(\,0)}}$ & \boldmath$78{{\scriptstyle\,(\,5)}}$ & $72{{\scriptstyle\,(\,2)}}$ & $70{{\scriptstyle\,(\,2)}}$ \\
            \bottomrule
        \end{tabular}
        }
        \caption{SST2}
    \end{subtable}
    \hfill
    \begin{subtable}[t]{0.48\textwidth}
        \centering
        \resizebox{0.98\textwidth}{!}{ 
        \begin{tabular}{l c c c c c} 
            \toprule
            {} & \Gen $\uparrow$ & \Val $\uparrow$ & \ValH $\uparrow$ & \ED $\downarrow$ & \EDH  \\
            \midrule
            \textbf{\llamaS} & $96{{\scriptstyle\,(\,1)}}$ & \boldmath$6{{\scriptstyle\,(\,1)}}$ & \boldmath$52{{\scriptstyle\,(\,2)}}$ & \boldmath$64{{\scriptstyle\,(\,3)}}$ & \boldmath$58{{\scriptstyle\,(\,0)}}$ \\
            \textbf{\llamaM} & $100{{\scriptstyle\,(\,0)}}$ & \boldmath$13{{\scriptstyle\,(\,1)}}$ & \boldmath$80{{\scriptstyle\,(\,9)}}$ & $57{{\scriptstyle\,(\,1)}}$ & $58{{\scriptstyle\,(\,0)}}$ \\
            \textbf{\mistralS} & $100{{\scriptstyle\,(\,0)}}$ & \boldmath$5{{\scriptstyle\,(\,1)}}$ & \boldmath$34{{\scriptstyle\,(\,4)}}$ & $57{{\scriptstyle\,(\,2)}}$ & $59{{\scriptstyle\,(\,1)}}$ \\
            \textbf{\mistralM} & $100{{\scriptstyle\,(\,0)}}$ & \boldmath$10{{\scriptstyle\,(\,0)}}$ & \boldmath$83{{\scriptstyle\,(\,0)}}$ & \boldmath$55{{\scriptstyle\,(\,0)}}$ & \boldmath$58{{\scriptstyle\,(\,0)}}$ \\
            \textbf{\gemmaS} & $27{{\scriptstyle\,(\,1)}}$ & \boldmath$3{{\scriptstyle\,(\,1)}}$ & \boldmath$48{{\scriptstyle\,(\,11)}}$ & $77{{\scriptstyle\,(\,6)}}$ & $74{{\scriptstyle\,(\,9)}}$ \\
            \textbf{\gemmaM} & $89{{\scriptstyle\,(\,1)}}$ & \boldmath$4{{\scriptstyle\,(\,0)}}$ & \boldmath$88{{\scriptstyle\,(\,3)}}$ & $57{{\scriptstyle\,(\,1)}}$ & $58{{\scriptstyle\,(\,0)}}$ \\
            \textbf{\rd} & $100{{\scriptstyle\,(\,0)}}$ & \boldmath$27{{\scriptstyle\,(\,3)}}$ & \boldmath$52{{\scriptstyle\,(\,5)}}$ & $69{{\scriptstyle\,(\,4)}}$ & $70{{\scriptstyle\,(\,7)}}$ \\
            \bottomrule
        \end{tabular}
        }
        \caption{GSM8K}        
    \end{subtable}
    \hfill
    \begin{subtable}[t]{0.48\textwidth}
        \centering
        \resizebox{0.98\textwidth}{!}{ 
        \begin{tabular}{l c c c c c} 
            \toprule
            {} & \Gen $\uparrow$ & \Val $\uparrow$ & \ValH $\uparrow$ & \ED $\downarrow$ & \EDH  \\
            \midrule
            \textbf{\llamaS} & $93{{\scriptstyle\,(\,0)}}$ & \boldmath$59{{\scriptstyle\,(\,1)}}$ & \boldmath$53{{\scriptstyle\,(\,2)}}$ & $73{{\scriptstyle\,(\,0)}}$ & $74{{\scriptstyle\,(\,1)}}$ \\
            \textbf{\llamaM} & $100{{\scriptstyle\,(\,0)}}$ & $88{{\scriptstyle\,(\,1)}}$ & $86{{\scriptstyle\,(\,6)}}$ & $72{{\scriptstyle\,(\,0)}}$ & $72{{\scriptstyle\,(\,0)}}$ \\
            \textbf{\mistralS} & $99{{\scriptstyle\,(\,0)}}$ & \boldmath$59{{\scriptstyle\,(\,1)}}$ & \boldmath$84{{\scriptstyle\,(\,0)}}$ & $74{{\scriptstyle\,(\,0)}}$ & $74{{\scriptstyle\,(\,0)}}$ \\
            \textbf{\mistralM} & $100{{\scriptstyle\,(\,0)}}$ & \boldmath$84{{\scriptstyle\,(\,0)}}$ & \boldmath$96{{\scriptstyle\,(\,1)}}$ & $78{{\scriptstyle\,(\,0)}}$ & $78{{\scriptstyle\,(\,0)}}$ \\
            \textbf{\gemmaS} & $97{{\scriptstyle\,(\,0)}}$ & \boldmath$78{{\scriptstyle\,(\,0)}}$ & \boldmath$86{{\scriptstyle\,(\,1)}}$ & $78{{\scriptstyle\,(\,0)}}$ & $78{{\scriptstyle\,(\,0)}}$ \\
            \textbf{\gemmaM} & $100{{\scriptstyle\,(\,0)}}$ & \boldmath$74{{\scriptstyle\,(\,1)}}$ & \boldmath$92{{\scriptstyle\,(\,0)}}$ & \boldmath$76{{\scriptstyle\,(\,0)}}$ & \boldmath$77{{\scriptstyle\,(\,0)}}$ \\
            \textbf{\rd} & $100{{\scriptstyle\,(\,0)}}$ & $77{{\scriptstyle\,(\,5)}}$ & $76{{\scriptstyle\,(\,14)}}$ & $78{{\scriptstyle\,(\,3)}}$ & $76{{\scriptstyle\,(\,1)}}$ \\
            \bottomrule
        \end{tabular}
        }
        \caption{MGNLI}
    \end{subtable}
    \caption{Performance of LLMs in generating \SCEs under unconstrained prompting at $T=0.5$, measured in terms of percentage of times the models are able to generate a \SCE (\Gen), percentage of times the model predictions on \SCEs yield the target label (\Val), and the normalized edit distance (\ED) between the original inputs and \SCEs. \ValH and \EDH denotes the metric values when the instructions for prediction on the original input and the \SCE generation are provided in the context while computing the validity of the \SCE (Section~\ref{sec:ce_eval}). Values in parentheses indicate marginal confidence intervals. See \autoref{sec:stat} for details. Values are bolded when the differences in with and without context conditions (\eg, \Val and \ValH) are statistically significant. $\uparrow$ means higher values are better.}
    \label{table:direct_prompting_temp05}
\end{table*}

\begin{table*}[!htb]
    \centering
    \begin{subtable}[t]{0.48\textwidth}
        \centering
        \resizebox{0.98\textwidth}{!}{ 
        \begin{tabular}{l c c c c c} 
            \toprule
            {} & \Gen $\uparrow$ & \Val $\uparrow$ & \ValH $\uparrow$ & \ED $\downarrow$ & \EDH  \\
            \midrule
            \textbf{\llamaS} & $81{{\scriptstyle\,(\,3)}}$ & \boldmath$55{{\scriptstyle\,(\,1)}}$ & \boldmath$84{{\scriptstyle\,(\,1)}}$ & $33{{\scriptstyle\,(\,3)}}$ & $33{{\scriptstyle\,(\,1)}}$ \\
            \textbf{\llamaM} & $100{{\scriptstyle\,(\,0)}}$ & $60{{\scriptstyle\,(\,1)}}$ & $67{{\scriptstyle\,(\,7)}}$ & \boldmath$25{{\scriptstyle\,(\,1)}}$ & \boldmath$22{{\scriptstyle\,(\,1)}}$ \\
            \textbf{\mistralS} & $99{{\scriptstyle\,(\,0)}}$ & \boldmath$88{{\scriptstyle\,(\,0)}}$ & \boldmath$91{{\scriptstyle\,(\,0)}}$ & $39{{\scriptstyle\,(\,1)}}$ & $38{{\scriptstyle\,(\,1)}}$ \\
            \textbf{\mistralM} & $100{{\scriptstyle\,(\,0)}}$ & \boldmath$59{{\scriptstyle\,(\,0)}}$ & \boldmath$83{{\scriptstyle\,(\,0)}}$ & \boldmath$12{{\scriptstyle\,(\,0)}}$ & \boldmath$11{{\scriptstyle\,(\,0)}}$ \\
            \textbf{\gemmaS} & $2{{\scriptstyle\,(\,2)}}$ & \boldmath$0{{\scriptstyle\,(\,0)}}$ & \boldmath$34{{\scriptstyle\,(\,27)}}$ & \boldmath$0{{\scriptstyle\,(\,0)}}$ & \boldmath$16{{\scriptstyle\,(\,0)}}$ \\
            \textbf{\gemmaM} & $81{{\scriptstyle\,(\,4)}}$ & \boldmath$47{{\scriptstyle\,(\,2)}}$ & \boldmath$98{{\scriptstyle\,(\,1)}}$ & $18{{\scriptstyle\,(\,1)}}$ & $17{{\scriptstyle\,(\,0)}}$ \\
            \textbf{\rd} & $100{{\scriptstyle\,(\,0)}}$ & \boldmath$62{{\scriptstyle\,(\,5)}}$ & \boldmath$87{{\scriptstyle\,(\,5)}}$ & \boldmath$23{{\scriptstyle\,(\,1)}}$ & \boldmath$21{{\scriptstyle\,(\,0)}}$ \\
            \bottomrule
        \end{tabular}
        }
        \caption{DiscrimEval}
    \end{subtable}
    \hfill
    \begin{subtable}[t]{0.48\textwidth}
        \centering
        \resizebox{0.98\textwidth}{!}{ 
        \begin{tabular}{l c c c c c} 
            \toprule
            {} & \Gen $\uparrow$ & \Val $\uparrow$ & \ValH $\uparrow$ & \ED $\downarrow$ & \EDH  \\
            \midrule
            \textbf{\llamaS} & $81{{\scriptstyle\,(\,10)}}$ & \boldmath$71{{\scriptstyle\,(\,0)}}$ & \boldmath$85{{\scriptstyle\,(\,1)}}$ & $37{{\scriptstyle\,(\,3)}}$ & $38{{\scriptstyle\,(\,4)}}$ \\
            \textbf{\llamaM} & $96{{\scriptstyle\,(\,2)}}$ & \boldmath$48{{\scriptstyle\,(\,3)}}$ & \boldmath$62{{\scriptstyle\,(\,5)}}$ & $36{{\scriptstyle\,(\,1)}}$ & $35{{\scriptstyle\,(\,0)}}$ \\
            \textbf{\mistralS} & $98{{\scriptstyle\,(\,0)}}$ & \boldmath$99{{\scriptstyle\,(\,0)}}$ & \boldmath$82{{\scriptstyle\,(\,2)}}$ & $48{{\scriptstyle\,(\,1)}}$ & $50{{\scriptstyle\,(\,1)}}$ \\
            \textbf{\mistralM} & $92{{\scriptstyle\,(\,0)}}$ & \boldmath$58{{\scriptstyle\,(\,0)}}$ & \boldmath$91{{\scriptstyle\,(\,0)}}$ & \boldmath$33{{\scriptstyle\,(\,0)}}$ & \boldmath$32{{\scriptstyle\,(\,0)}}$ \\
            \textbf{\gemmaS} & $8{{\scriptstyle\,(\,0)}}$ & \boldmath$4{{\scriptstyle\,(\,1)}}$ & \boldmath$92{{\scriptstyle\,(\,2)}}$ & \boldmath$43{{\scriptstyle\,(\,3)}}$ & \boldmath$33{{\scriptstyle\,(\,0)}}$ \\
            \textbf{\gemmaM} & $30{{\scriptstyle\,(\,3)}}$ & \boldmath$61{{\scriptstyle\,(\,6)}}$ & \boldmath$97{{\scriptstyle\,(\,0)}}$ & \boldmath$34{{\scriptstyle\,(\,0)}}$ & \boldmath$33{{\scriptstyle\,(\,0)}}$ \\
            \textbf{\rd} & $73{{\scriptstyle\,(\,15)}}$ & \boldmath$64{{\scriptstyle\,(\,0)}}$ & \boldmath$86{{\scriptstyle\,(\,7)}}$ & $40{{\scriptstyle\,(\,3)}}$ & $37{{\scriptstyle\,(\,3)}}$ \\
            \bottomrule
        \end{tabular}
        }
        \caption{FolkTexts}        
    \end{subtable}
     \hfill
    \begin{subtable}[t]{0.48\textwidth}
        \centering
        \resizebox{0.98\textwidth}{!}{ 
        \begin{tabular}{l c c c c c} 
            \toprule
            {} & \Gen $\uparrow$ & \Val $\uparrow$ & \ValH $\uparrow$ & \ED $\downarrow$ & \EDH  \\
            \midrule
            \textbf{\llamaS} & $85{{\scriptstyle\,(\,0)}}$ & $74{{\scriptstyle\,(\,1)}}$ & $81{{\scriptstyle\,(\,8)}}$ & \boldmath$59{{\scriptstyle\,(\,3)}}$ & \boldmath$54{{\scriptstyle\,(\,0)}}$ \\
            \textbf{\llamaM} & $99{{\scriptstyle\,(\,0)}}$ & \boldmath$92{{\scriptstyle\,(\,0)}}$ & \boldmath$73{{\scriptstyle\,(\,10)}}$ & $70{{\scriptstyle\,(\,3)}}$ & $67{{\scriptstyle\,(\,6)}}$ \\
            \textbf{\mistralS} & $100{{\scriptstyle\,(\,0)}}$ & \boldmath$90{{\scriptstyle\,(\,1)}}$ & \boldmath$96{{\scriptstyle\,(\,0)}}$ & $74{{\scriptstyle\,(\,0)}}$ & $74{{\scriptstyle\,(\,0)}}$ \\
            \textbf{\mistralM} & $100{{\scriptstyle\,(\,0)}}$ & \boldmath$77{{\scriptstyle\,(\,0)}}$ & \boldmath$99{{\scriptstyle\,(\,0)}}$ & \boldmath$49{{\scriptstyle\,(\,0)}}$ & \boldmath$48{{\scriptstyle\,(\,0)}}$ \\
            \textbf{\gemmaS} & $97{{\scriptstyle\,(\,0)}}$ & \boldmath$78{{\scriptstyle\,(\,0)}}$ & \boldmath$96{{\scriptstyle\,(\,0)}}$ & \boldmath$50{{\scriptstyle\,(\,0)}}$ & \boldmath$49{{\scriptstyle\,(\,0)}}$ \\
            \textbf{\gemmaM} & $100{{\scriptstyle\,(\,0)}}$ & \boldmath$87{{\scriptstyle\,(\,0)}}$ & \boldmath$92{{\scriptstyle\,(\,4)}}$ & $51{{\scriptstyle\,(\,1)}}$ & $49{{\scriptstyle\,(\,1)}}$ \\
            \textbf{\rd} & $100{{\scriptstyle\,(\,0)}}$ & $73{{\scriptstyle\,(\,2)}}$ & $80{{\scriptstyle\,(\,5)}}$ & $59{{\scriptstyle\,(\,3)}}$ & $58{{\scriptstyle\,(\,4)}}$ \\
            \bottomrule
        \end{tabular}
        }
        \caption{Twitter Financial News}
    \end{subtable}
    \hfill
    \begin{subtable}[t]{0.48\textwidth}
        \centering
        \resizebox{0.98\textwidth}{!}{ 
        \begin{tabular}{l c c c c c} 
            \toprule
            {} & \Gen $\uparrow$ & \Val $\uparrow$ & \ValH $\uparrow$ & \ED $\downarrow$ & \EDH  \\
            \midrule
            \textbf{\llamaS} & $87{{\scriptstyle\,(\,2)}}$ & \boldmath$49{{\scriptstyle\,(\,1)}}$ & \boldmath$58{{\scriptstyle\,(\,5)}}$ & \boldmath$73{{\scriptstyle\,(\,2)}}$ & \boldmath$69{{\scriptstyle\,(\,0)}}$ \\
            \textbf{\llamaM} & $99{{\scriptstyle\,(\,0)}}$ & \boldmath$87{{\scriptstyle\,(\,0)}}$ & \boldmath$67{{\scriptstyle\,(\,2)}}$ & $76{{\scriptstyle\,(\,1)}}$ & $77{{\scriptstyle\,(\,0)}}$ \\
            \textbf{\mistralS} & $85{{\scriptstyle\,(\,2)}}$ & \boldmath$93{{\scriptstyle\,(\,0)}}$ & \boldmath$89{{\scriptstyle\,(\,2)}}$ & $77{{\scriptstyle\,(\,1)}}$ & $77{{\scriptstyle\,(\,1)}}$ \\
            \textbf{\mistralM} & $100{{\scriptstyle\,(\,0)}}$ & \boldmath$85{{\scriptstyle\,(\,0)}}$ & \boldmath$98{{\scriptstyle\,(\,0)}}$ & \boldmath$66{{\scriptstyle\,(\,0)}}$ & \boldmath$65{{\scriptstyle\,(\,0)}}$ \\
            \textbf{\gemmaS} & $95{{\scriptstyle\,(\,1)}}$ & \boldmath$74{{\scriptstyle\,(\,2)}}$ & \boldmath$97{{\scriptstyle\,(\,0)}}$ & $66{{\scriptstyle\,(\,1)}}$ & $64{{\scriptstyle\,(\,1)}}$ \\
            \textbf{\gemmaM} & $100{{\scriptstyle\,(\,0)}}$ & \boldmath$83{{\scriptstyle\,(\,2)}}$ & \boldmath$95{{\scriptstyle\,(\,2)}}$ & $66{{\scriptstyle\,(\,1)}}$ & $65{{\scriptstyle\,(\,1)}}$ \\
            \textbf{\rd} & $99{{\scriptstyle\,(\,0)}}$ & \boldmath$77{{\scriptstyle\,(\,1)}}$ & \boldmath$72{{\scriptstyle\,(\,1)}}$ & $65{{\scriptstyle\,(\,1)}}$ & $63{{\scriptstyle\,(\,1)}}$ \\
            \bottomrule
        \end{tabular}
        }
        \caption{SST2}
    \end{subtable}
     \hfill
    \begin{subtable}[t]{0.48\textwidth}
        \centering
        \resizebox{0.98\textwidth}{!}{ 
        \begin{tabular}{l c c c c c} 
            \toprule
            {} & \Gen $\uparrow$ & \Val $\uparrow$ & \ValH $\uparrow$ & \ED $\downarrow$ & \EDH  \\
            \midrule
            \textbf{\llamaS} & $95{{\scriptstyle\,(\,1)}}$ & \boldmath$11{{\scriptstyle\,(\,0)}}$ & \boldmath$49{{\scriptstyle\,(\,7)}}$ & \boldmath$68{{\scriptstyle\,(\,1)}}$ & \boldmath$62{{\scriptstyle\,(\,3)}}$ \\
            \textbf{\llamaM} & $100{{\scriptstyle\,(\,0)}}$ & \boldmath$25{{\scriptstyle\,(\,1)}}$ & \boldmath$60{{\scriptstyle\,(\,2)}}$ & $63{{\scriptstyle\,(\,0)}}$ & $62{{\scriptstyle\,(\,1)}}$ \\
            \textbf{\mistralS} & $100{{\scriptstyle\,(\,0)}}$ & $57{{\scriptstyle\,(\,5)}}$ & $64{{\scriptstyle\,(\,6)}}$ & $59{{\scriptstyle\,(\,1)}}$ & $60{{\scriptstyle\,(\,1)}}$ \\
            \textbf{\mistralM} & $100{{\scriptstyle\,(\,0)}}$ & \boldmath$10{{\scriptstyle\,(\,0)}}$ & \boldmath$75{{\scriptstyle\,(\,0)}}$ & \boldmath$55{{\scriptstyle\,(\,0)}}$ & \boldmath$58{{\scriptstyle\,(\,0)}}$ \\
            \textbf{\gemmaS} & $30{{\scriptstyle\,(\,0)}}$ & \boldmath$6{{\scriptstyle\,(\,1)}}$ & \boldmath$48{{\scriptstyle\,(\,4)}}$ & $55{{\scriptstyle\,(\,3)}}$ & $57{{\scriptstyle\,(\,1)}}$ \\
            \textbf{\gemmaM} & $93{{\scriptstyle\,(\,2)}}$ & \boldmath$7{{\scriptstyle\,(\,0)}}$ & \boldmath$76{{\scriptstyle\,(\,1)}}$ & $57{{\scriptstyle\,(\,1)}}$ & $58{{\scriptstyle\,(\,1)}}$ \\
            \textbf{\rd} & $99{{\scriptstyle\,(\,0)}}$ & \boldmath$19{{\scriptstyle\,(\,0)}}$ & \boldmath$37{{\scriptstyle\,(\,6)}}$ & $63{{\scriptstyle\,(\,0)}}$ & $62{{\scriptstyle\,(\,4)}}$ \\
            \bottomrule
        \end{tabular}
        }
        \caption{GSM8K}        
    \end{subtable}
    \hfill
    \begin{subtable}[t]{0.48\textwidth}
        \centering
        \resizebox{0.98\textwidth}{!}{ 
        \begin{tabular}{l c c c c c} 
            \toprule
            {} & \Gen $\uparrow$ & \Val $\uparrow$ & \ValH $\uparrow$ & \ED $\downarrow$ & \EDH  \\
            \midrule
            \textbf{\llamaS} & $93{{\scriptstyle\,(\,0)}}$ & $61{{\scriptstyle\,(\,1)}}$ & $64{{\scriptstyle\,(\,11)}}$ & $77{{\scriptstyle\,(\,1)}}$ & $75{{\scriptstyle\,(\,1)}}$ \\
            \textbf{\llamaM} & $99{{\scriptstyle\,(\,0)}}$ & \boldmath$90{{\scriptstyle\,(\,1)}}$ & \boldmath$60{{\scriptstyle\,(\,20)}}$ & $74{{\scriptstyle\,(\,0)}}$ & $73{{\scriptstyle\,(\,1)}}$ \\
            \textbf{\mistralS} & $98{{\scriptstyle\,(\,2)}}$ & $89{{\scriptstyle\,(\,1)}}$ & $88{{\scriptstyle\,(\,4)}}$ & $73{{\scriptstyle\,(\,0)}}$ & $73{{\scriptstyle\,(\,0)}}$ \\
            \textbf{\mistralM} & $100{{\scriptstyle\,(\,0)}}$ & \boldmath$68{{\scriptstyle\,(\,0)}}$ & \boldmath$87{{\scriptstyle\,(\,0)}}$ & $75{{\scriptstyle\,(\,0)}}$ & $75{{\scriptstyle\,(\,0)}}$ \\
            \textbf{\gemmaS} & $91{{\scriptstyle\,(\,5)}}$ & \boldmath$66{{\scriptstyle\,(\,1)}}$ & \boldmath$84{{\scriptstyle\,(\,2)}}$ & $76{{\scriptstyle\,(\,0)}}$ & $76{{\scriptstyle\,(\,0)}}$ \\
            \textbf{\gemmaM} & $100{{\scriptstyle\,(\,0)}}$ & \boldmath$74{{\scriptstyle\,(\,1)}}$ & \boldmath$89{{\scriptstyle\,(\,3)}}$ & $75{{\scriptstyle\,(\,0)}}$ & $75{{\scriptstyle\,(\,0)}}$ \\
            \textbf{\rd} & $100{{\scriptstyle\,(\,0)}}$ & \boldmath$64{{\scriptstyle\,(\,2)}}$ & \boldmath$86{{\scriptstyle\,(\,1)}}$ & $73{{\scriptstyle\,(\,0)}}$ & $73{{\scriptstyle\,(\,0)}}$ \\
            \bottomrule
        \end{tabular}
        }
        \caption{MGNLI}
    \end{subtable}
    \caption{Performance of LLMs in generating \SCEs under rationale-based prompting at $T=0.5$, measured in terms of percentage of times the models are able to generate a \SCE (\Gen), percentage of times the model predictions on \SCEs yield the target label (\Val), and the normalized edit distance (\ED) between the original inputs and \SCEs. \ValH and \EDH denotes the metric values when the instructions for prediction on the original input and the \SCE generation are provided in the context while computing the validity of the \SCE (Section~\ref{sec:ce_eval}). Values in parentheses indicate marginal confidence intervals. See \autoref{sec:stat} for details. Values are bolded when the differences in with and without context conditions (\eg, \Val and \ValH) are statistically significant. $\uparrow$ means higher values are better.}
    \label{table:rationale_prompting_temp05}
\end{table*}

\begin{table*}[!htb]
    \centering
    \begin{subtable}[t]{0.48\textwidth}
        \centering
        \resizebox{0.98\textwidth}{!}{ 
        \begin{tabular}{l c c c c c} 
            \toprule
            {} & \Gen $\uparrow$ & \Val $\uparrow$ & \ValH $\uparrow$ & \ED $\downarrow$ & \EDH  \\
            \midrule
            \textbf{\llamaS} & $97{{\scriptstyle\,(\,4)}}$ & $84{{\scriptstyle\,(\,9)}}$ & $75{{\scriptstyle\,(\,10)}}$ & $52{{\scriptstyle\,(\,5)}}$ & $53{{\scriptstyle\,(\,5)}}$ \\
            \textbf{\llamaM} & $100{{\scriptstyle\,(\,0)}}$ & \boldmath$76{{\scriptstyle\,(\,10)}}$ & \boldmath$53{{\scriptstyle\,(\,12)}}$ & $34{{\scriptstyle\,(\,3)}}$ & $38{{\scriptstyle\,(\,4)}}$ \\
            \textbf{\mistralS} & $90{{\scriptstyle\,(\,7)}}$ & $86{{\scriptstyle\,(\,9)}}$ & $90{{\scriptstyle\,(\,7)}}$ & $37{{\scriptstyle\,(\,4)}}$ & $36{{\scriptstyle\,(\,4)}}$ \\
            \textbf{\mistralM} & $97{{\scriptstyle\,(\,4)}}$ & \boldmath$82{{\scriptstyle\,(\,9)}}$ & \boldmath$100{{\scriptstyle\,(\,0)}}$ & $24{{\scriptstyle\,(\,3)}}$ & $23{{\scriptstyle\,(\,3)}}$ \\
            \textbf{\gemmaS} & $89{{\scriptstyle\,(\,7)}}$ & \boldmath$63{{\scriptstyle\,(\,12)}}$ & \boldmath$94{{\scriptstyle\,(\,6)}}$ & $24{{\scriptstyle\,(\,3)}}$ & $23{{\scriptstyle\,(\,3)}}$ \\
            \textbf{\gemmaM} & $100{{\scriptstyle\,(\,0)}}$ & \boldmath$94{{\scriptstyle\,(\,6)}}$ & \boldmath$71{{\scriptstyle\,(\,11)}}$ & $22{{\scriptstyle\,(\,2)}}$ & $24{{\scriptstyle\,(\,3)}}$ \\
            \textbf{\rd} & $100{{\scriptstyle\,(\,0)}}$ & \boldmath$76{{\scriptstyle\,(\,10)}}$ & \boldmath$99{{\scriptstyle\,(\,2)}}$ & $37{{\scriptstyle\,(\,3)}}$ & $35{{\scriptstyle\,(\,3)}}$ \\
            \bottomrule
        \end{tabular}
        }
        \caption{DiscrimEval}
    \end{subtable}
    \hfill
    \begin{subtable}[t]{0.48\textwidth}
        \centering
        \resizebox{0.98\textwidth}{!}{ 
        \begin{tabular}{l c c c c c} 
            \toprule
            {} & \Gen $\uparrow$ & \Val $\uparrow$ & \ValH $\uparrow$ & \ED $\downarrow$ & \EDH  \\
            \midrule
            \textbf{\llamaS} & $99{{\scriptstyle\,(\,1)}}$ & \boldmath$80{{\scriptstyle\,(\,4)}}$ & \boldmath$96{{\scriptstyle\,(\,2)}}$ & $48{{\scriptstyle\,(\,2)}}$ & $46{{\scriptstyle\,(\,2)}}$ \\
            \textbf{\llamaM} & $99{{\scriptstyle\,(\,1)}}$ & \boldmath$84{{\scriptstyle\,(\,3)}}$ & \boldmath$64{{\scriptstyle\,(\,4)}}$ & $37{{\scriptstyle\,(\,1)}}$ & $37{{\scriptstyle\,(\,1)}}$ \\
            \textbf{\mistralS} & $82{{\scriptstyle\,(\,3)}}$ & \boldmath$85{{\scriptstyle\,(\,3)}}$ & \boldmath$99{{\scriptstyle\,(\,1)}}$ & $32{{\scriptstyle\,(\,1)}}$ & $30{{\scriptstyle\,(\,1)}}$ \\
            \textbf{\mistralM} & $100{{\scriptstyle\,(\,0)}}$ & \boldmath$54{{\scriptstyle\,(\,4)}}$ & \boldmath$98{{\scriptstyle\,(\,1)}}$ & $32{{\scriptstyle\,(\,0)}}$ & $32{{\scriptstyle\,(\,0)}}$ \\
            \textbf{\gemmaS} & $94{{\scriptstyle\,(\,2)}}$ & \boldmath$88{{\scriptstyle\,(\,3)}}$ & \boldmath$99{{\scriptstyle\,(\,1)}}$ & $40{{\scriptstyle\,(\,0)}}$ & $39{{\scriptstyle\,(\,0)}}$ \\
            \textbf{\gemmaM} & $100{{\scriptstyle\,(\,0)}}$ & \boldmath$99{{\scriptstyle\,(\,1)}}$ & \boldmath$100{{\scriptstyle\,(\,0)}}$ & $38{{\scriptstyle\,(\,0)}}$ & $38{{\scriptstyle\,(\,0)}}$ \\
            \textbf{\rd} & $99{{\scriptstyle\,(\,1)}}$ & \boldmath$75{{\scriptstyle\,(\,4)}}$ & \boldmath$40{{\scriptstyle\,(\,4)}}$ & \boldmath$62{{\scriptstyle\,(\,2)}}$ & \boldmath$57{{\scriptstyle\,(\,3)}}$ \\
            \bottomrule
        \end{tabular}
        }
        \caption{FolkTexts}        
    \end{subtable}
    \hfill
    \begin{subtable}[t]{0.48\textwidth}
        \centering
        \resizebox{0.98\textwidth}{!}{ 
        \begin{tabular}{l c c c c c} 
            \toprule
            {} & \Gen $\uparrow$ & \Val $\uparrow$ & \ValH $\uparrow$ & \ED $\downarrow$ & \EDH  \\
            \midrule
            \textbf{\llamaS} & $85{{\scriptstyle\,(\,3)}}$ & $85{{\scriptstyle\,(\,3)}}$ & $83{{\scriptstyle\,(\,3)}}$ & $77{{\scriptstyle\,(\,2)}}$ & $76{{\scriptstyle\,(\,2)}}$ \\
            \textbf{\llamaM} & $100{{\scriptstyle\,(\,0)}}$ & \boldmath$87{{\scriptstyle\,(\,2)}}$ & \boldmath$75{{\scriptstyle\,(\,3)}}$ & $60{{\scriptstyle\,(\,1)}}$ & $60{{\scriptstyle\,(\,1)}}$ \\
            \textbf{\mistralS} & $99{{\scriptstyle\,(\,1)}}$ & \boldmath$90{{\scriptstyle\,(\,2)}}$ & \boldmath$96{{\scriptstyle\,(\,1)}}$ & $64{{\scriptstyle\,(\,1)}}$ & $64{{\scriptstyle\,(\,1)}}$ \\
            \textbf{\mistralM} & $100{{\scriptstyle\,(\,0)}}$ & \boldmath$82{{\scriptstyle\,(\,3)}}$ & \boldmath$100{{\scriptstyle\,(\,0)}}$ & $61{{\scriptstyle\,(\,1)}}$ & $61{{\scriptstyle\,(\,1)}}$ \\
            \textbf{\gemmaS} & $98{{\scriptstyle\,(\,1)}}$ & \boldmath$84{{\scriptstyle\,(\,3)}}$ & \boldmath$96{{\scriptstyle\,(\,1)}}$ & $63{{\scriptstyle\,(\,1)}}$ & $62{{\scriptstyle\,(\,1)}}$ \\
            \textbf{\gemmaM} & $100{{\scriptstyle\,(\,0)}}$ & \boldmath$75{{\scriptstyle\,(\,3)}}$ & \boldmath$91{{\scriptstyle\,(\,2)}}$ & $67{{\scriptstyle\,(\,1)}}$ & $67{{\scriptstyle\,(\,1)}}$ \\
            \textbf{\rd} & $100{{\scriptstyle\,(\,0)}}$ & \boldmath$77{{\scriptstyle\,(\,3)}}$ & \boldmath$94{{\scriptstyle\,(\,2)}}$ & \boldmath$62{{\scriptstyle\,(\,1)}}$ & \boldmath$59{{\scriptstyle\,(\,1)}}$ \\
            \bottomrule
        \end{tabular}
        }
        \caption{Twitter Financial News}        
    \end{subtable}
    \hfill
    \begin{subtable}[t]{0.48\textwidth}
        \centering
        \resizebox{0.98\textwidth}{!}{ 
        \begin{tabular}{l c c c c c} 
            \toprule
            {} & \Gen $\uparrow$ & \Val $\uparrow$ & \ValH $\uparrow$ & \ED $\downarrow$ & \EDH  \\
            \midrule
            \textbf{\llamaS} & $93{{\scriptstyle\,(\,2)}}$ & $59{{\scriptstyle\,(\,4)}}$ & $53{{\scriptstyle\,(\,5)}}$ & $77{{\scriptstyle\,(\,2)}}$ & $78{{\scriptstyle\,(\,2)}}$ \\
            \textbf{\llamaM} & $94{{\scriptstyle\,(\,2)}}$ & \boldmath$92{{\scriptstyle\,(\,2)}}$ & \boldmath$58{{\scriptstyle\,(\,4)}}$ & $70{{\scriptstyle\,(\,2)}}$ & $72{{\scriptstyle\,(\,2)}}$ \\
            \textbf{\mistralS} & $89{{\scriptstyle\,(\,3)}}$ & \boldmath$92{{\scriptstyle\,(\,3)}}$ & \boldmath$80{{\scriptstyle\,(\,4)}}$ & $80{{\scriptstyle\,(\,1)}}$ & $80{{\scriptstyle\,(\,1)}}$ \\
            \textbf{\mistralM} & $96{{\scriptstyle\,(\,2)}}$ & $97{{\scriptstyle\,(\,2)}}$ & $96{{\scriptstyle\,(\,2)}}$ & $67{{\scriptstyle\,(\,1)}}$ & $66{{\scriptstyle\,(\,1)}}$ \\
            \textbf{\gemmaS} & $76{{\scriptstyle\,(\,4)}}$ & $93{{\scriptstyle\,(\,3)}}$ & $92{{\scriptstyle\,(\,3)}}$ & $72{{\scriptstyle\,(\,1)}}$ & $72{{\scriptstyle\,(\,1)}}$ \\
            \textbf{\gemmaM} & $98{{\scriptstyle\,(\,1)}}$ & \boldmath$99{{\scriptstyle\,(\,1)}}$ & \boldmath$80{{\scriptstyle\,(\,4)}}$ & $76{{\scriptstyle\,(\,1)}}$ & $76{{\scriptstyle\,(\,1)}}$ \\
            \textbf{\rd} & $100{{\scriptstyle\,(\,0)}}$ & \boldmath$91{{\scriptstyle\,(\,3)}}$ & \boldmath$77{{\scriptstyle\,(\,4)}}$ & $73{{\scriptstyle\,(\,1)}}$ & $72{{\scriptstyle\,(\,1)}}$ \\
            \bottomrule
        \end{tabular}
        }
        \caption{SST2}        
    \end{subtable}
    \hfill
    \begin{subtable}[t]{0.48\textwidth}
        \centering
        \resizebox{0.98\textwidth}{!}{ 
        \begin{tabular}{l c c c c c} 
            \toprule
            {} & \Gen $\uparrow$ & \Val $\uparrow$ & \ValH $\uparrow$ & \ED $\downarrow$ & \EDH  \\
            \midrule
            \textbf{\llamaS} & $95{{\scriptstyle\,(\,3)}}$ & \boldmath$5{{\scriptstyle\,(\,3)}}$ & \boldmath$53{{\scriptstyle\,(\,6)}}$ & $61{{\scriptstyle\,(\,7)}}$ & $59{{\scriptstyle\,(\,2)}}$ \\
            \textbf{\llamaM} & $100{{\scriptstyle\,(\,0)}}$ & \boldmath$14{{\scriptstyle\,(\,4)}}$ & \boldmath$72{{\scriptstyle\,(\,6)}}$ & $54{{\scriptstyle\,(\,3)}}$ & $58{{\scriptstyle\,(\,1)}}$ \\
            \textbf{\mistralS} & $100{{\scriptstyle\,(\,0)}}$ & \boldmath$10{{\scriptstyle\,(\,4)}}$ & \boldmath$39{{\scriptstyle\,(\,6)}}$ & $56{{\scriptstyle\,(\,5)}}$ & $57{{\scriptstyle\,(\,2)}}$ \\
            \textbf{\mistralM} & $100{{\scriptstyle\,(\,0)}}$ & \boldmath$14{{\scriptstyle\,(\,4)}}$ & \boldmath$84{{\scriptstyle\,(\,5)}}$ & $56{{\scriptstyle\,(\,3)}}$ & $58{{\scriptstyle\,(\,1)}}$ \\
            \textbf{\gemmaS} & $13{{\scriptstyle\,(\,4)}}$ & $12{{\scriptstyle\,(\,11)}}$ & $27{{\scriptstyle\,(\,15)}}$ & $61{{\scriptstyle\,(\,18)}}$ & $66{{\scriptstyle\,(\,12)}}$ \\
            \textbf{\gemmaM} & $96{{\scriptstyle\,(\,2)}}$ & \boldmath$4{{\scriptstyle\,(\,2)}}$ & \boldmath$86{{\scriptstyle\,(\,4)}}$ & $55{{\scriptstyle\,(\,5)}}$ & $58{{\scriptstyle\,(\,1)}}$ \\
            \textbf{\rd} & $100{{\scriptstyle\,(\,0)}}$ & \boldmath$26{{\scriptstyle\,(\,5)}}$ & \boldmath$63{{\scriptstyle\,(\,6)}}$ & \boldmath$73{{\scriptstyle\,(\,3)}}$ & \boldmath$83{{\scriptstyle\,(\,3)}}$ \\
            \bottomrule
        \end{tabular}
        }
        \caption{GSM8K}        
    \end{subtable}
    \hfill
    \begin{subtable}[t]{0.48\textwidth}
        \centering
        \resizebox{0.98\textwidth}{!}{ 
        \begin{tabular}{l c c c c c} 
            \toprule
            {} & \Gen $\uparrow$ & \Val $\uparrow$ & \ValH $\uparrow$ & \ED $\downarrow$ & \EDH  \\
            \midrule
            \textbf{\llamaS} & $95{{\scriptstyle\,(\,2)}}$ & \boldmath$56{{\scriptstyle\,(\,4)}}$ & \boldmath$79{{\scriptstyle\,(\,3)}}$ & $73{{\scriptstyle\,(\,1)}}$ & $73{{\scriptstyle\,(\,1)}}$ \\
            \textbf{\llamaM} & $97{{\scriptstyle\,(\,1)}}$ & \boldmath$81{{\scriptstyle\,(\,3)}}$ & \boldmath$73{{\scriptstyle\,(\,3)}}$ & $71{{\scriptstyle\,(\,1)}}$ & $71{{\scriptstyle\,(\,1)}}$ \\
            \textbf{\mistralS} & $100{{\scriptstyle\,(\,0)}}$ & \boldmath$62{{\scriptstyle\,(\,3)}}$ & \boldmath$82{{\scriptstyle\,(\,3)}}$ & $74{{\scriptstyle\,(\,1)}}$ & $74{{\scriptstyle\,(\,1)}}$ \\
            \textbf{\mistralM} & $100{{\scriptstyle\,(\,0)}}$ & \boldmath$85{{\scriptstyle\,(\,3)}}$ & \boldmath$96{{\scriptstyle\,(\,1)}}$ & $76{{\scriptstyle\,(\,1)}}$ & $76{{\scriptstyle\,(\,1)}}$ \\
            \textbf{\gemmaS} & $97{{\scriptstyle\,(\,1)}}$ & \boldmath$76{{\scriptstyle\,(\,3)}}$ & \boldmath$89{{\scriptstyle\,(\,2)}}$ & $77{{\scriptstyle\,(\,1)}}$ & $77{{\scriptstyle\,(\,1)}}$ \\
            \textbf{\gemmaM} & $100{{\scriptstyle\,(\,0)}}$ & \boldmath$85{{\scriptstyle\,(\,3)}}$ & \boldmath$98{{\scriptstyle\,(\,1)}}$ & $75{{\scriptstyle\,(\,1)}}$ & $75{{\scriptstyle\,(\,1)}}$ \\
            \textbf{\rd} & $100{{\scriptstyle\,(\,0)}}$ & $79{{\scriptstyle\,(\,3)}}$ & $84{{\scriptstyle\,(\,3)}}$ & $77{{\scriptstyle\,(\,1)}}$ & $76{{\scriptstyle\,(\,1)}}$ \\
            \bottomrule
        \end{tabular}
        }
        \caption{MGNLI}        
    \end{subtable}
    \caption{Performance of LLMs in generating \SCEs under CoT prompting at $T=0$, measured in terms of percentage of times the models are able to generate a \SCE (\Gen),  percentage of times the model predictions on \SCEs yield the target label (\Val), and the normalized edit distance (\ED) between the original inputs and \SCEs. \ValH and \EDH denote the metric values when the instructions for prediction on the original input and the \SCE generation are provided in the context while computing the validity of the \SCE (Section~\ref{sec:ce_eval}). Values in parentheses indicate marginal confidence intervals. See \autoref{sec:stat} for details. Values are bolded when the differences in with and without context conditions (\eg, \Val and \ValH) are statistically significant. $\uparrow$ means higher values are better.}
    \label{table:COT_prompting_temp0}
\end{table*}

\begin{table*}[!htb]
    \centering
    \begin{subtable}[t]{0.48\textwidth}
        \centering
        \resizebox{0.98\textwidth}{!}{ 
        \begin{tabular}{l c c c c c} 
            \toprule
            {} & \Gen $\uparrow$ & \Val $\uparrow$ & \ValH $\uparrow$ & \ED $\downarrow$ & \EDH  \\
            \midrule
            \textbf{\llamaS} & $89{{\scriptstyle\,(\,7)}}$ & $63{{\scriptstyle\,(\,12)}}$ & $81{{\scriptstyle\,(\,10)}}$ & $39{{\scriptstyle\,(\,6)}}$ & $42{{\scriptstyle\,(\,5)}}$ \\
            \textbf{\llamaM} & $99{{\scriptstyle\,(\,2)}}$ & \boldmath$84{{\scriptstyle\,(\,9)}}$ & \boldmath$55{{\scriptstyle\,(\,12)}}$ & $35{{\scriptstyle\,(\,4)}}$ & $37{{\scriptstyle\,(\,5)}}$ \\
            \textbf{\mistralS} & $91{{\scriptstyle\,(\,7)}}$ & $81{{\scriptstyle\,(\,10)}}$ & $88{{\scriptstyle\,(\,8)}}$ & $40{{\scriptstyle\,(\,4)}}$ & $37{{\scriptstyle\,(\,3)}}$ \\
            \textbf{\mistralM} & $97{{\scriptstyle\,(\,4)}}$ & \boldmath$78{{\scriptstyle\,(\,10)}}$ & \boldmath$97{{\scriptstyle\,(\,4)}}$ & $25{{\scriptstyle\,(\,3)}}$ & $24{{\scriptstyle\,(\,3)}}$ \\
            \textbf{\gemmaS} & $77{{\scriptstyle\,(\,10)}}$ & \boldmath$59{{\scriptstyle\,(\,13)}}$ & \boldmath$91{{\scriptstyle\,(\,8)}}$ & $25{{\scriptstyle\,(\,3)}}$ & $23{{\scriptstyle\,(\,2)}}$ \\
            \textbf{\gemmaM} & $100{{\scriptstyle\,(\,0)}}$ & $83{{\scriptstyle\,(\,9)}}$ & $86{{\scriptstyle\,(\,8)}}$ & $25{{\scriptstyle\,(\,3)}}$ & $25{{\scriptstyle\,(\,2)}}$ \\
            \textbf{\rd} & $93{{\scriptstyle\,(\,6)}}$ & \boldmath$75{{\scriptstyle\,(\,11)}}$ & \boldmath$100{{\scriptstyle\,(\,0)}}$ & $41{{\scriptstyle\,(\,5)}}$ & $41{{\scriptstyle\,(\,5)}}$ \\
            \bottomrule
        \end{tabular}
        }
        \caption{DiscrimEval}
    \end{subtable}
    \hfill
    \begin{subtable}[t]{0.48\textwidth}
        \centering
        \resizebox{0.98\textwidth}{!}{ 
        \begin{tabular}{l c c c c c} 
            \toprule
            {} & \Gen $\uparrow$ & \Val $\uparrow$ & \ValH $\uparrow$ & \ED $\downarrow$ & \EDH  \\
            \midrule
            \textbf{\llamaS} & $92{{\scriptstyle\,(\,2)}}$ & \boldmath$72{{\scriptstyle\,(\,4)}}$ & \boldmath$82{{\scriptstyle\,(\,4)}}$ & $48{{\scriptstyle\,(\,3)}}$ & $47{{\scriptstyle\,(\,2)}}$ \\
            \textbf{\llamaM} & $97{{\scriptstyle\,(\,2)}}$ & \boldmath$80{{\scriptstyle\,(\,4)}}$ & \boldmath$66{{\scriptstyle\,(\,4)}}$ & $38{{\scriptstyle\,(\,1)}}$ & $37{{\scriptstyle\,(\,1)}}$ \\
            \textbf{\mistralS} & $76{{\scriptstyle\,(\,4)}}$ & \boldmath$83{{\scriptstyle\,(\,4)}}$ & \boldmath$92{{\scriptstyle\,(\,3)}}$ & $34{{\scriptstyle\,(\,1)}}$ & $33{{\scriptstyle\,(\,1)}}$ \\
            \textbf{\mistralM} & $100{{\scriptstyle\,(\,0)}}$ & \boldmath$65{{\scriptstyle\,(\,4)}}$ & \boldmath$98{{\scriptstyle\,(\,1)}}$ & $34{{\scriptstyle\,(\,0)}}$ & $33{{\scriptstyle\,(\,0)}}$ \\
            \textbf{\gemmaS} & $82{{\scriptstyle\,(\,3)}}$ & \boldmath$81{{\scriptstyle\,(\,4)}}$ & \boldmath$97{{\scriptstyle\,(\,2)}}$ & $41{{\scriptstyle\,(\,1)}}$ & $39{{\scriptstyle\,(\,1)}}$ \\
            \textbf{\gemmaM} & $99{{\scriptstyle\,(\,1)}}$ & \boldmath$99{{\scriptstyle\,(\,1)}}$ & \boldmath$100{{\scriptstyle\,(\,0)}}$ & $39{{\scriptstyle\,(\,0)}}$ & $39{{\scriptstyle\,(\,0)}}$ \\
            \textbf{\rd} & $67{{\scriptstyle\,(\,4)}}$ & \boldmath$50{{\scriptstyle\,(\,5)}}$ & \boldmath$88{{\scriptstyle\,(\,3)}}$ & $38{{\scriptstyle\,(\,2)}}$ & $36{{\scriptstyle\,(\,2)}}$ \\
            \bottomrule
        \end{tabular}
        }
        \caption{FolkTexts}
    \end{subtable}
    \hfill
    \begin{subtable}[t]{0.48\textwidth}
        \centering
        \resizebox{0.98\textwidth}{!}{ 
        \begin{tabular}{l c c c c c} 
            \toprule
            {} & \Gen $\uparrow$ & \Val $\uparrow$ & \ValH $\uparrow$ & \ED $\downarrow$ & \EDH  \\
            \midrule
            \textbf{\llamaS} & $86{{\scriptstyle\,(\,2)}}$ & $80{{\scriptstyle\,(\,3)}}$ & $82{{\scriptstyle\,(\,3)}}$ & $76{{\scriptstyle\,(\,2)}}$ & $75{{\scriptstyle\,(\,2)}}$ \\
            \textbf{\llamaM} & $100{{\scriptstyle\,(\,0)}}$ & \boldmath$87{{\scriptstyle\,(\,2)}}$ & \boldmath$78{{\scriptstyle\,(\,3)}}$ & $61{{\scriptstyle\,(\,1)}}$ & $61{{\scriptstyle\,(\,1)}}$ \\
            \textbf{\mistralS} & $91{{\scriptstyle\,(\,2)}}$ & \boldmath$81{{\scriptstyle\,(\,3)}}$ & \boldmath$92{{\scriptstyle\,(\,2)}}$ & $64{{\scriptstyle\,(\,1)}}$ & $64{{\scriptstyle\,(\,1)}}$ \\
            \textbf{\mistralM} & $100{{\scriptstyle\,(\,0)}}$ & \boldmath$81{{\scriptstyle\,(\,3)}}$ & \boldmath$100{{\scriptstyle\,(\,0)}}$ & $58{{\scriptstyle\,(\,1)}}$ & $57{{\scriptstyle\,(\,1)}}$ \\
            \textbf{\gemmaS} & $97{{\scriptstyle\,(\,1)}}$ & \boldmath$87{{\scriptstyle\,(\,2)}}$ & \boldmath$95{{\scriptstyle\,(\,2)}}$ & $63{{\scriptstyle\,(\,1)}}$ & $63{{\scriptstyle\,(\,1)}}$ \\
            \textbf{\gemmaM} & $100{{\scriptstyle\,(\,0)}}$ & \boldmath$74{{\scriptstyle\,(\,3)}}$ & \boldmath$91{{\scriptstyle\,(\,2)}}$ & $67{{\scriptstyle\,(\,1)}}$ & $67{{\scriptstyle\,(\,1)}}$ \\
            \textbf{\rd} & $99{{\scriptstyle\,(\,1)}}$ & \boldmath$77{{\scriptstyle\,(\,3)}}$ & \boldmath$91{{\scriptstyle\,(\,2)}}$ & \boldmath$62{{\scriptstyle\,(\,1)}}$ & \boldmath$59{{\scriptstyle\,(\,1)}}$ \\
            \bottomrule
        \end{tabular}
        }
        \caption{Twitter Financial News}        
    \end{subtable}
    \hfill
    \begin{subtable}[t]{0.48\textwidth}
        \centering
        \resizebox{0.98\textwidth}{!}{ 
        \begin{tabular}{l c c c c c} 
            \toprule
            {} & \Gen $\uparrow$ & \Val $\uparrow$ & \ValH $\uparrow$ & \ED $\downarrow$ & \EDH  \\
            \midrule
            \textbf{\llamaS} & $92{{\scriptstyle\,(\,2)}}$ & $59{{\scriptstyle\,(\,4)}}$ & $53{{\scriptstyle\,(\,5)}}$ & $79{{\scriptstyle\,(\,2)}}$ & $79{{\scriptstyle\,(\,2)}}$ \\
            \textbf{\llamaM} & $95{{\scriptstyle\,(\,2)}}$ & \boldmath$87{{\scriptstyle\,(\,3)}}$ & \boldmath$54{{\scriptstyle\,(\,4)}}$ & $70{{\scriptstyle\,(\,2)}}$ & $72{{\scriptstyle\,(\,2)}}$ \\
            \textbf{\mistralS} & $87{{\scriptstyle\,(\,3)}}$ & \boldmath$92{{\scriptstyle\,(\,3)}}$ & \boldmath$78{{\scriptstyle\,(\,4)}}$ & $80{{\scriptstyle\,(\,1)}}$ & $80{{\scriptstyle\,(\,1)}}$ \\
            \textbf{\mistralM} & $96{{\scriptstyle\,(\,2)}}$ & $93{{\scriptstyle\,(\,2)}}$ & $89{{\scriptstyle\,(\,3)}}$ & $69{{\scriptstyle\,(\,1)}}$ & $68{{\scriptstyle\,(\,1)}}$ \\
            \textbf{\gemmaS} & $70{{\scriptstyle\,(\,4)}}$ & $89{{\scriptstyle\,(\,3)}}$ & $93{{\scriptstyle\,(\,3)}}$ & $73{{\scriptstyle\,(\,1)}}$ & $73{{\scriptstyle\,(\,1)}}$ \\
            \textbf{\gemmaM} & $98{{\scriptstyle\,(\,1)}}$ & \boldmath$97{{\scriptstyle\,(\,2)}}$ & \boldmath$81{{\scriptstyle\,(\,4)}}$ & $77{{\scriptstyle\,(\,1)}}$ & $77{{\scriptstyle\,(\,1)}}$ \\
            \textbf{\rd} & $98{{\scriptstyle\,(\,1)}}$ & \boldmath$85{{\scriptstyle\,(\,3)}}$ & \boldmath$72{{\scriptstyle\,(\,4)}}$ & $75{{\scriptstyle\,(\,1)}}$ & $75{{\scriptstyle\,(\,2)}}$ \\
            \bottomrule
        \end{tabular}
        }
        \caption{SST2}        
    \end{subtable}
    \hfill
    \begin{subtable}[t]{0.48\textwidth}
        \centering
        \resizebox{0.98\textwidth}{!}{ 
        \begin{tabular}{l c c c c c} 
            \toprule
            {} & \Gen $\uparrow$ & \Val $\uparrow$ & \ValH $\uparrow$ & \ED $\downarrow$ & \EDH  \\
            \midrule
            \textbf{\llamaS} & $92{{\scriptstyle\,(\,3)}}$ & \boldmath$4{{\scriptstyle\,(\,3)}}$ & \boldmath$58{{\scriptstyle\,(\,6)}}$ & $55{{\scriptstyle\,(\,11)}}$ & $57{{\scriptstyle\,(\,2)}}$ \\
            \textbf{\llamaM} & $99{{\scriptstyle\,(\,1)}}$ & \boldmath$18{{\scriptstyle\,(\,5)}}$ & \boldmath$63{{\scriptstyle\,(\,6)}}$ & $57{{\scriptstyle\,(\,4)}}$ & $59{{\scriptstyle\,(\,2)}}$ \\
            \textbf{\mistralS} & $99{{\scriptstyle\,(\,1)}}$ & \boldmath$8{{\scriptstyle\,(\,3)}}$ & \boldmath$36{{\scriptstyle\,(\,6)}}$ & $56{{\scriptstyle\,(\,5)}}$ & $60{{\scriptstyle\,(\,2)}}$ \\
            \textbf{\mistralM} & $99{{\scriptstyle\,(\,1)}}$ & \boldmath$6{{\scriptstyle\,(\,3)}}$ & \boldmath$82{{\scriptstyle\,(\,5)}}$ & $59{{\scriptstyle\,(\,5)}}$ & $59{{\scriptstyle\,(\,1)}}$ \\
            \textbf{\gemmaS} & $28{{\scriptstyle\,(\,6)}}$ & \boldmath$3{{\scriptstyle\,(\,4)}}$ & \boldmath$39{{\scriptstyle\,(\,11)}}$ & $76{{\scriptstyle\,(\,45)}}$ & $76{{\scriptstyle\,(\,9)}}$ \\
            \textbf{\gemmaM} & $96{{\scriptstyle\,(\,2)}}$ & \boldmath$3{{\scriptstyle\,(\,2)}}$ & \boldmath$84{{\scriptstyle\,(\,5)}}$ & $58{{\scriptstyle\,(\,8)}}$ & $58{{\scriptstyle\,(\,1)}}$ \\
            \textbf{\rd} & $100{{\scriptstyle\,(\,0)}}$ & \boldmath$27{{\scriptstyle\,(\,6)}}$ & \boldmath$54{{\scriptstyle\,(\,6)}}$ & $75{{\scriptstyle\,(\,3)}}$ & $73{{\scriptstyle\,(\,3)}}$ \\
            \bottomrule
        \end{tabular}
        }
        \caption{GSM8K}        
    \end{subtable}
    \hfill
    \begin{subtable}[t]{0.48\textwidth}
        \centering
        \resizebox{0.98\textwidth}{!}{ 
        \begin{tabular}{l c c c c c} 
            \toprule
            {} & \Gen $\uparrow$ & \Val $\uparrow$ & \ValH $\uparrow$ & \ED $\downarrow$ & \EDH  \\
            \midrule
            \textbf{\llamaS} & $91{{\scriptstyle\,(\,2)}}$ & \boldmath$56{{\scriptstyle\,(\,4)}}$ & \boldmath$76{{\scriptstyle\,(\,3)}}$ & $76{{\scriptstyle\,(\,1)}}$ & $75{{\scriptstyle\,(\,1)}}$ \\
            \textbf{\llamaM} & $99{{\scriptstyle\,(\,1)}}$ & \boldmath$84{{\scriptstyle\,(\,3)}}$ & \boldmath$75{{\scriptstyle\,(\,3)}}$ & $73{{\scriptstyle\,(\,1)}}$ & $72{{\scriptstyle\,(\,1)}}$ \\
            \textbf{\mistralS} & $99{{\scriptstyle\,(\,1)}}$ & \boldmath$61{{\scriptstyle\,(\,4)}}$ & \boldmath$83{{\scriptstyle\,(\,3)}}$ & $73{{\scriptstyle\,(\,1)}}$ & $73{{\scriptstyle\,(\,1)}}$ \\
            \textbf{\mistralM} & $99{{\scriptstyle\,(\,1)}}$ & \boldmath$86{{\scriptstyle\,(\,2)}}$ & \boldmath$97{{\scriptstyle\,(\,1)}}$ & $77{{\scriptstyle\,(\,1)}}$ & $76{{\scriptstyle\,(\,1)}}$ \\
            \textbf{\gemmaS} & $93{{\scriptstyle\,(\,2)}}$ & \boldmath$77{{\scriptstyle\,(\,3)}}$ & \boldmath$92{{\scriptstyle\,(\,2)}}$ & $77{{\scriptstyle\,(\,1)}}$ & $77{{\scriptstyle\,(\,1)}}$ \\
            \textbf{\gemmaM} & $100{{\scriptstyle\,(\,0)}}$ & \boldmath$85{{\scriptstyle\,(\,3)}}$ & \boldmath$97{{\scriptstyle\,(\,1)}}$ & $76{{\scriptstyle\,(\,1)}}$ & $76{{\scriptstyle\,(\,1)}}$ \\
            \textbf{\rd} & $97{{\scriptstyle\,(\,1)}}$ & \boldmath$78{{\scriptstyle\,(\,3)}}$ & \boldmath$84{{\scriptstyle\,(\,3)}}$ & $78{{\scriptstyle\,(\,1)}}$ & $77{{\scriptstyle\,(\,1)}}$ \\            
            \bottomrule
        \end{tabular}
        }
        \caption{MGNLI}        
    \end{subtable}
    \caption{Performance of LLMs in generating \SCEs under CoT prompting at $T=0.5$, measured in terms of percentage of times the models are able to generate a \SCE (\Gen), percentage of times the model predictions on \SCEs yield the target label (\Val), and the normalized edit distance (\ED) between the original inputs and \SCEs. \ValH and \EDH denote the metric values when the instructions for prediction on the original input and the \SCE generation are provided in the context while computing the validity of the \SCE (Section~\ref{sec:ce_eval}). Values in parentheses indicate marginal confidence intervals. See \autoref{sec:stat} for details. Values are bolded when the differences in with and without context conditions (\eg, \Val and \ValH) are statistically significant. $\uparrow$ means higher values are better.}
    \label{table:COT_prompting_temp05}
\end{table*}

\begin{table*}[ht]
\small
\centering
\resizebox{0.98\textwidth}{!}{
\begin{tabular}{l 
                c@{\hspace{3pt}}c 
                c@{\hspace{3pt}}c 
                c@{\hspace{3pt}}c 
                c@{\hspace{3pt}}c 
                c@{\hspace{3pt}}c 
                c@{\hspace{3pt}}c}
\toprule
{} & \multicolumn{2}{c}{\textbf{DEV}} & \multicolumn{2}{c}{\textbf{TWT}} & \multicolumn{2}{c}{\textbf{SST}} & \multicolumn{2}{c}{\textbf{FLK}} & \multicolumn{2}{c}{\textbf{NLI}} & \multicolumn{2}{c}{\textbf{MTH}} \\
{} & w/o & w/ & w/o & w/ & w/o & w/ & w/o & w/ & w/o & w/ & w/o & w/ \\
\midrule
\textbf{\llamaS} & $40{{\scriptstyle\,(\,19)}}$ & $19{{\scriptstyle\,(\,30)}}$ & $\bm{6{{\scriptstyle\,(\,7)}}}$ & $\bm{44{{\scriptstyle\,(\,6)}}}$ & $37{{\scriptstyle\,(\,8)}}$ & $20{{\scriptstyle\,(\,9)}}$ & $13{{\scriptstyle\,(\,10)}}$ & $4{{\scriptstyle\,(\,2)}}$ & $1{{\scriptstyle\,(\,22)}}$ & $21{{\scriptstyle\,(\,20)}}$ & $26{{\scriptstyle\,(\,30)}}$ & $45{{\scriptstyle\,(\,13)}}$ \\
\textbf{\llamaM} & $\bm{16{{\scriptstyle\,(\,11)}}}$ & $\bm{67{{\scriptstyle\,(\,2)}}}$ & $5{{\scriptstyle\,(\,6)}}$ & $11{{\scriptstyle\,(\,5)}}$ & $26{{\scriptstyle\,(\,11)}}$ & $20{{\scriptstyle\,(\,8)}}$ & $\bm{0{{\scriptstyle\,(\,0)}}}$ & $\bm{100{{\scriptstyle\,(\,0)}}}$ & $\bm{0{{\scriptstyle\,(\,5)}}}$ & $\bm{15{{\scriptstyle\,(\,5)}}}$ & $\bm{22{{\scriptstyle\,(\,9)}}}$ & $\bm{100{{\scriptstyle\,(\,0)}}}$ \\
\textbf{\mistralS} & $4{{\scriptstyle\,(\,6)}}$ & $14{{\scriptstyle\,(\,6)}}$ & $\bm{1{{\scriptstyle\,(\,7)}}}$ & $\bm{19{{\scriptstyle\,(\,5)}}}$ & $27{{\scriptstyle\,(\,6)}}$ & $26{{\scriptstyle\,(\,8)}}$ & $\bm{3{{\scriptstyle\,(\,1)}}}$ & $\bm{9{{\scriptstyle\,(\,1)}}}$ & $5{{\scriptstyle\,(\,5)}}$ & $9{{\scriptstyle\,(\,5)}}$ & $9{{\scriptstyle\,(\,16)}}$ & $18{{\scriptstyle\,(\,18)}}$ \\
\textbf{\mistralM} & $\bm{19{{\scriptstyle\,(\,6)}}}$ & $\bm{100{{\scriptstyle\,(\,0)}}}$ & $3{{\scriptstyle\,(\,3)}}$ & $4{{\scriptstyle\,(\,3)}}$ & $\bm{8{{\scriptstyle\,(\,6)}}}$ & $\bm{27{{\scriptstyle\,(\,5)}}}$ & $\bm{1{{\scriptstyle\,(\,0)}}}$ & $\bm{2{{\scriptstyle\,(\,0)}}}$ & $\bm{3{{\scriptstyle\,(\,5)}}}$ & $\bm{16{{\scriptstyle\,(\,6)}}}$ & $19{{\scriptstyle\,(\,10)}}$ & $28{{\scriptstyle\,(\,4)}}$ \\
\textbf{\gemmaS} & $0{{\scriptstyle\,(\,0)}}$ & $0{{\scriptstyle\,(\,0)}}$ & $4{{\scriptstyle\,(\,4)}}$ & $6{{\scriptstyle\,(\,4)}}$ & $100{{\scriptstyle\,(\,0)}}$ & $100{{\scriptstyle\,(\,0)}}$ & $0{{\scriptstyle\,(\,0)}}$ & $0{{\scriptstyle\,(\,0)}}$ & $6{{\scriptstyle\,(\,4)}}$ & $7{{\scriptstyle\,(\,5)}}$ & $17{{\scriptstyle\,(\,26)}}$ & $11{{\scriptstyle\,(\,18)}}$ \\
\textbf{\gemmaM} & $\bm{11{{\scriptstyle\,(\,6)}}}$ & $\bm{100{{\scriptstyle\,(\,0)}}}$ & $3{{\scriptstyle\,(\,4)}}$ & $7{{\scriptstyle\,(\,3)}}$ & $\bm{6{{\scriptstyle\,(\,5)}}}$ & $\bm{49{{\scriptstyle\,(\,3)}}}$ & $\bm{4{{\scriptstyle\,(\,0)}}}$ & $\bm{100{{\scriptstyle\,(\,0)}}}$ & $1{{\scriptstyle\,(\,5)}}$ & $6{{\scriptstyle\,(\,5)}}$ & $\bm{31{{\scriptstyle\,(\,15)}}}$ & $\bm{9{{\scriptstyle\,(\,5)}}}$ \\
\textbf{\rd} & $\bm{16{{\scriptstyle\,(\,22)}}}$ & $\bm{100{{\scriptstyle\,(\,0)}}}$ & $37{{\scriptstyle\,(\,15)}}$ & $44{{\scriptstyle\,(\,5)}}$ & $\bm{35{{\scriptstyle\,(\,18)}}}$ & $\bm{72{{\scriptstyle\,(\,8)}}}$ & $\bm{1{{\scriptstyle\,(\,7)}}}$ & $\bm{26{{\scriptstyle\,(\,5)}}}$ & $11{{\scriptstyle\,(\,4)}}$ & $12{{\scriptstyle\,(\,4)}}$ & $63{{\scriptstyle\,(\,9)}}$ & $70{{\scriptstyle\,(\,9)}}$ \\
\bottomrule
\end{tabular}
}
\caption{Normalized difference in lengths of valid and invalid counterfactuals. For DiscrimEval (\DSEV), Twitter Financial News (\TWTR), SST2 (\SST), FolkTexts (\FOLK), MGNLI (\NLI), and GSM8K (\MATH) datasets under unconstrained prompting with $T = 0$. Left columns (w/o) show the differences without prediction and counterfactual generations provided as context (Section~\ref{sec:ce_eval}), whereas right columns (w/) show the differences with this information.}
\label{table:response_len_diff_direct_prompting_temp0}
\end{table*}

\begin{table*}[ht]
\small
\centering
\resizebox{0.98\textwidth}{!}{
\begin{tabular}{l 
                c@{\hspace{3pt}}c 
                c@{\hspace{3pt}}c 
                c@{\hspace{3pt}}c 
                c@{\hspace{3pt}}c 
                c@{\hspace{3pt}}c 
                c@{\hspace{3pt}}c}
\toprule
{} & \multicolumn{2}{c}{\textbf{DEV}} & \multicolumn{2}{c}{\textbf{TWT}} & \multicolumn{2}{c}{\textbf{SST}} & \multicolumn{2}{c}{\textbf{FLK}} & \multicolumn{2}{c}{\textbf{NLI}} & \multicolumn{2}{c}{\textbf{MTH}} \\
{} & w/o & w/ & w/o & w/ & w/o & w/ & w/o & w/ & w/o & w/ & w/o & w/ \\
\midrule
\textbf{\llamaS} & $\bm{23{{\scriptstyle\,(\,14)}}}$ & $\bm{52{{\scriptstyle\,(\,7)}}}$ & $80{{\scriptstyle\,(\,3)}}$ & $81{{\scriptstyle\,(\,3)}}$ & $18{{\scriptstyle\,(\,15)}}$ & $1{{\scriptstyle\,(\,17)}}$ & $2{{\scriptstyle\,(\,9)}}$ & $8{{\scriptstyle\,(\,9)}}$ & $25{{\scriptstyle\,(\,17)}}$ & $46{{\scriptstyle\,(\,10)}}$ & $46{{\scriptstyle\,(\,16)}}$ & $40{{\scriptstyle\,(\,11)}}$ \\
\textbf{\llamaM} & $1{{\scriptstyle\,(\,10)}}$ & $7{{\scriptstyle\,(\,9)}}$ & $5{{\scriptstyle\,(\,6)}}$ & $4{{\scriptstyle\,(\,5)}}$ & $29{{\scriptstyle\,(\,14)}}$ & $38{{\scriptstyle\,(\,11)}}$ & $\bm{5{{\scriptstyle\,(\,2)}}}$ & $\bm{12{{\scriptstyle\,(\,2)}}}$ & $10{{\scriptstyle\,(\,12)}}$ & $1{{\scriptstyle\,(\,10)}}$ & $12{{\scriptstyle\,(\,13)}}$ & $0{{\scriptstyle\,(\,6)}}$ \\
\textbf{\mistralS} & $\bm{2{{\scriptstyle\,(\,6)}}}$ & $\bm{100{{\scriptstyle\,(\,0)}}}$ & $\bm{1{{\scriptstyle\,(\,7)}}}$ & $\bm{21{{\scriptstyle\,(\,6)}}}$ & $17{{\scriptstyle\,(\,10)}}$ & $7{{\scriptstyle\,(\,11)}}$ & $6{{\scriptstyle\,(\,3)}}$ & $2{{\scriptstyle\,(\,3)}}$ & $19{{\scriptstyle\,(\,7)}}$ & $17{{\scriptstyle\,(\,7)}}$ & $13{{\scriptstyle\,(\,14)}}$ & $24{{\scriptstyle\,(\,7)}}$ \\
\textbf{\mistralM} & $\bm{2{{\scriptstyle\,(\,7)}}}$ & $\bm{100{{\scriptstyle\,(\,0)}}}$ & $\bm{6{{\scriptstyle\,(\,3)}}}$ & $\bm{100{{\scriptstyle\,(\,0)}}}$ & $\bm{10{{\scriptstyle\,(\,6)}}}$ & $\bm{40{{\scriptstyle\,(\,5)}}}$ & $0{{\scriptstyle\,(\,0)}}$ & $1{{\scriptstyle\,(\,0)}}$ & $\bm{4{{\scriptstyle\,(\,5)}}}$ & $\bm{14{{\scriptstyle\,(\,5)}}}$ & $17{{\scriptstyle\,(\,11)}}$ & $13{{\scriptstyle\,(\,5)}}$ \\
\textbf{\gemmaS} & $16{{\scriptstyle\,(\,7)}}$ & $10{{\scriptstyle\,(\,5)}}$ & $4{{\scriptstyle\,(\,4)}}$ & $6{{\scriptstyle\,(\,4)}}$ & $\bm{10{{\scriptstyle\,(\,9)}}}$ & $\bm{34{{\scriptstyle\,(\,7)}}}$ & $\bm{3{{\scriptstyle\,(\,1)}}}$ & $\bm{22{{\scriptstyle\,(\,3)}}}$ & $6{{\scriptstyle\,(\,6)}}$ & $7{{\scriptstyle\,(\,6)}}$ & $10{{\scriptstyle\,(\,25)}}$ & $11{{\scriptstyle\,(\,27)}}$ \\
\textbf{\gemmaM} & $12{{\scriptstyle\,(\,6)}}$ & $15{{\scriptstyle\,(\,6)}}$ & $3{{\scriptstyle\,(\,4)}}$ & $7{{\scriptstyle\,(\,3)}}$ & $\bm{25{{\scriptstyle\,(\,5)}}}$ & $\bm{43{{\scriptstyle\,(\,4)}}}$ & $\bm{0{{\scriptstyle\,(\,1)}}}$ & $\bm{100{{\scriptstyle\,(\,0)}}}$ & $9{{\scriptstyle\,(\,6)}}$ & $16{{\scriptstyle\,(\,5)}}$ & $\bm{20{{\scriptstyle\,(\,23)}}}$ & $\bm{100{{\scriptstyle\,(\,0)}}}$ \\
\textbf{\rd} & $\bm{6{{\scriptstyle\,(\,10)}}}$ & $\bm{55{{\scriptstyle\,(\,4)}}}$ & $\bm{37{{\scriptstyle\,(\,15)}}}$ & $\bm{93{{\scriptstyle\,(\,1)}}}$ & $31{{\scriptstyle\,(\,20)}}$ & $33{{\scriptstyle\,(\,21)}}$ & $\bm{1{{\scriptstyle\,(\,7)}}}$ & $\bm{26{{\scriptstyle\,(\,5)}}}$ & $\bm{17{{\scriptstyle\,(\,15)}}}$ & $\bm{81{{\scriptstyle\,(\,3)}}}$ & $63{{\scriptstyle\,(\,8)}}$ & $48{{\scriptstyle\,(\,12)}}$ \\
\bottomrule
\end{tabular}
}
\caption{Normalized difference in lengths of valid and invalid counterfactuals. For DiscrimEval (\DSEV), Twitter Financial News (\TWTR), SST2 (\SST), FolkTexts (\FOLK), MGNLI (\NLI), and GSM8K (\MATH) datasets under CoT prompting with $T = 0$. Left columns (w/o) show the differences without prediction and counterfactual generations provided as context (Section~\ref{sec:ce_eval}), whereas right columns (w/) show the differences with this information.}
\label{table:response_len_diff_CoT_prompting_temp0}
\end{table*}

\begin{table*}[ht]
\small
    \centering
    \begin{subtable}{0.48\textwidth}
        \centering
        \resizebox{0.98\textwidth}{!}{
        \begin{tabular}{l c c c c c c} 
            \toprule
            {} & \textbf{\DSEV} & \textbf{\TWTR} & \textbf{\SST} & \textbf{\FOLK} & \textbf{\NLI} & \textbf{\MATH}\\
            \midrule
            \textbf{\llamaS} & $54{{\scriptstyle\,(\,12)}}$ & $77{{\scriptstyle\,(\,3)}}$ & $82{{\scriptstyle\,(\,3)}}$ & $55{{\scriptstyle\,(\,4)}}$ & $66{{\scriptstyle\,(\,3)}}$ & $13{{\scriptstyle\,(\,4)}}$ \\
            \textbf{\llamaM} & $86{{\scriptstyle\,(\,8)}}$ & $80{{\scriptstyle\,(\,3)}}$ & $92{{\scriptstyle\,(\,2)}}$ & $69{{\scriptstyle\,(\,4)}}$ & $76{{\scriptstyle\,(\,3)}}$ & $39{{\scriptstyle\,(\,6)}}$ \\
            \textbf{\mistralS} & $82{{\scriptstyle\,(\,9)}}$ & $82{{\scriptstyle\,(\,3)}}$ & $60{{\scriptstyle\,(\,4)}}$ & $60{{\scriptstyle\,(\,4)}}$ & $75{{\scriptstyle\,(\,3)}}$ & $8{{\scriptstyle\,(\,3)}}$ \\
            \textbf{\mistralM} & $63{{\scriptstyle\,(\,11)}}$ & $84{{\scriptstyle\,(\,3)}}$ & $81{{\scriptstyle\,(\,3)}}$ & $71{{\scriptstyle\,(\,4)}}$ & $86{{\scriptstyle\,(\,2)}}$ & $38{{\scriptstyle\,(\,6)}}$ \\
            \textbf{\gemmaS} & $80{{\scriptstyle\,(\,9)}}$ & $81{{\scriptstyle\,(\,3)}}$ & $90{{\scriptstyle\,(\,3)}}$ & $76{{\scriptstyle\,(\,4)}}$ & $77{{\scriptstyle\,(\,3)}}$ & $24{{\scriptstyle\,(\,5)}}$ \\
            \textbf{\gemmaM} & $76{{\scriptstyle\,(\,10)}}$ & $85{{\scriptstyle\,(\,3)}}$ & $91{{\scriptstyle\,(\,2)}}$ & $74{{\scriptstyle\,(\,4)}}$ & $82{{\scriptstyle\,(\,3)}}$ & $0{{\scriptstyle\,(\,1)}}$ \\
            \textbf{\rd} & $39{{\scriptstyle\,(\,11)}}$ & $79{{\scriptstyle\,(\,3)}}$ & $95{{\scriptstyle\,(\,2)}}$ & $30{{\scriptstyle\,(\,4)}}$ & $82{{\scriptstyle\,(\,3)}}$ & $13{{\scriptstyle\,(\,4)}}$ \\
            \bottomrule
        \end{tabular}
        }
        \caption{Accuracy under unconstrained and rationale-based prompting ($T = 0$)}
    \end{subtable}
    \hfill
 \begin{subtable}{0.48\textwidth}
        \centering
        \resizebox{0.98\textwidth}{!}{
        \begin{tabular}{l c c c c c c} 
            \toprule
            {} & \textbf{\DSEV} & \textbf{\TWTR} & \textbf{\SST} & \textbf{\FOLK} & \textbf{\NLI} & \textbf{\MATH}\\
            \midrule
            \textbf{\llamaS} & $51{{\scriptstyle\,(\,12)}}$ & $77{{\scriptstyle\,(\,3)}}$ & $83{{\scriptstyle\,(\,3)}}$ & $55{{\scriptstyle\,(\,4)}}$ & $65{{\scriptstyle\,(\,3)}}$ & $12{{\scriptstyle\,(\,4)}}$ \\
            \textbf{\llamaM} & $85{{\scriptstyle\,(\,8)}}$ & $82{{\scriptstyle\,(\,3)}}$ & $92{{\scriptstyle\,(\,2)}}$ & $70{{\scriptstyle\,(\,4)}}$ & $76{{\scriptstyle\,(\,3)}}$ & $40{{\scriptstyle\,(\,6)}}$ \\
            \textbf{\mistralS} & $80{{\scriptstyle\,(\,9)}}$ & $81{{\scriptstyle\,(\,3)}}$ & $61{{\scriptstyle\,(\,4)}}$ & $59{{\scriptstyle\,(\,4)}}$ & $76{{\scriptstyle\,(\,3)}}$ & $8{{\scriptstyle\,(\,3)}}$ \\
            \textbf{\mistralM} & $68{{\scriptstyle\,(\,11)}}$ & $82{{\scriptstyle\,(\,3)}}$ & $81{{\scriptstyle\,(\,3)}}$ & $69{{\scriptstyle\,(\,4)}}$ & $84{{\scriptstyle\,(\,3)}}$ & $41{{\scriptstyle\,(\,6)}}$ \\
            \textbf{\gemmaS} & $80{{\scriptstyle\,(\,9)}}$ & $81{{\scriptstyle\,(\,3)}}$ & $90{{\scriptstyle\,(\,3)}}$ & $75{{\scriptstyle\,(\,4)}}$ & $78{{\scriptstyle\,(\,3)}}$ & $22{{\scriptstyle\,(\,5)}}$ \\
            \textbf{\gemmaM} & $79{{\scriptstyle\,(\,10)}}$ & $85{{\scriptstyle\,(\,3)}}$ & $90{{\scriptstyle\,(\,3)}}$ & $74{{\scriptstyle\,(\,4)}}$ & $82{{\scriptstyle\,(\,3)}}$ & $27{{\scriptstyle\,(\,6)}}$ \\
            \textbf{\rd} & $46{{\scriptstyle\,(\,12)}}$ & $79{{\scriptstyle\,(\,3)}}$ & $94{{\scriptstyle\,(\,2)}}$ & $36{{\scriptstyle\,(\,4)}}$ & $78{{\scriptstyle\,(\,3)}}$ & $19{{\scriptstyle\,(\,5)}}$ \\
            \bottomrule
        \end{tabular}
        }
        \caption{Accuracy under unconstrained and rationale-based prompting ($T = 0.5$)}
    \end{subtable}
    
\vspace{3mm}

    \begin{subtable}{0.48\textwidth}
        \centering
        \resizebox{0.98\textwidth}{!}{
    \begin{tabular}{l c c c c c c} 
        \toprule
        {} & \textbf{\DSEV} & \textbf{\TWTR} & \textbf{\SST} & \textbf{\FOLK} & \textbf{\NLI} & \textbf{\MATH}\\
        \midrule
        \textbf{\llamaS} & $85{{\scriptstyle\,(\,8)}}$ & $75{{\scriptstyle\,(\,3)}}$ & $93{{\scriptstyle\,(\,2)}}$ & $68{{\scriptstyle\,(\,4)}}$ & $62{{\scriptstyle\,(\,3)}}$ & $86{{\scriptstyle\,(\,4)}}$ \\
        \textbf{\llamaM} & $84{{\scriptstyle\,(\,9)}}$ & $78{{\scriptstyle\,(\,3)}}$ & $96{{\scriptstyle\,(\,2)}}$ & $52{{\scriptstyle\,(\,5)}}$ & $78{{\scriptstyle\,(\,3)}}$ & $29{{\scriptstyle\,(\,6)}}$ \\
        \textbf{\mistralS} & $63{{\scriptstyle\,(\,11)}}$ & $76{{\scriptstyle\,(\,3)}}$ & $78{{\scriptstyle\,(\,4)}}$ & $31{{\scriptstyle\,(\,4)}}$ & $63{{\scriptstyle\,(\,3)}}$ & $11{{\scriptstyle\,(\,4)}}$ \\
        \textbf{\mistralM} & $66{{\scriptstyle\,(\,11)}}$ & $78{{\scriptstyle\,(\,3)}}$ & $91{{\scriptstyle\,(\,2)}}$ & $72{{\scriptstyle\,(\,4)}}$ & $80{{\scriptstyle\,(\,3)}}$ & $96{{\scriptstyle\,(\,2)}}$ \\
        \textbf{\gemmaS} & $72{{\scriptstyle\,(\,10)}}$ & $79{{\scriptstyle\,(\,3)}}$ & $86{{\scriptstyle\,(\,3)}}$ & $67{{\scriptstyle\,(\,4)}}$ & $77{{\scriptstyle\,(\,3)}}$ & $61{{\scriptstyle\,(\,6)}}$ \\
        \textbf{\gemmaM} & $69{{\scriptstyle\,(\,11)}}$ & $81{{\scriptstyle\,(\,3)}}$ & $82{{\scriptstyle\,(\,3)}}$ & $69{{\scriptstyle\,(\,4)}}$ & $76{{\scriptstyle\,(\,3)}}$ & $29{{\scriptstyle\,(\,6)}}$ \\
        \textbf{\rd} & $17{{\scriptstyle\,(\,9)}}$ & $72{{\scriptstyle\,(\,3)}}$ & $94{{\scriptstyle\,(\,2)}}$ & $13{{\scriptstyle\,(\,3)}}$ & $76{{\scriptstyle\,(\,3)}}$ & $31{{\scriptstyle\,(\,6)}}$ \\
        \bottomrule
    \end{tabular}
    }
    \caption{Accuracy under CoT prompting ($T = 0$)}
     \end{subtable}
    \hfill
    \begin{subtable}{0.48\textwidth}
        \centering
        \resizebox{0.98\textwidth}{!}{
    \begin{tabular}{l c c c c c c}
        \toprule
        {} & \textbf{\DSEV} & \textbf{\TWTR} & \textbf{\SST} & \textbf{\FOLK} & \textbf{\NLI} & \textbf{\MATH}\\
        \midrule
        \textbf{\llamaS} & $83{{\scriptstyle\,(\,9)}}$ & $75{{\scriptstyle\,(\,3)}}$ & $92{{\scriptstyle\,(\,2)}}$ & $65{{\scriptstyle\,(\,4)}}$ & $62{{\scriptstyle\,(\,3)}}$ & $82{{\scriptstyle\,(\,5)}}$ \\
        \textbf{\llamaM} & $89{{\scriptstyle\,(\,7)}}$ & $80{{\scriptstyle\,(\,3)}}$ & $96{{\scriptstyle\,(\,2)}}$ & $61{{\scriptstyle\,(\,5)}}$ & $80{{\scriptstyle\,(\,3)}}$ & $98{{\scriptstyle\,(\,2)}}$ \\
        \textbf{\mistralS} & $62{{\scriptstyle\,(\,11)}}$ & $75{{\scriptstyle\,(\,3)}}$ & $80{{\scriptstyle\,(\,3)}}$ & $38{{\scriptstyle\,(\,4)}}$ & $62{{\scriptstyle\,(\,3)}}$ & $12{{\scriptstyle\,(\,4)}}$ \\
        \textbf{\mistralM} & $66{{\scriptstyle\,(\,11)}}$ & $80{{\scriptstyle\,(\,3)}}$ & $90{{\scriptstyle\,(\,3)}}$ & $73{{\scriptstyle\,(\,4)}}$ & $80{{\scriptstyle\,(\,3)}}$ & $94{{\scriptstyle\,(\,3)}}$ \\
        \textbf{\gemmaS} & $72{{\scriptstyle\,(\,10)}}$ & $79{{\scriptstyle\,(\,3)}}$ & $85{{\scriptstyle\,(\,3)}}$ & $69{{\scriptstyle\,(\,4)}}$ & $77{{\scriptstyle\,(\,3)}}$ & $64{{\scriptstyle\,(\,6)}}$ \\
        \textbf{\gemmaM} & $66{{\scriptstyle\,(\,11)}}$ & $78{{\scriptstyle\,(\,3)}}$ & $83{{\scriptstyle\,(\,3)}}$ & $69{{\scriptstyle\,(\,4)}}$ & $73{{\scriptstyle\,(\,3)}}$ & $27{{\scriptstyle\,(\,6)}}$ \\
        \textbf{\rd} & $15{{\scriptstyle\,(\,8)}}$ & $68{{\scriptstyle\,(\,3)}}$ & $93{{\scriptstyle\,(\,2)}}$ & $17{{\scriptstyle\,(\,3)}}$ & $65{{\scriptstyle\,(\,3)}}$ & $34{{\scriptstyle\,(\,6)}}$ \\
        \bottomrule
    \end{tabular}
    }
    \caption{Accuracy under CoT prompting ($T = 0.5$)}
\end{subtable}
    \caption{Task-specific accuracy (\%) of models on each dataset under (a) $T = 0$ and (b) $T = 0.5$. Since the prompts used for unconstrained and rationale-based generations are identical when obtaining model predictions, their accuracy values are shared. However, because CoT uses a different prompt format, we independently report its accuracy. Values in parentheses indicate marginal confidence intervals. See \autoref{sec:stat} for details.}
    \label{tables:acc}
    \end{table*}

\section{Statistical Analysis of Results}
\label{sec:stat}
We computed 95\% Confidence Intervals (CIs) for generation percentage, validity percentage, and edit distance to assess whether the differences between the \textit{with context} and \textit{without context} conditions are statistically significant. Non-overlapping CIs mean that the results for the two conditions differ more than what we would expect just from random variation. This usually points to a statistically significant difference (roughly corresponding to $p < 0.05$). The CIs were calculated using the standard error of the mean:

\[
\text{CI} = \text{mean} \pm 1.96 \times \left( \frac{\text{sd}}{\sqrt{n}} \right)
\]
Here, \textit{mean} is the average value, \textit{sd} is the standard deviation, and $n$ is the number of samples. The factor $1.96$ corresponds to a 95\% confidence level under a normal distribution.

\section{Correlation between validity and popular performance metrics}
\label{app:correlation}
We explored the relationship between the validity of \textsc{\SCEs} and several model properties, including \textit{Model Size}, \textit{Perplexity}, and \textit{Open LLM Leaderboard Rank}\footnote{\url{https://huggingface.co/spaces/open-llm-leaderboard}} (see \autoref{fig:rank_vs_plp}).
However, we did not observe any clear or consistent patterns. Additionally, we performed both Pearson and Spearman correlation tests to check for non-zero correlation coefficient,%
\footnote{Using \url{https://scipy.org}}
but \textbf{none of the correlations were statistically significant, with all p-values exceeding $0.05$}. In the following subsection, we present results from these analyses under unconstrained prompting with temperature $T=0$.

\begin{figure}[!htb]
    \centering
    \includegraphics[width=\linewidth]{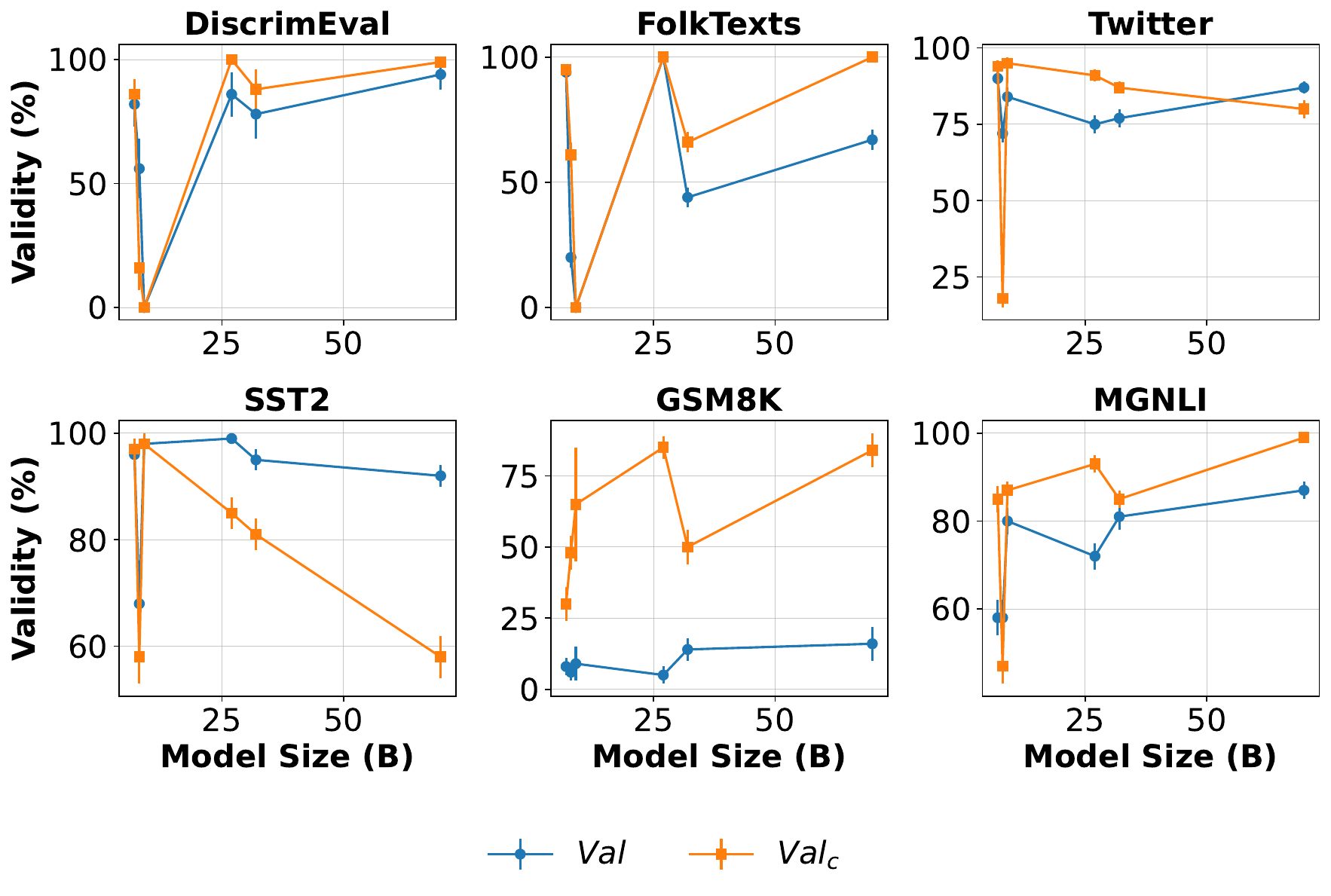}
    \caption{Validity of \SCEs vs. Model Size across Datasets. Orange lines show validity with context (\ValH); blue lines show validity without context (\Val).}
    \label{fig:model_size_vs_validity}
\end{figure}

\xhdr{Validity of \SCEs vs. Model Size across Datasets}
\autoref{fig:model_size_vs_validity} shows how \SCE validity varies with model size across datasets. Scaling generally improves validity on some tasks (e.g., \textsc{DiscrimEval}, \textsc{FolkTexts}, \textsc{MGNLI}), but yields diminishing returns or even declines on others (\textsc{Twitter}, \textsc{SST2}) and remains poor on \textsc{GSM8K}. Notably, smaller models sometimes outperform larger ones (e.g., \textsc{SST2}, \textsc{GSM8K}), indicating that counterfactual validity does not scale monotonically with model size.

\begin{figure}[ht]
    \centering
    \includegraphics[width=\linewidth]{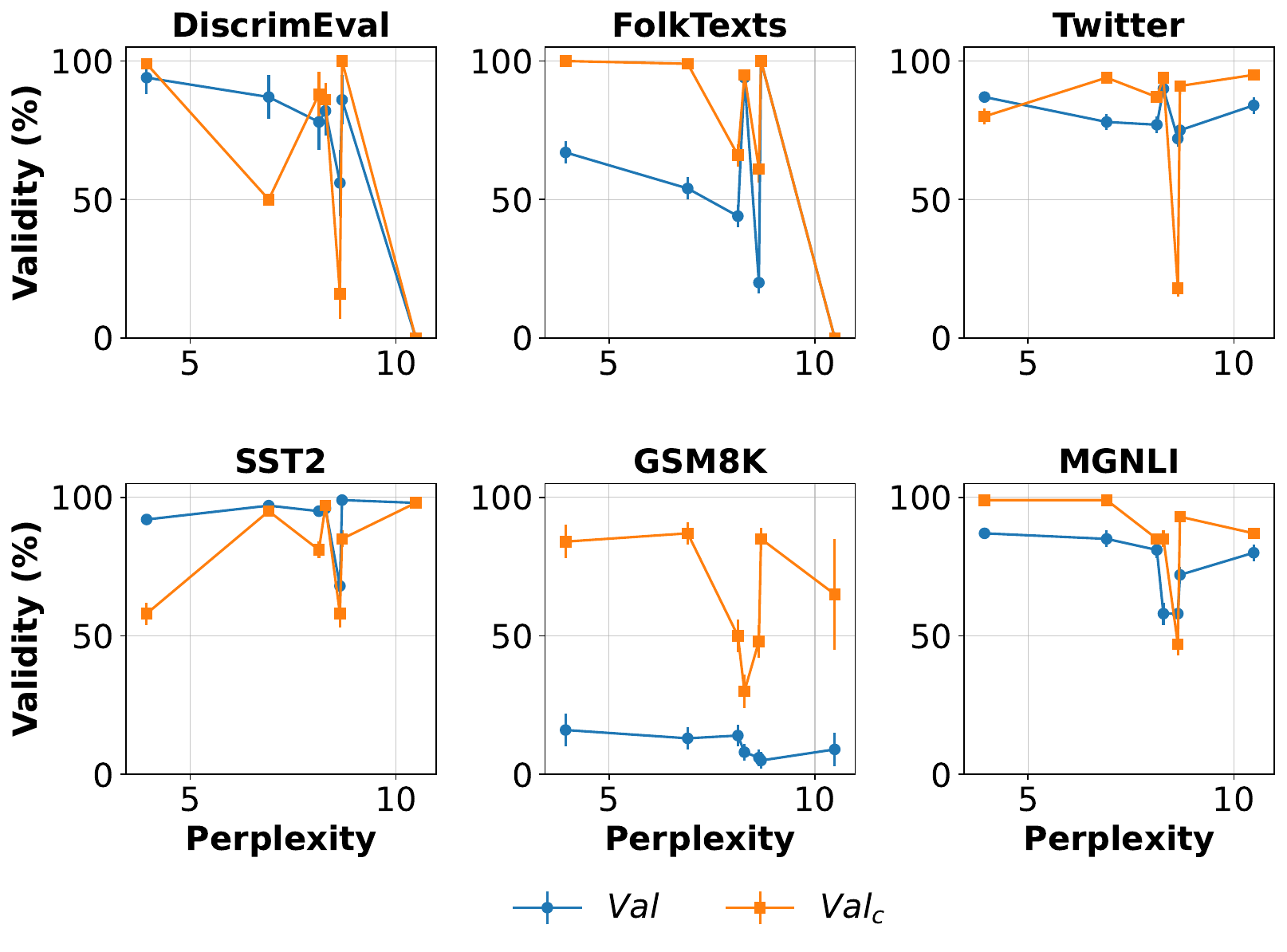}
    \caption{Line plots of few-shot perplexity (measured on \textsc{Wikitext}) versus \SCE validity across datasets. Blue lines indicate validity without context (\Val) and orange lines indicate validity with context (\ValH).}
    \label{fig:model_plp_vs_validity}
\end{figure}

\begin{figure}[ht]
    \centering
    \includegraphics[width=\linewidth]{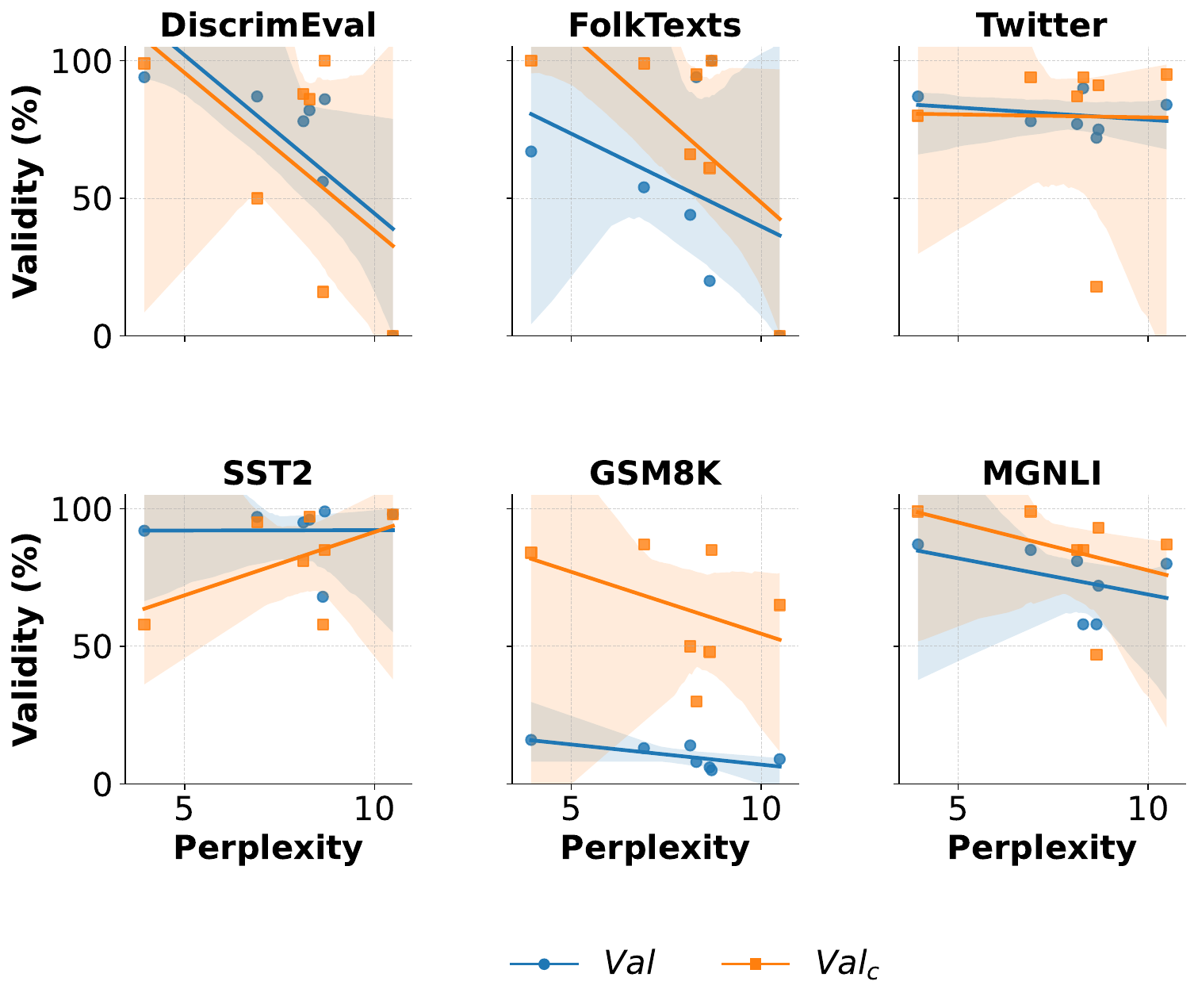}
    \caption{Regression plots of few-shot perplexity versus \SCE validity across datasets. Blue lines indicate validity without context (\Val) and orange lines indicate validity with context (\ValH), with shaded regions denoting 95\% confidence intervals.}
    \label{fig:model_plp_reg_vs_validity}
\end{figure}

\begin{figure}[ht]
    \centering
    \includegraphics[width=\linewidth]{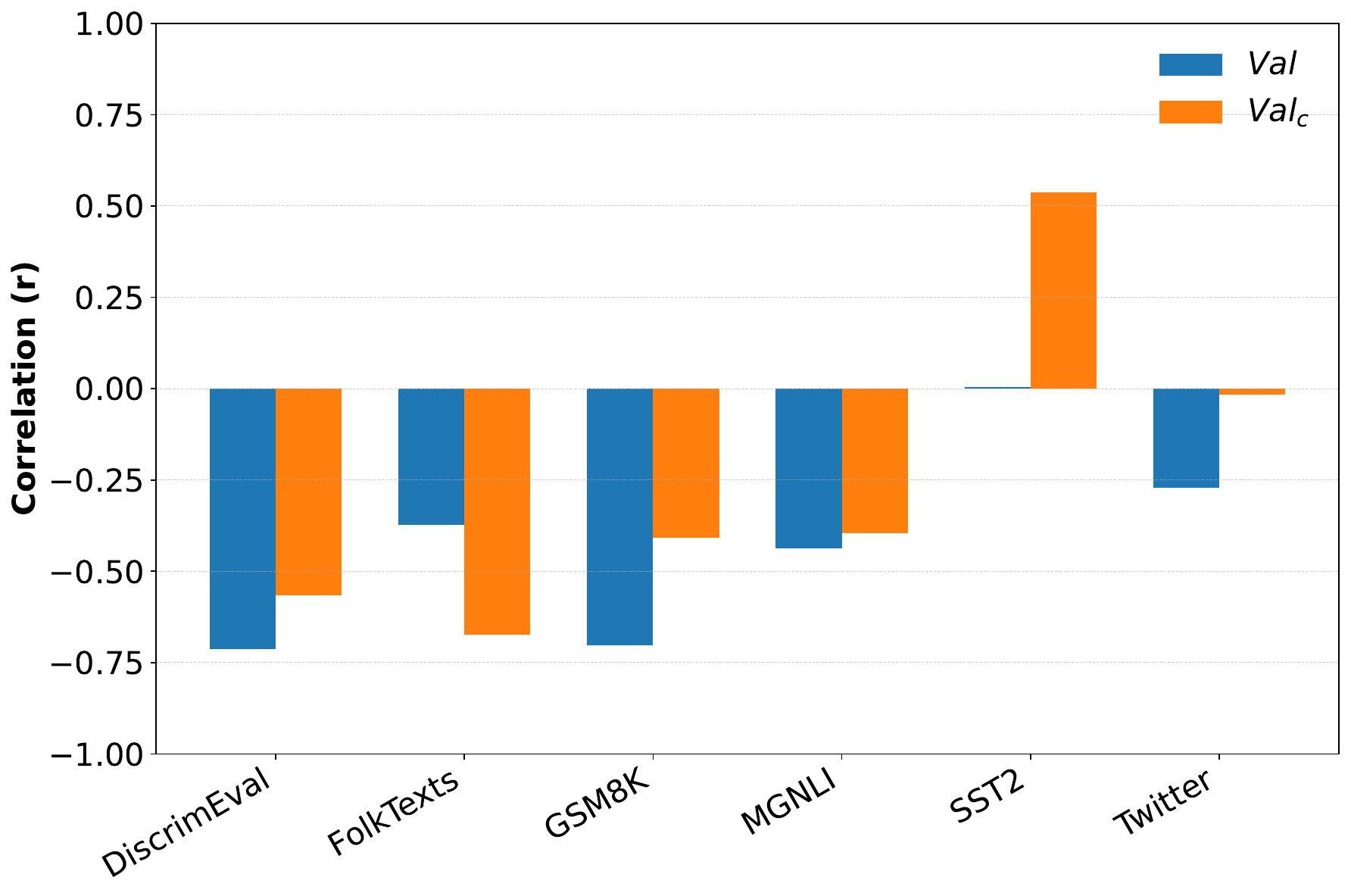}
    \caption{Pearson correlation coefficients between few-shot perplexity and \SCE validity across datasets. Blue bars represent validity without context (\Val) and orange bars represent validity with context (\ValH).}
    \label{fig:pearson_plp_validity}
\end{figure}

\xhdr{Model perplexity vs. \SCEs validity}
We used the lm-eval framework%
\footnote{\url{https://github.com/EleutherAI/lm-evaluation-harness}}
to compute five-shot perplexity on the \textsc{Wikitext}~\cite{merity2016pointer} benchmark for each model, and then analyzed its correlation with the percentage of valid \SCEs generated. The decision to use lm-eval aligns with best practices for reproducible, transparent, and comparable evaluation, as emphasized by~\citet{biderman2024lessons}. By adopting a controlled few-shot setup, we reduce variance across evaluations and ensure our perplexity scores reflect meaningful differences in model behavior rather than implementation artifacts. Measuring perplexity in this standardized way enables a principled comparison with \SCEs validity, allowing us to probe whether language models with lower perplexity exhibit stronger counterfactual reasoning.
However, as shown in line plots (\autoref{fig:model_plp_vs_validity}), regression fits (\autoref{fig:model_plp_reg_vs_validity}), and correlation analysis (\autoref{fig:pearson_plp_validity}), we did not observe a clear relationship between few-shot perplexity and \SCE validity across models.

\section{Annotation Protocol}
\label{app:AnnotProt}
We conducted a human annotation study, as reported in Section~\ref{sec:CharaFail}.  
The protocol was as follows. We randomly selected $50$ examples from \textsc{GSM8K} under CoT prompting at $T=0$, for each of the $7$ models, resulting in $350$ examples overall. Each example was independently assessed by two annotators (the authors), who determined whether the \SCE yielded a solution matching the correct target label ($\hat{y}_{CE}$). Disagreements, observed in roughly $5\%$ of the cases, were resolved through in-person discussion. These disagreements typically arose from ambiguities in the counterfactual statements or occasional mistakes in solving the math problems. The resulting consensus labels were then used to compute correlations between \textbf{validity} and \textbf{correctness}.

\section{Clustering of \SCE Representations: Methodology and Results}
\label{app:FailAnal}
As introduced in Section~\ref{sec:CharaFail}, we applied $K$-means clustering to the embedding space of \SCEs in order to probe potential task misunderstandings. In the following, we detail the methodology and results of this analysis, focusing on the systematic differences in the hidden representations of valid and invalid \SCEs. We evaluated whether different clustering strategies and distance metrics provide consistent separation between valid and invalid \SCEs. Specifically, we compared three strategies: using the representations at the \textit{First Generated Token} and \textit{Last Generated Token} of the \SCE, and the \textit{Last Input Token} of the prompt that elicited the \SCE. For each strategy, we evaluated four distance metrics: raw Euclidean distance, normalized Euclidean distance, raw cosine distance (that is, 1 - cosine similarity), and normalized cosine distance. Here, ``normalized'' means that last-layer hidden-state vectors were standardized to zero mean and unit variance before distance computation.
We ran k-means clustering with each of the above four metrics as the distance metric.
To quantify performance, we define the \textbf{average separation score} as:
\[
\text{SepScore} = \frac{1}{N} \sum_{i=1}^{N} \big( \Delta_{0}^{(i)} + \Delta_{1}^{(i)} \big),
\]
where $\Delta_{0}^{(i)}$ and $\Delta_{1}^{(i)}$ are the absolute differences between valid and invalid \SCEs in clusters~0 and 1 for the $i$-th (model, dataset) pair, and $N$ is the total number of evaluated pairs.

When averaging across all models and datasets, we found that the separation scores do not differ much between various distance metrics and that normalized cosine distance yielded the highest separation score (178.9), outperforming raw Cosine (176.5), normalized Euclidean (175.7), and raw Euclidean (175.2).
Therefore, we adopted \textbf{normalized Cosine distance} as our primary metric.

Detailed results for each model and dataset are reported in \autoref{table:combined_strategies}, where $\Delta_0$ and $\Delta_1$ denote the absolute difference between valid and invalid cases assigned to cluster~0 and cluster~1, respectively. Larger $\Delta$ values indicate clearer separation. For example, \textsc{GSM8K} shows consistently low $\Delta$ scores, suggesting weaker separation, whereas \textsc{Twitter} and \textsc{SST2} yield higher $\Delta$ values, indicating stronger clustering of valid vs.\ invalid cases.
\begin{table*}[ht]
\small
\centering
\begin{subtable}{\textwidth}
\centering
\resizebox{\textwidth}{!}{
\begin{tabular}{l
  c c @{\hskip 6pt} c c @{\hskip 12pt} %
  c c @{\hskip 6pt} c c @{\hskip 12pt} %
  c c @{\hskip 6pt} c c @{\hskip 12pt} %
  c c @{\hskip 6pt} c c @{\hskip 12pt} %
  c c @{\hskip 6pt} c c @{\hskip 12pt} %
  c c @{\hskip 6pt} c c                %
}

\toprule
 & \multicolumn{4}{c}{\textbf{DEV}} & \multicolumn{4}{c}{\textbf{TWT}} & \multicolumn{4}{c}{\textbf{SST}} & \multicolumn{4}{c}{\textbf{FLK}} & \multicolumn{4}{c}{\textbf{NLI}} & \multicolumn{4}{c}{\textbf{MTH}} \\
 & \multicolumn{2}{c}{$\Delta_0$} & \multicolumn{2}{c}{$\Delta_1$} & \multicolumn{2}{c}{$\Delta_0$} & \multicolumn{2}{c}{$\Delta_1$} & \multicolumn{2}{c}{$\Delta_0$} & \multicolumn{2}{c}{$\Delta_1$} & \multicolumn{2}{c}{$\Delta_0$} & \multicolumn{2}{c}{$\Delta_1$} & \multicolumn{2}{c}{$\Delta_0$} & \multicolumn{2}{c}{$\Delta_1$} & \multicolumn{2}{c}{$\Delta_0$} & \multicolumn{2}{c}{$\Delta_1$} \\
 & w/o & w/ & w/o & w/ & w/o & w/ & w/o & w/ & w/o & w/ & w/o & w/ & w/o & w/ & w/o & w/ & w/o & w/ & w/o & w/ & w/o & w/ & w/o & w/ \\
\midrule
\textbf{\llamaS} & 22 & 3 & 11 & 5 & 424 & 356 & 70 & 203 & 306 & 92 & 108 & 66 & 272 & 197 & 97 & 88 & 185 & 144 & 87 & 69 & 1 & 11 & 8 & 23 \\
\textbf{\llamaM} & 27 & 21 & 33 & 17 & 429 & 432 & 80 & 317 & 244 & 258 & 236 & 242 & 92 & 91 & 249 & 183 & 404 & 465 & 131 & 211 & 44 & 90 & 32 & 119 \\
\textbf{\mistralS} & 10 & 18 & 11 & 22 & 539 & 275 & 103 & 62 & 26 & 15 & 63 & 40 & 90 & 52 & 100 & 109 & 142 & 101 & 135 & 40 & 6 & 3 & 32 & 27 \\
\textbf{\mistralM} & 26 & 28 & 31 & 39 & 246 & 97 & 189 & 377 & 161 & 157 & 256 & 271 & 111 & 121 & 283 & 3 & 238 & 130 & 163 & 145 & 7 & 13 & 27 & 49 \\
\textbf{\gemmaS} & 19 & 2 & 22 & 1 & 402 & 402 & 65 & 19 & 231 & 235 & 155 & 143 & 98 & 45 & 104 & 71 & 264 & 141 & 427 & 147 & 2 & 20 & 3 & 4 \\
\textbf{\gemmaM} & 32 & 0 & 28 & 0 & 33 & 84 & 358 & 379 & 181 & 174 & 187 & 187 & 325 & 282 & 171 & 94 & 382 & 369 & 148 & 100 & 1 & 119 & 2 & 60 \\
\textbf{\rd} & 4 & 15 & 34 & 9 & 52 & 64 & 29 & 93 & 8 & 10 & 29 & 35 & 197 & 88 & 234 & 65 & 107 & 54 & 96 & 14 & 89 & 65 & 78 & 91 \\
\bottomrule
\end{tabular}
}
\caption{Clustering results using the \textbf{first generated token representation}. 
Entries show $\Delta_0$ and $\Delta_1$ (absolute differences between valid and invalid \SCEs in clusters~0 and~1) under the w/o (without context) and w/ (with context) settings.}

\label{table:combined_FirstTok}
\end{subtable}

\vspace{0.3cm}
\begin{subtable}{\textwidth}
\centering
\resizebox{\textwidth}{!}{
\begin{tabular}{l
  c c @{\hskip 6pt} c c @{\hskip 12pt} %
  c c @{\hskip 6pt} c c @{\hskip 12pt} %
  c c @{\hskip 6pt} c c @{\hskip 12pt} %
  c c @{\hskip 6pt} c c @{\hskip 12pt} %
  c c @{\hskip 6pt} c c @{\hskip 12pt} %
  c c @{\hskip 6pt} c c                %
}
\toprule
 & \multicolumn{4}{c}{\textbf{DEV}} & \multicolumn{4}{c}{\textbf{TWT}} & \multicolumn{4}{c}{\textbf{SST}} & \multicolumn{4}{c}{\textbf{FLK}} & \multicolumn{4}{c}{\textbf{NLI}} & \multicolumn{4}{c}{\textbf{MTH}} \\
 & \multicolumn{2}{c}{$\Delta_0$} & \multicolumn{2}{c}{$\Delta_1$} & \multicolumn{2}{c}{$\Delta_0$} & \multicolumn{2}{c}{$\Delta_1$} & \multicolumn{2}{c}{$\Delta_0$} & \multicolumn{2}{c}{$\Delta_1$} & \multicolumn{2}{c}{$\Delta_0$} & \multicolumn{2}{c}{$\Delta_1$} & \multicolumn{2}{c}{$\Delta_0$} & \multicolumn{2}{c}{$\Delta_1$} & \multicolumn{2}{c}{$\Delta_0$} & \multicolumn{2}{c}{$\Delta_1$} \\
 & w/o & w/ & w/o & w/ & w/o & w/ & w/o & w/ & w/o & w/ & w/o & w/ & w/o & w/ & w/o & w/ & w/o & w/ & w/o & w/ & w/o & w/ & w/o & w/ \\
\midrule
\textbf{\llamaS} & 17 & 2 & 16 & 0 & 417 & 362 & 77 & 197 & 198 & 67 & 216 & 91 & 19 & 48 & 350 & 61 & 153 & 116 & 119 & 97 & 9 & 26 & 0 & 8 \\
\textbf{\llamaM} & 28 & 19 & 32 & 19 & 310 & 462 & 199 & 287 & 247 & 261 & 233 & 239 & 98 & 20 & 243 & 72 & 319 & 373 & 216 & 303 & 57 & 147 & 19 & 62 \\
\textbf{\mistralS} & 10 & 17 & 11 & 23 & 320 & 187 & 322 & 150 & 206 & 170 & 117 & 145 & 137 & 85 & 53 & 76 & 104 & 183 & 173 & 42 & 36 & 6 & 2 & 30 \\
\textbf{\mistralM} & 24 & 26 & 33 & 41 & 200 & 253 & 235 & 221 & 205 & 214 & 212 & 214 & 142 & 191 & 252 & 309 & 301 & 228 & 100 & 47 & 25 & 46 & 9 & 16 \\
\textbf{\gemmaS} & 24 & 1 & 17 & 2 & 159 & 210 & 178 & 211 & 205 & 197 & 181 & 181 & 86 & 118 & 116 & 2 & 138 & 94 & 553 & 194 & 1 & 21 & 4 & 3 \\
\textbf{\gemmaM} & 24 & 1 & 36 & 1 & 78 & 122 & 247 & 341 & 50 & 40 & 318 & 321 & 328 & 283 & 168 & 95 & 293 & 300 & 237 & 169 & 1 & 125 & 0 & 54 \\
\textbf{\rd} & 4 & 15 & 34 & 9 & 66 & 70 & 43 & 99 & 59 & 47 & 38 & 2 & 220 & 98 & 211 & 55 & 73 & 22 & 130 & 46 & 78 & 54 & 89 & 102 \\
\bottomrule
\end{tabular}
}
\caption{Clustering results using the \textbf{last input token representation}. 
Entries show $\Delta_0$ and $\Delta_1$ under w/o (without context) and w/ (with context).}

\label{table:combined_LastInput}
\end{subtable}

\vspace{0.3cm}
\begin{subtable}{\textwidth}
\centering
\resizebox{\textwidth}{!}{
\begin{tabular}{l
  c c @{\hskip 6pt} c c @{\hskip 12pt} %
  c c @{\hskip 6pt} c c @{\hskip 12pt} %
  c c @{\hskip 6pt} c c @{\hskip 12pt} %
  c c @{\hskip 6pt} c c @{\hskip 12pt} %
  c c @{\hskip 6pt} c c @{\hskip 12pt} %
  c c @{\hskip 6pt} c c                %
}

\toprule
 & \multicolumn{4}{c}{\textbf{DEV}} & \multicolumn{4}{c}{\textbf{TWT}} & \multicolumn{4}{c}{\textbf{SST}} & \multicolumn{4}{c}{\textbf{FLK}} & \multicolumn{4}{c}{\textbf{NLI}} & \multicolumn{4}{c}{\textbf{MTH}} \\
 & \multicolumn{2}{c}{$\Delta_0$} & \multicolumn{2}{c}{$\Delta_1$} & \multicolumn{2}{c}{$\Delta_0$} & \multicolumn{2}{c}{$\Delta_1$} & \multicolumn{2}{c}{$\Delta_0$} & \multicolumn{2}{c}{$\Delta_1$} & \multicolumn{2}{c}{$\Delta_0$} & \multicolumn{2}{c}{$\Delta_1$} & \multicolumn{2}{c}{$\Delta_0$} & \multicolumn{2}{c}{$\Delta_1$} & \multicolumn{2}{c}{$\Delta_0$} & \multicolumn{2}{c}{$\Delta_1$} \\
 & w/o & w/ & w/o & w/ & w/o & w/ & w/o & w/ & w/o & w/ & w/o & w/ & w/o & w/ & w/o & w/ & w/o & w/ & w/o & w/ & w/o & w/ & w/o & w/ \\
\midrule
\textbf{\llamaS} & 26 & 2 & 7 & 0 & 257 & 280 & 237 & 279 & 231 & 85 & 183 & 73 & 239 & 37 & 130 & 72 & 135 & 74 & 137 & 139 & 2 & 12 & 7 & 22 \\
\textbf{\llamaM} & 32 & 26 & 28 & 12 & 401 & 425 & 108 & 324 & 225 & 245 & 255 & 255 & 247 & 8 & 94 & 84 & 226 & 327 & 309 & 349 & 45 & 111 & 31 & 98 \\
\textbf{\mistralS} & 10 & 22 & 11 & 18 & 400 & 207 & 242 & 130 & 118 & 114 & 29 & 89 & 86 & 78 & 104 & 83 & 171 & 18 & 106 & 159 & 39 & 78 & 1 & 54 \\
\textbf{\mistralM} & 32 & 40 & 25 & 27 & 390 & 282 & 45 & 192 & 202 & 210 & 215 & 218 & 192 & 168 & 202 & 50 & 151 & 61 & 250 & 214 & 24 & 34 & 10 & 28 \\
\textbf{\gemmaS} & 15 & 5 & 26 & 8 & 67 & 86 & 270 & 335 & 314 & 305 & 72 & 73 & 193 & 49 & 9 & 67 & 270 & 154 & 421 & 134 & 0 & 23 & 5 & 1 \\
\textbf{\gemmaM}  & 41 & 0 & 19 & 0 & 73 & 50 & 398 & 413 & 135 & 127 & 233 & 234 & 82 & 12 & 414 & 176 & 234 & 369 & 296 & 100 & 6 & 67 & 7 & 112 \\
\textbf{\rd} & 5 & 24 & 25 & 0 & 48 & 36 & 25 & 65 & 18 & 50 & 39 & 95 & 219 & 107 & 212 & 46 & 105 & 140 & 98 & 208 & 75 & 77 & 92 & 79 \\
\bottomrule
\end{tabular}
}
\caption{Clustering results using the \textbf{last generated token representation}. 
Entries show $\Delta_0$ and $\Delta_1$ under w/o (without context) and w/ (with context).}

\label{table:combined_LastGen}
\end{subtable}
\caption{Comparison of clustering strategies for separating valid vs. invalid \SCEs. 
Each panel reports results for one token-based representation (first generated token, last input token, last generated token). 
Performance is measured by $\Delta_0$ and $\Delta_1$, which quantify how well valid and invalid cases are separated within clusters under both w/o (without context) and w/ (with context) settings, where larger values indicate stronger separation.}

\label{table:combined_strategies}
\end{table*}

\section{Statistical Significance via Permutation Testing}
\label{sec:permtest}
To complement the confidence interval comparisons reported in the \autoref{table:direct_prompt_temp0} and \autoref{table:rationale_prompting_temp0}, we additionally performed nonparametric permutation tests to assess whether the differences between the two conditions (with context and without context) are statistically significant.
We applied paired permutation tests with the null hypothesis that the two conditions are drawn from the same distribution, \ie, any observed difference in validity or normalized edit distance arises purely from random variation in the sample. In each test, the assignment of condition labels was randomly permuted across paired examples, and the distribution of mean differences was computed over $10,000$ resamples. Two-sided p-values were then obtained by comparing the observed effect size to this null distribution.
\autoref{tables:permu} reports the effect size (mean difference between the two conditions) for both validity and normalized edit distance under two prompting strategies:
(i) Unconstrained prompting ($T=0$; see \autoref{tab:direct_validity} and~\autoref{tab:direct_edit}), and
(ii) Rationale-based prompting ($T=0$; see \autoref{tab:rationale_validity} and~\autoref{tab:rationale_edit}).
The table shows that when comparing validity, permutation testing detects more statistically significant differences than CI overlap alone. The effect magnitude varies across datasets and prompting strategies.

\begin{table*}[!t]
\small
    \centering
    \begin{subtable}[t]{0.48\textwidth}
        \centering
        \resizebox{0.6\textwidth}{!}{
        \begin{tabular}{l c c c c c c} 
            \toprule
            {} & \textbf{\DSEV} & \textbf{\TWTR} & \textbf{\SST} & \textbf{\FOLK} & \textbf{\NLI} & \textbf{\MATH}\\
            \midrule
            \textbf{\llamaS} & \textbf{-55} & \textbf{-55} & 0 & \textbf{-81} & \textbf{15} & \textbf{43} \\
            \textbf{\llamaM} & 4 & \textbf{1} & \textbf{-33} & \textbf{33} & \textbf{12} & \textbf{33} \\
            \textbf{\mistralS} & \textbf{16} & \textbf{6} & \textbf{1} & 1 & \textbf{26} & \textbf{44} \\
            \textbf{\mistralM} & \textbf{13} & \textbf{16} & \textbf{1} & \textbf{46} & \textbf{15} & \textbf{54} \\
            \textbf{\gemmaS} & \textbf{N/A} & \textbf{14} & 1  & \textbf{N/A} & \textbf{16} & -8 \\
            \textbf{\gemmaM} & \textbf{11} & \textbf{16} & \textbf{-14} & 0 & \textbf{21} & \textbf{33} \\
            \textbf{\rd} & \textbf{25} & \textbf{25} & \textbf{-14} & \textbf{50} & \textbf{25} & \textbf{21} \\
            \bottomrule
        \end{tabular}
        }
        \caption{Unconstrained prompting: effect size on \textbf{validity}.}

        \label{tab:direct_validity}
    \end{subtable}
    \hfill
 \begin{subtable}[t]{0.48\textwidth}
        \centering
        \resizebox{0.6\textwidth}{!}{
        \begin{tabular}{l c c c c c c} 
            \toprule
            {} & \textbf{\DSEV} & \textbf{\TWTR} & \textbf{\SST} & \textbf{\FOLK} & \textbf{\NLI} & \textbf{\MATH}\\
            \midrule
            \textbf{\llamaS} & -19 & \textbf{-8} & 0 & \textbf{7} & 0 & -2 \\
            \textbf{\llamaM} & 0 & 1 & 2 & 0 & 0 & 1 \\
            \textbf{\mistralS} & -2 & -1 & 0 & \textbf{-1} & 0 & -1 \\
            \textbf{\mistralM} & 0 & 0 & 0 & 0 & 0 & 0 \\
            \textbf{\gemmaS} & N/A & -1 & 0 & N/A & 0 & -10 \\
            \textbf{\gemmaM} & -1 & 0 & -1 & 0 & 0 & -4 \\
            \textbf{\rd} & 2 & \textbf{-3} & \textbf{-2} & \textbf{-1} & 0 & -6 \\
            \bottomrule
        \end{tabular}
        }
        \caption{Unconstrained prompting: effect size on \textbf{normalized edit distance}.}

        \label{tab:direct_edit}
    \end{subtable}
\vspace{3mm}
    \begin{subtable}[t]{0.48\textwidth}
        \centering
        \resizebox{0.6\textwidth}{!}{
    \begin{tabular}{l c c c c c c} 
        \toprule
        {} & \textbf{\DSEV} & \textbf{\TWTR} & \textbf{\SST} & \textbf{\FOLK} & \textbf{\NLI} & \textbf{\MATH}\\
        \midrule
        \textbf{\llamaS} & \textbf{42} & \textbf{12} & \textbf{10} & \textbf{30} & \textbf{28} & -43 \\
        \textbf{\llamaM} & \textbf{9} & -1 & \textbf{-14} & \textbf{38} & \textbf{23} & \textbf{15} \\
        \textbf{\mistralS} & \textbf{-60} & \textbf{-1} & -1 & \textbf{-5} & \textbf{-56} & \textbf{-100} \\
        \textbf{\mistralM} & \textbf{40} & \textbf{24} & \textbf{12} & \textbf{50} & \textbf{99} & \textbf{39} \\
        \textbf{\gemmaS} & N/A & \textbf{19} & \textbf{18} & N/A & \textbf{23} & 23 \\
        \textbf{\gemmaM} & \textbf{51} & \textbf{14} & \textbf{7} & \textbf{36} & \textbf{25} & \textbf{21} \\
        \textbf{\rd} & \textbf{45} & \textbf{8} & \textbf{-16} & \textbf{38} & \textbf{24} & \textbf{44} \\

        \bottomrule
    \end{tabular}
    }
    \caption{Rationale-based prompting: effect size on \textbf{validity}.}

    \label{tab:rationale_validity}
     \end{subtable}
    \hfill
    \begin{subtable}[t]{0.48\textwidth}
        \centering
        \resizebox{0.6\textwidth}{!}{
    \begin{tabular}{l c c c c c c}
        \toprule
        {} & \textbf{\DSEV} & \textbf{\TWTR} & \textbf{\SST} & \textbf{\FOLK} & \textbf{\NLI} & \textbf{\MATH}\\
        \midrule
        \textbf{\llamaS} & 3 & \textbf{-8} & \textbf{-4} & \textbf{10} & 0 & -12 \\
        \textbf{\llamaM} & -1 & \textbf{-4} & 2 & 0 & \textbf{-5} & \textbf{-6} \\
        \textbf{\mistralS} & -6 & \textbf{-1} & 0 & -2 & \textbf{-1} & 0 \\
        \textbf{\mistralM} & -1 & \textbf{1} & -1 & 0 & 0 & 0 \\
        \textbf{\gemmaS} & N/A & -1 & -2 & N/A & 0 & -5 \\
        \textbf{\gemmaM} & -1 & -1 & -2 & 0 & 0 & -2 \\
        \textbf{\rd} & 1 & -1 & \textbf{-7} & \textbf{-5} & -1 & 6 \\
        \bottomrule
    \end{tabular}
    }
   \caption{Rationale-based prompting: effect size on \textbf{normalized edit distance}.}
     \label{tab:rationale_edit}
\end{subtable}
\caption{Effect sizes (mean difference between with-context and without-context conditions) for validity and normalized edit distance under two prompting strategies (unconstrained and rationale-based) at $T=0$ across datasets. 
Positive values indicate higher scores with context (\ValH) compared to without context (\Val), and bolded entries mark statistically significant differences.}
    \label{tables:permu}
    \end{table*}

\section{Bootstrap Confidence Intervals}
\label{app:bootstrap_ci}

To avoid reliance on normality assumptions and to allow for asymmetric intervals, we computed confidence intervals for the normalized differences in \SCE lengths using nonparametric bootstrap resampling \citep{tibshirani1993introduction}.
Specifically, $10{,}000$ bootstrap samples with replacement were drawn from the valid and invalid counterfactual length distributions. 
For each resample, we calculated the normalized difference, and reported the bootstrap mean together with the 2.5$^{\text{th}}$ and 97.5$^{\text{th}}$ percentiles. 
This yields a 95\% confidence interval that does not rely on normality assumptions and naturally accommodates asymmetry. 
The original results with normality-based intervals are provided in \autoref{app:additional_results}.

\end{document}